\documentclass[twoside,11pt]{article}

\usepackage[abbrvbib, preprint]{jmlr2e}

\usepackage{microtype}
\usepackage{graphicx}
\usepackage{subfigure}
\usepackage{booktabs} %
\usepackage{amssymb}
\usepackage{graphics}
\usepackage{multirow}
\usepackage{amsmath}
\usepackage[dvipsnames]{xcolor}
\usepackage{floatrow}
\usepackage{enumitem}

\usepackage{tablefootnote}

\usepackage{nicefrac}
\usepackage{amsfonts} %

\usepackage[toc,page,header]{appendix}
\usepackage{minitoc}
\usepackage{placeins}

\usepackage{hyperref}
\usepackage{caption}

\usepackage{pifont}%
\newcommand{\xmark}{\ding{55}}%
\newcommand{\dashrule}[1][black]{%
  \color{#1}\rule[\dimexpr.5ex-.2pt]{4pt}{.4pt}\xleaders\hbox{\rule{4pt}{0pt}\rule[\dimexpr.5ex-.2pt]{4pt}{.4pt}}\hfill\kern0pt%
}

\usepackage{color,soul,colortbl}
\definecolor{lightcyan}{rgb}{0.92, 1.0, 1.0}
\definecolor{pink}{rgb}{0.99, 0.76, 0.8}
\definecolor{lightgray}{gray}{0.95}
\definecolor{fbteal}{HTML}{e6f5f0}

\newcommand{\rownumber}[1]{\textcolor{gray}{#1}}
\newcommand{\sem}[1]{{\scriptsize $\pm$#1}}
\newcommand{\gsem}[1]{\textcolor{gray}{\scriptsize{$\pm$#1}}}
\newcommand{\na}{}
\newcommand{\gdel}[1]{(\textbf{\textcolor{ForestGreen}{#1}}) }
\newcommand{\rdel}[1]{(\textbf{\textcolor{BrickRed}{#1}}) }

\definecolor{darkgray}{rgb}{0.6,0.6,0.6}
\definecolor{babyblue}{rgb}{0.54, 0.81, 0.94}
\definecolor{citrine}{rgb}{0.89, 0.82, 0.04}
\definecolor{misogreen}{rgb}{0.25,0.6,0.0}
\definecolor{PalePurp}{rgb}{0.66,0.57,0.66}
\definecolor{todocolor}{rgb}{0.66,0.99,0.99}
\definecolor{pearcomp}{HTML}{B97E29}
\definecolor{pearDark}{HTML}{2980B9}
\definecolor{pearDarker}{HTML}{1D2DEC}

\usepackage{color}
\usepackage{xcolor}

\newcommand{\bgrm}{\texttt{BG\textunderscore RM}}
\newcommand{\bgswaps}{\texttt{BG\textunderscore Swaps}}
\newcommand{\bgrand}{\texttt{BG\textunderscore Random}}

\newcommand{\pneg}{$p_{\text{neg}}$}
\newcommand{\ppos}{$p_{\text{pos}}$}

\newcommand{\moco}{MoCo-v2}
\newcommand{\byol}{BYOL}
\newcommand{\swav}{SwAV}

\ShortHeadings{Background Augmentations for Self-Supervised Learning}{Ryali, Schwab and Morcos}
\firstpageno{1}

\begin{document}

\title{Characterizing and Improving the Robustness \\ of Self-Supervised Learning through  \\ Background Augmentations}

\author{\name Chaitanya K.\ Ryali\thanks{Work done while an intern and student researcher at Facebook AI Research (FAIR).} \email rckrishn@eng.ucsd.edu \\
       \addr Department of Computer Science and Engineering\\
       University of California San Diego, CA, USA\\
       \AND
       \name David J.\ Schwab \email dschwab@gc.cuny.edu \\
       \addr ITS, CUNY Graduate Center, NY, USA\\ 
       Facebook AI Research, NY, USA\\
       \AND
       \name Ari S.\ Morcos \email arimorcos@fb.com \\
       \addr Facebook AI Research, CA, USA\\
       }

\editor{}

\maketitle

\begin{abstract}%
Recent progress in self-supervised learning has demonstrated promising results in multiple visual tasks. An important ingredient in high-performing self-supervised methods is the use of data augmentation by training models to place different augmented views of the same image nearby in embedding space. However, commonly used augmentation pipelines treat images holistically, ignoring the semantic relevance of parts of an image—e.g. a subject \textit{vs}. a background—which can lead to the learning of spurious correlations. Our work addresses this problem by investigating a class of simple, yet highly effective “background augmentations", which encourage models to focus on semantically-relevant content by discouraging them from focusing on image backgrounds. Through a systematic investigation, we show that background augmentations lead to substantial improvements in performance across a spectrum of state-of-the-art self-supervised methods (\moco, BYOL, SwAV) on a variety of tasks, e.g. $\sim+$1-2\% gains on ImageNet, enabling performance on par with the supervised baseline. Further, we find the improvement in limited-labels settings is even larger (up to 4.2\%). Background augmentations also improve robustness to a number of distribution shifts, including natural adversarial examples, ImageNet-9, adversarial attacks, ImageNet-Renditions. We also make progress in completely unsupervised saliency detection, in the process of generating saliency masks used for background augmentations. 
\end{abstract}

\begin{keywords}
  self-supervised learning, contrastive learning, representation learning, background augmentation, out-of-distribution generalization, robustness
\end{keywords}

\section{Introduction} \label{sec:intro}

Learning useful representations in the absence of labels is a critical challenge in machine learning. Recently, self-supervised (SSL) methods such as SimCLR \citep{chen2020simple}, \moco~\citep{he2019moco, chen2020mocov2}, BYOL \citep{grill2020bootstrap}, and SwAV \citep{swav} have risen to prominence because they are able to produce high-quality representations that rival supervised representations on vision tasks. These methods differ in the details of their approach---e.g. some are instance based (\moco, SimCLR) while others are cluster based (SwAV), some explicitly utilize negatives while others do not (BYOL), and some use a memory bank (\moco). In fact, competitive performance has recently been achieved by SimSiam \citep{simsiam} without any of these additions. However, a central ingredient common to all high performing SSL methods is their reliance on {\em data augmentation} as a means of encoding desired invariances. Two \textit{views} of an image are created via independent samples from the data augmentation pipeline, and the objective is \textit{view-invariance}, i.e. the encoder is trained to place them near each other in representational space. Thus, the choice of data augmentation is critical, as augmentations and the invariances they encourage are the primary teaching signal these methods utilize to create semantically meaningful representations. 

In fact, \citet{chen2020simple} explored a large space of standard augmentations and demonstrated that the choice of these augmentations can have dramatic effects on performance. However, this standard suite of augmentations used in most SSL methods was modified from augmentations designed for supervised approaches. It may therefore be useful to design new augmentation schemes for SSL that specifically target semantic focus for this setting. 

A parallel line of inquiry has found that supervised models often rely on non-semantic features that may nonetheless be predictive at test time. Models often overly focus on backgrounds \citep{xiao2020noise, sehwag2020time, beery_recognition_2018}, are brittle to distribution shift in foreground-background statistics, and rely on high-frequency information \citep{jo2017measuring, ilyas_neurips_2020_features_not_bugs}. Models are also susceptible to adversarial attacks \citep{goodfellow_fgsm, jo2017measuring}, often rely on texture over shape \citep{geirhos2019imagenettrained, geirhos2020surprising, hermann2019origins} and are brittle to distribution shift in local texture (e.g. paintings, sculpture, \citet{hendrycks_many_2021}) as well as to corruptions (e.g. blur, contrast, \citet{hendrycks_benchmarking_2019}). Importantly, the benefits or limitations of a modeling choice on robustness are not apparent from metrics on standard tasks \citep{hendrycks_using_2019}. All of these results showcase the need for comprehensive model evaluation across diverse data sets and settings. We broadly encompass such comprehensive evaluation under robustness, e.g. robustness to distribution shifts (e.g. paintings, blurring, different background statistics), robustness to adversarial attacks, robustness to label scarcity.

While there has been much work investigating robustness properties in the supervised setting, the self-supervised setting has received relatively less attention. As SSL methods shrink the gap to their supervised counterparts, it has become increasingly important to characterize their robustness properties and gain a more holistic understanding.
The aim of this work is twofold: characterizing the robustness of high performing SSL methods and investigating approaches for improved semantic focus via a class of augmentations called background augmentations.

We conduct a systematic, comprehensive investigation of the robustness properties of SSL methods as well as the impact of background augmentations in improving semantic focus across \textit{a}) a spectrum of high performing SSL methods, \textit{b}) training durations, \textit{c}) three variants of background augmentations, \textit{d}) different foreground extraction methods used in background augmentations, and \textit{e}) a wide range of downstream data sets and tasks, including 17 distribution shift settings.

Specifically, we study three classes of approaches: \bgrm, in which a subset of backgrounds are removed during the augmentation process, \bgrand, in which backgrounds are replaced with random backgrounds from other images in the mini-batch, and \bgswaps, in which a selection of backgrounds are swapped between positive and negative images to match backgrounds across the query and the negative, thereby explicitly penalizing background focus. 

We highlight the following contributions:
\begin{itemize}[leftmargin=2em]
    
    \item \textbf{Novel background augmentation method.} We develop and analyze a novel, highly effective background augmentation method \bgswaps, which manipulates the backgrounds of positives and negatives in a structured manner, yielding large performance and robustness benefits.

    \item \textbf{Sizeable performance benefits.} We show sizeable performance improvements for all view-invariant SSL methods, yielding consistent improvements of $\sim$1-2\% in linear evaluation on ImageNet; these improvements allow us to reach an accuracy of 76.1\% (63.8\%) on ImageNet (ImageNet-v2), on par with the standard supervised baseline 76.4\% (63.8\%) for ResNet-50. Further, background augmentations enable us to reach a benchmark accuracy of 74.4\%, outperforming Barlow Twins \citep{zbontar_barlow_2021}, MoCo-v3 \citep{chen_empirical_2021} and BYOL trained for 800-1000 epochs in \textit{only} 100 epochs; this result takes a large step forward in reducing the amount of training required for competitive performance in SSL.
    
    In the \textit{limited-label setting}, we show the performance benefits are even larger, e.g. in the 1\% (ImageNet) label setting, \bgswaps~confers a 4.0\% accuracy gain for \moco~and in the 10\% label setting \bgrand~enables BYOL to reach 72\% accuracy using only 10\% of ImageNet labels.
    
    \item \textbf{Improved robustness.} We find that background augmentations (especially \bgswaps) lead to significantly improved robustness in many settings including ImageNet-9 (shift in foreground-background statistics), ImageNet-A (natural adversarial examples), ImageNet-R (ImageNet-Renditions), against adversarial attack, and ImageNet ReaL.

    \item \textbf{Scientific Insight.} We investigate the impact of background augmentations in \textit{a}) the supervised setting and \textit{b}) RotNet, and find that they do not confer a performance gain, giving us insight into \textit{when} and \textit{how} background augmentations work. We also gain further insight by shape-bias probing as well as by systematically perturbing the quality of the augmentations.
    
    \item \textbf{Improvement in saliency detection.} In order to separate foregrounds and backgrounds without any supervision, we also make progress in completely unsupervised saliency detection, matching or outperforming weakly supervised as well as many supervised methods.
    
\end{itemize}

\section{Methods} \label{sec:methods}

\subsection{Self-Supervised Learning Methods}
We consider a diverse test bed of high performing self-supervised learning methods: \textbf{\moco} \citep{chen2020mocov2}, \textbf{BYOL} \citep{grill2020bootstrap}, and \textbf{SwAV} \citep{swav} to ensure generality of our results. As in the respective original works, we use a standard ResNet-50 \citep{resnet} as the default architecture in all experiments (SSL and supervised) unless otherwise noted. A small subset of our experiments are based on \textbf{RotNet}\citep{gidaris2018unsupervised}, using an AlexNet \citep{alexnet} architecture following the respective original work. All reported numbers are based on our reproduction unless otherwise stated. Where possible, we follow the protocol from the original works.

Here, we provide a brief overview of \moco, \byol~and \swav~and some implementation details, with further details in Appendix \ref{app:training_details}. We defer an overview of RotNet to Section \ref{sec:rotnet} and relegate implementation details to Appendix \ref{app:training_details}.

\paragraph{Overview.} Broadly, each method uses a pair of Siamese networks \citep{siamese_bromley_1994}---i.e. weight-sharing neural networks, to encode differently augmented ``views" of the same image and maximize similarity between them, thereby encouraging the learning of ``desirable" invariances. Concretely, two views $v_{s},v_{t}$ of an image $x$ are generated by sampling from a random augmentation pipeline. The student network $f_{\theta}$ is used to encode $v_{s}$ as $z_s=f_{\theta}(v_{s})$ and similarly the teacher network\footnote{The weight-sharing between the student and teacher may be direct as $\xi \leftarrow \theta$ (as in SwAV) or indirect as $\xi \leftarrow m\xi+(1-m)\theta$, where $\xi$ is an exponential moving average of $\theta$ (as in \moco~and BYOL).} $f_{\xi}$, is used to encode $v_{t}$ as $z_t=f_{\xi}(v_{t})$. Then, $z_s$ is used to \textit{predict} a target generated from $z_t$; the specific form of this   pretext prediction task varies with the SSL method. Learning/``pre-training" is by optimization of the prediction loss over $\theta$.

\begin{figure}
    \centering
    \includegraphics[width=0.97\textwidth]{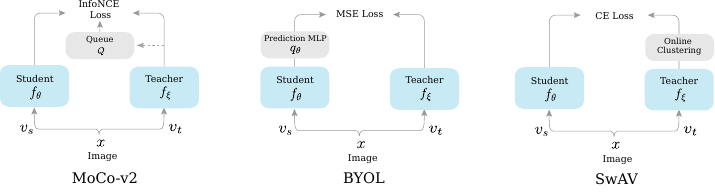}
    \caption{ \textbf{Schematic of Siamese SSL methods.} A simplified schematic of the Siamese SSL methods in our test bed. Dashed line in \moco~denotes enqueuing the positives $k^+$ from the previous mini-batch (and dequeuing the oldest mini-batch).}
    \label{fig:ssl_methods_summary}
\end{figure}

\textbf{\moco}~is an instance of \textit{contrastive} learning \citep{hadsell_dimensionality_2006}, a framework for learning representations from data that are organized into similar/dissimilar pairs. The prediction task in \moco~is one of \textit{instance discrimination}: a differently augmented view of the same image $x$ needs to be discriminated from a set $Q$ of ``distractors"---views of images different from $x$, in a ($|Q|+1$)-way classification. Two images form a similar/positive pair if they are views of the same image and otherwise form a negative pair. \moco~uses the InfoNCE \citep{infonce_2018} loss for this task and instantiates $Q$ as a queue comprised of previous mini-batches of $\ell_2$ normalized outputs from the teacher. The prediction is $\bar{z}_s$ and the target is $\bar{z}_t$, where $\bar{z}=\nicefrac{z}{||z||_2}$.

In the terminology of the original work, the prediction $\bar{z}_s$ is called the query (denoted $q$), the target $\bar{z}_t$ is called the positive key (denoted $k^+$) and the distractors (here elements of $Q$) are known as negatives keys (denoted $Q=\{k^-\}$). Thus, the loss encourages similarity between $q$ and $k^+$ and dissimilarity between $q$ and $k^-$. 

 In \textbf{BYOL}, a prediction Multi-Layer Perceptron (MLP) $q_{\theta}$ is used to generate the prediction $\overline{q_{\theta}}(z_{s})$, the target is $\bar{z}_t$ and the loss used is Mean Squared Error (MSE). In \textbf{SwAV}, the target is generated by an online-clustering process and $\bar{z}_s$ is used to predict the cluster assignment of $\bar{z}_t$; the loss used is Cross-Entropy (CE). Thus, SwAV is a \textit{clustering-based} approach, while \moco~and \byol~are \textit{instance-based} approaches. \swav~and \byol~are not explicitly contrastive, since they do use negative instances.

All methods use 2 ``global" views, while SwAV additionally uses $L$ ``local" views---low resolution crops that cover only small parts of the image; by default $L=6$. Using global and local views is known as \texttt{multi-crop} augmentation. Local views are typically only used for prediction and not used in generating the targets. Intuitively, since local views are expected to be predictive of global views, models are discouraged from representing only the most discriminative features for solving the pretext prediction task.

It is typical to use a projection MLP \citep{chen2020simple} on top of a backbone network and discard the projection MLP after pre-training (but see \citet{chen2020big}). In our notation, $f$ subsumes the backbone $g$ and the projection MLP $h$, i.e. $f=h\circ g$. At the end of pre-training, only the backbone $g_{\theta}$ is kept. The outputs of $g_{\theta}$ are called representations and the corresponding outputs of $h$ are called the embeddings/projections.

\textit{(Abuse of) Notation}: For simplicity, we refer to the embedding from the student network as the query $q$ and the embedding from the teacher corresponding to the same image $x$ as the positive key $k^+$, across all methods. We also use the terms student (teacher) and query (key) network interchangeably.

\paragraph{Implementation.}
\moco~is trained using SGD and a larger (than the standard 256) batch size  of 1024 (distributed across 32 GPUs) with a 20 epoch linear warmup for 220 (800) epochs in the medium (full) setting. These settings were chosen to increase training speed while matching the reported performance at a similar number of epochs in \citet{chen2020mocov2}.

BYOL and SwAV were trained using LARS \citep{you2017large} using a batch size of 4096, distributed across 64 GPUs with synchronized batch normalization \citep{ioffe_batch_2015} for groups of 8 GPUs. BYOL (SwAV) is trained for 300 (100) epochs in the medium setting and 1000 (800) epochs in the full setting. See Appendix \ref{app:training_details} for more details.

\begin{figure}
    \centering
    \includegraphics[width=0.9\textwidth]{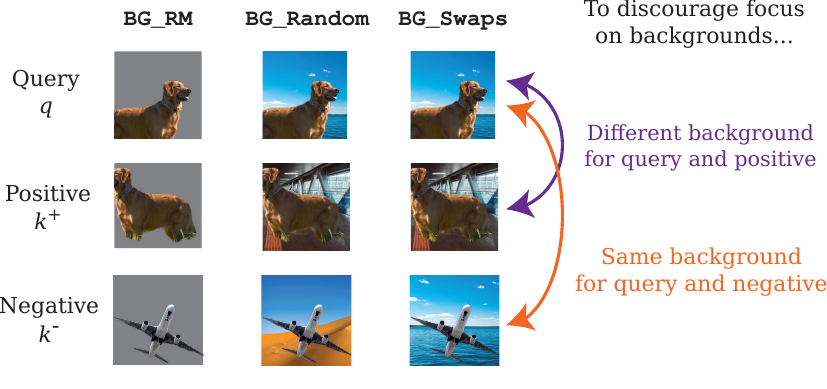}
    \caption{\textbf{Schematic of different types of background augmentations.} \bgrm~(left) replaces backgrounds with grayscale, effectively removing any background information. \bgrand~(middle) replaces backgrounds with random backgrounds, creating a random signal which is uncorrelated with the foreground. \bgswaps~(right) exploits the structure of contrastive learning to ensure that the query and the positive have the same foreground but different backgrounds, while the query and one negative have matched backgrounds. As a result, \bgswaps~makes it so that models are penalized for focusing on the background.}
    \label{fig:schematic}
\end{figure}

\subsection{Background Augmentations}
\label{sec:bg_augs_methods}
We apply all background augmentations (\bgrm, \bgrand, \bgswaps) after all other augmentations in the respective augmentation pipeline. However, we note that we observed similar results applying background augmentations before all other augmentations as well (Appendix \ref{app:aug_order}). While we apply background augmentations to (views of) \textit{images}, when it is clear from context, we will refer instead to the corresponding embeddings $q, k^+, k^-$. Unless otherwise mentioned, background augmentations are applied independently with a probability $p_{\text{pos}}$ to both $q$ and $k^+$ (the positive teaching pair). When a method has explicit negative instances (\moco), we denote by $p_{\text{neg}}$ the probability of including a negative whose background matches $q$; by default, this is independent of background augmentation in $q$ and $k^+$. Values for \ppos~and \pneg~were optimized independently for each background augmentation. When it is clear from context, we will sometimes drop the subscript. Note that in \moco, $k^+$ is placed in the queue $Q$ for use in subsequent batches as a negative, so that augmentations applied to $k^+$, also indirectly apply to $k^-$ via $Q$.  When \texttt{multi-crop} augmentation is used (as in SwAV), we apply background augmentations only to the global views. Background augmentations are only applied during self-supervised pre-training and are not applied when training linear classification layers for evaluation. Below, we describe the details of each of the background augmentations we study.

In \bgrm, the background of an image is removed by using a foreground mask (obtained using a saliency detector, see Section \ref{sec:saliency_det}), and replaced with a solid grayscale background whose intensity is drawn uniformly from $[0,1]$, though we note that a solid black background produced similar results. See illustrative examples in Figure \ref{fig:schematic}, left column. 

In \bgrand, we replace the background with a background from a different image in the same batch. As in \citet{xiao2020noise}, tiled backgrounds corresponding to an image are generated by filling in the foreground information using the surrounding background.

In \bgswaps, we generate a negative image with a background \textit{matched} to that of the query $q$. In practice, we create a background matched negative as $m_q r+(1-m_q)q$, where $m_q$ is the binary foreground mask of the query $q$ and $r$ is a random image. We generate all foreground masks and tiled backgrounds offline and cache them to increase throughput at train time. Note that foreground masks may include multiple foreground objects when they are present (e.g. last row of Figure \ref{fig:saliency_examples} or Figure \ref{fig: bbox_mask}). Substantial noise is tolerable in the quality of the foreground masks (see Appendix \ref{app:noisy_masks}). More generally, there is substantial flexibility and tolerance in instantiating the main ingredients of background augmentations, which we expand on in Appendix \ref{app:ablations}. 

\subsection{Supervised Training}
We largely follow the protocol from \citet{goyal_accurate_2018}, unless otherwise indicated. We train all supervised models (with or without background augmentation) with a batch size of 4096 with a 5 epoch linear warmup due to the large batchsize.  Models are trained for 90 epochs, with a step schedule (30, 60, 80) and a decay factor of 0.1, using SGD with a base learning rate of 0.1 scaled linearly ($lr$=BatchSize/256$\times0.1$) and momentum  of 0.9, and the standard augmentations \texttt{RandomResizedCrop} and \texttt{RandomHorizontalFlip}. We also exclude bias and batch normalization parameters from weight decay, which was set to $1\times10^{-4}$. The $\gamma$ in each residual block’s last BatchNorm layer is zero initialized. Our supervised baseline for ResNet-50 reaches the standard baseline \citep{goyal_accurate_2018} performance of $\sim$76.4\% Top-1 accuracy on ImageNet \citep{imnet1k}.

\section{Saliency Detection} 
\label{sec:saliency_det}

We use saliency detection to generate the foreground masks used in background augmentations (see methods, Section \ref{sec:bg_augs_methods}). However, state-of-the-art saliency detection methods (e.g. U$^2$Net, \citet{u2net}) are generally reliant on manually annotated, accurate pixel-level Ground Truth (GT) saliency labels for training, making their usage inappropriate in a truly self-supervised benchmark.

\subsection{Weakly Supervised Saliency Detection} 
Recent ``unsupervised" saliency detection methods \citep{nguyen_deepusps_2019, zhang_deep_2018, zhang_supervision_2017} demonstrate promising results by leveraging psuedo-labels generated by hand-crafted saliency methods in lieu of manually annotated GT saliency labels. Briefly, noisy psuedo-labels generated by hand-crafted saliency methods are iteratively \textit{refined} by using them as targets to train a Fully Convolutional Network (FCN) for saliency detection, and obtaining refined pseudo-labels from the denoised predictions. Refined pseudo-labels from multiple hand-crafted methods are then jointly used to train a re-initialized FCN to obtain the final saliency detector. While these methods are ``unsupervised" in that they do not use manually annotated saliency labels, their success implicitly relies on human annotation---the FCN used is pre-trained in a \textit{supervised} manner using ImageNet class and CityScapes \citep{cordts_cityscapes_2016} segmentation labels. Indeed, we find that if we use a randomly initialized FCN instead, the resulting saliency predictions are \textit{worse} than the noisy psuedo-labels used as targets. As such, these methods are also not appropriate to generate foreground masks for our purpose; we thus refer to these methods as weakly supervised methods in this context.

\begin{figure}
    \centering
    \includegraphics[width=0.45\textwidth]{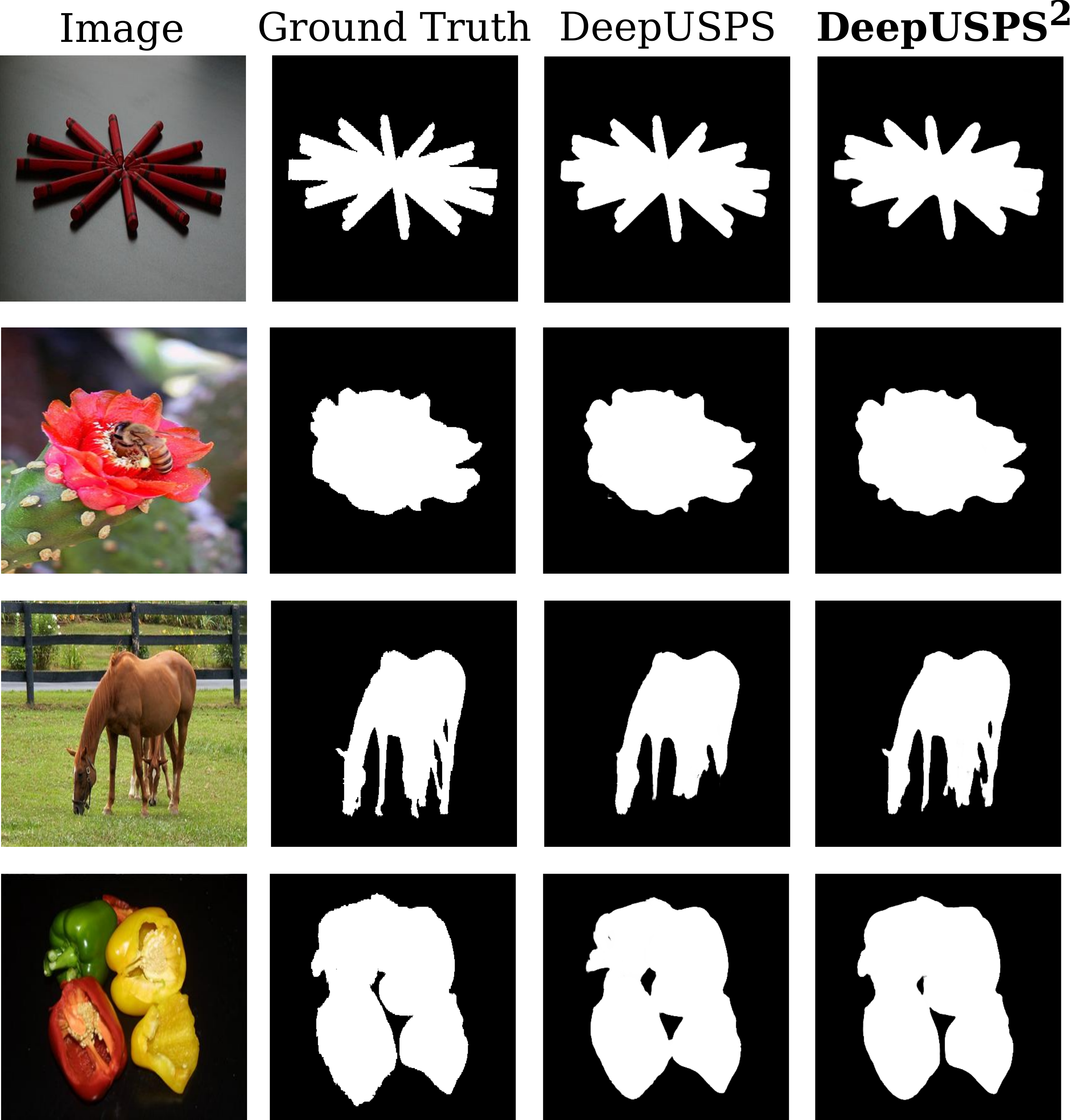}
    \caption{ Examples of saliency masks generated by DeepUSPS$^2$.}
    \label{fig:saliency_examples}
\end{figure}

\begin{table}[ht]
    \centering
    \begin{tabular}{lcccccc}\toprule
 Method  & \multicolumn{2}{c}{MSRA-B}  & \multicolumn{2}{c}{ECSSD} & \multicolumn{2}{c}{DUT}
    \\\cmidrule(lr){2-3} \cmidrule(lr){4-5} \cmidrule(lr){6-7} 
 &  F$\uparrow$ &  MAE$\downarrow$ & F$\uparrow$ &  MAE$\downarrow$ & F$\uparrow$ &  MAE$\downarrow$\\ \midrule
\multicolumn{7}{c}{\shortstack{Supervised \\ {\small (GT saliency labels used for training.)} } }\\ \midrule
\cite{hou_deeply_2017} & 89.4 & 4.7 & 88.0 & 7.0 & 72.9 & 7.6 \\
\cite{luo_non-local_2017} & 89.7 & 4.8 & 89.1 & 6.6 & 73.6 & 8.0 \\
\cite{zhang_amulet_2017} & - & - & 88.3 & 6.1 & 69.3 & 9.8\\
\cite{zhang_learning_2017} & - & - & 85.2 & 8.0 & 66.0 & 13.2\\
\cite{wang_stagewise_2017} & 85.1 & 6.7 & 82.6 & 9.2 & 67.2 & 8.5\\
\cite{li_deepsaliency_2016} & - & - & 75.9 & 16.0 & 60.5 & 7.6 \\
\cite{wang_saliency_2016} & - & - & 84.3 & 9.7 & 69.2 & 9.5\\
\midrule
\multicolumn{7}{c}{\shortstack{Weakly Supervised \\ {\small (Class labels used in pre-trained backbone, GT saliency labels not used in training.)} } }\\ \midrule
SBF \footnotesize{\citep{zhang_supervision_2017}} & - &  - &  78.7  & 8.5  & 58.3 & 13.5 \\
USD \footnotesize{\citep{zhang_deep_2018}}   & 87.7 & 5.6 & 87.8 & 7.0 & 71.6 & 8.6 \\
DeepUSPS & 90.3 & 4.0 & 87.4 & 6.3 & \textbf{73.6} & \textbf{6.3} \\
DeepUSPS {\scriptsize{(\textit{repro.})}} & 90.5{\scriptsize{$\pm$0.1}} & 3.9{\scriptsize{$\pm$0.0}} & 87.9{\scriptsize{$\pm$0.1}} & 6.3{\scriptsize{$\pm$0.0}} & 72.1{\scriptsize{$\pm$0.2}} & 6.8{\scriptsize{$\pm$0.1}} \\
\midrule
\multicolumn{7}{c}{\shortstack{Completely Unsupervised \\ {\small (No human annotation at any stage in the pipeline.)}}}\\
\midrule
\rowcolor{lightgray}
DeepUSPS$^2$ {\scriptsize{(\textit{ours})}} & \textbf{91.3}{\scriptsize{$\pm$0.0}} & \textbf{3.6}{\scriptsize{$\pm$0.0}} & \textbf{90.0}{\scriptsize{$\pm$0.0}}
& \textbf{5.4}{\scriptsize{$\pm$0.0}} & 71.1{\scriptsize{$\pm$0.0}} & 6.9{\scriptsize{$\pm$0.0}}
\\\bottomrule
    \end{tabular}
    \caption{\textbf{DeepUSPS$^2$ is on par with or outperforms weakly supervised saliency methods and several recent supervised saliency methods.} We report performance across 5 independent runs for DeepUSPS$^2$ (and also for DeepUSPS {\scriptsize{(\textit{repro.})}}). Notation: Mean$\pm$SEM (Standard Error of the Mean). Best results are in \textbf{bold}.}
    \label{tab:saliency_deepusps}
\end{table}

\subsection{Unsupervised Saliency Detection: DeepUSPS\texorpdfstring{$^2$}{2}}
In order to train a completely unsupervised saliency detector, we build upon DeepUSPS \citep{nguyen_deepusps_2019}, a recent state-of-the-art weakly supervised saliency detection method. We first pre-train a DRN-D-105 \citep{drn_yu_2017} network in a \textit{self-supervised} manner for 500 epochs on ImageNet, using BYOL. We then use this pre-trained network to refine pseudo-labels and train a saliency detector, which we call DeepUSPS$^2$, employing a training protocol modified from DeepUSPS (see Appendix \ref{app:train_saliency_impl_details}); some example saliency predictions are shown in Figure \ref{fig:saliency_examples}. Training images were 2500 images from the MSRA-B data set \citep{liu_learning_2011}.

We find that DeepUSPS$^2$ \textit{performs better than or on par} with DeepUSPS and other recent state-of-the-art weakly supervised and even some supervised saliency detectors on common saliency benchmark data sets MSRA-B, ECSSD \citep{yan_hierarchical_2013}, and DUT \citep{yan_hierarchical_2013}, yet DeepUSPS$^2$ does not rely on \textit{any human annotation at any stage in the pipeline}, see Table \ref{tab:saliency_deepusps}. For each data set, following common protocol \citep{nguyen_deepusps_2019, achanta_frequency-tuned_2009}, we report the F-score, 
\begin{equation*}
F_{\beta}=\frac{(1+\beta^2)\times\text{precision} \times \text{recall}}{\beta^2\times\text{precision}+\text{recall}},    
\end{equation*}
where $\beta^2=0.3$ to weigh precision more than recall and the MAE (Mean Absolute Error) on the test split. 

We use DeepUSPS$^2$ as the default saliency detector to generate foreground masks in our experiments unless otherwise indicated. To ablate the method of mask generation and for control experiments, we also use U$^2$Net \citep{u2net}, a state-of-the-art saliency detector that is trained in a \textit{supervised} manner on DUTS-TR \citep{duts_tr_wang2017}, which contains 10553 pixel-level manual saliency annotations. 

\section{Representation Learning with Background Augmentations} 
\label{sec:imnet_eval}

\begin{table}
    \centering
    \resizebox*{!}{0.825\textheight}{
     \begin{tabular}{lccc}\toprule
    Method &  Epochs & \multicolumn{2}{c}{ImageNet acc.} \\
           &   &  Original & ReaL \\\midrule
    \textcolor{gray}{Supervised}  & \textcolor{gray}{90} & \textcolor{gray}{76.4} & \textcolor{gray}{82.7} \\\midrule
    PCL-v2  \footnotesize{\citep{li_prototypical_2021}}    & 200 & 67.6 & -  \\ %
    CMC  \footnotesize{\citep{tian_contrastive_2020}}   & 200 & 66.2 &- \\ %
    SimCLR & 200 & 66.8 &- \\ %
    MoCo   & 200 & 60.6 & - \\ %
    SeLa  \footnotesize{\citep{asano_self-labelling_2020}}   & 400 & 61.5 & - \\ %
    \moco & 200 & 67.5 & - \\ %
    \rowcolor{lightgray}
    \moco~{\scriptsize{(\textit{repro.})}} & 220 & 67.7 & 74.7  \\
    \rowcolor{lightgray}
    \moco~+ \bgrm & 220 & 69.1{\scriptsize{$\pm$0.0}} (\textbf{\textcolor{ForestGreen}{+1.4}}) & 76.2{\scriptsize{$\pm$0.0}} (\textbf{\textcolor{ForestGreen}{+1.5}}) \\
    \rowcolor{lightgray}
    \moco~+ \bgswaps \footnote{We show \bgswaps~$>$~\bgrand~for \moco, see section \ref{sec:bgswap}.  \bgswaps~does not apply to BYOL and SwAV as they do not use negative instances.}
    & 220 & 69.5{\scriptsize{$\pm$0.1}} (\textbf{\textcolor{ForestGreen}{+1.8}}) 
    & 76.6{\scriptsize{$\pm$0.1}} (\textbf{\textcolor{ForestGreen}{+1.9}}) \\
    DiLo (\moco)  \footnotesize{\citep{ZhaoAAAI2021}} & 200 & 67.9 (\textbf{\textcolor{ForestGreen}{+0.2}}) & - \\
    BYOL & 300 & 72.5 & - \\
    \rowcolor{lightgray}
    BYOL {\scriptsize{(\textit{repro.})}} & 300 & 72.7 & 79.6  \\
    \rowcolor{lightgray}
    BYOL + \bgrm & 300 & 73.3{\scriptsize{$\pm$0.2}} (\textbf{\textcolor{ForestGreen}{+0.6}}) 
    & 80.4{\scriptsize{$\pm$0.3}} (\textbf{\textcolor{ForestGreen}{+0.8}}) \\
    \rowcolor{lightgray}
    BYOL + \bgrand & 300 & \textbf{73.9}{\scriptsize{$\pm$0.1}} (\textbf{\textcolor{ForestGreen}{+1.2}}) 
    & \textbf{81.0}{\scriptsize{$\pm$0.0}} (\textbf{\textcolor{ForestGreen}{+1.4}}) \\
    SwAV & 100 & 72.1 & - \\
    \rowcolor{lightgray}
    SwAV {\scriptsize{(\textit{repro.})}} & 100 & 72.2  & 79.1  \\
    \rowcolor{lightgray}
    SwAV + \bgrm & 100 & 73.6{\scriptsize{$\pm$0.1}} (\textbf{\textcolor{ForestGreen}{+1.4}})  
    & 80.6{\scriptsize{$\pm$0.1}} (\textbf{\textcolor{ForestGreen}{+1.5}}) \\
    \rowcolor{lightgray}
    SwAV + \bgrand & 100 & 73.4{\scriptsize{$\pm$0.0}} (\textbf{\textcolor{ForestGreen}{+1.2}}) 
    & 80.4{\scriptsize{$\pm$0.1}} (\textbf{\textcolor{ForestGreen}{+1.3}}) \\
    \midrule
    \multicolumn{1}{@{}l}{\textit{Longer Training}}\\
    PIRL  \footnotesize{\citep{misra2020pirl}}   & 800 & 63.6 & - \\ %
    SimCLR & 1000 & 69.3 & - \\ %
    Barlow Twins  \footnotesize{\citep{zbontar_barlow_2021}}  & 1000 & 73.2 & -\\
    \moco~& 800 & 71.1  & - \\
    \rowcolor{lightgray}
    \moco~{\scriptsize{(\textit{repro.})}} & 800 & 71.0 & 78.0 \\
    \rowcolor{lightgray}
    \moco~+ \bgrm & 800 & 71.9 (\textbf{\textcolor{ForestGreen}{+0.9}}) & 78.9 (\textbf{\textcolor{ForestGreen}{+0.9}}) \\ %
    \rowcolor{lightgray}
    \moco~+ \bgswaps & 800 & 72.2 (\textbf{\textcolor{ForestGreen}{+1.2}}) & 79.2 (\textbf{\textcolor{ForestGreen}{+1.2}})  \\ %
    BYOL & 1000 & 74.3 & - \\
    \rowcolor{lightgray}
    BYOL {\scriptsize{(\textit{repro.})}} & 1000 & 73.8  & 80.5\\
    \rowcolor{lightgray}
    BYOL + \bgrm & 1000 & 74.6 (\textbf{\textcolor{ForestGreen}{+0.8}}) & 81.3 (\textbf{\textcolor{ForestGreen}{+0.8}})  \\ %
    \rowcolor{lightgray}
    BYOL + \bgrand & 1000 & 74.8 (\textbf{\textcolor{ForestGreen}{+1.0}}) & 81.7 (\textbf{\textcolor{ForestGreen}{+1.2}})  \\ %
    SwAV & 800 & 75.3  & - \\
    \rowcolor{lightgray}
    SwAV {\scriptsize{(\textit{repro.})}} & 800 & 74.9 & 81.4 \\
    \rowcolor{lightgray}
    SwAV + \bgrm & 800 & \textbf{76.1} (\textbf{\textcolor{ForestGreen}{+1.2}}) & \textbf{82.5} (\textbf{\textcolor{ForestGreen}{+1.1}}) \\
    \rowcolor{lightgray}
    SwAV + \bgrand & 800 & \textbf{76.1} (\textbf{\textcolor{ForestGreen}{+1.2}}) & \textbf{82.6} (\textbf{\textcolor{ForestGreen}{+1.2}}) \\\bottomrule
    \end{tabular}
    }
    \caption{
    \looseness=-1
   \textbf{Background augmentations confer large performance benefits in linear evaluation on \textbf{ImageNet} across a spectrum of SSL methods using the original or reassessed labels.} For shorter training, we report metrics averaged over 3 independent runs, reflecting robust improvements. Number of training epochs are chosen to be consistent with previously published results. We highlight \textbf{\textcolor{ForestGreen}{performance gains}} due to background augmentations relative to our reproductions, but also include published baseline numbers for comparison. Notation: Mean$\pm$SEM (Standard Error of the Mean). Best results are in \textbf{bold}.
    }
    \label{tab:gain_summary_imgnet_deepusps}
\end{table}

\subsection{Do Background Augmentations that Encourage Semantic Focus Increase Performance?}

Deep neural networks often rely on non-semantic, superficial features and thus may be easily misled by backgrounds. Nonetheless, these non-semantic features are often predictive at test time \citep{xiao2020noise, sehwag2020time, ilyas_neurips_2020_features_not_bugs}, so it is not a priori obvious whether background augmentations that encourage semantic focus on the foreground will benefit performance. We investigate this question by exploring the space of possible background augmentations. First, we study removing backgrounds {\em probabilistically}, where the strength of the augmentation is controlled by a parameter $p$, which sets the probability that the background is removed from the query or positive key. See Figure \ref{fig:schematic}, left for an example of the \bgrm~setting.

Across SSL methods, we find that \bgrm~substantially improves linear classification on ImageNet, improving performance by $\sim$0.6-1.4\% (Table \ref{tab:gain_summary_imgnet_deepusps}). For all methods, we found that a moderate value of $p$ between 0.1 and 0.3 is generally a good setting. However, despite its improved performance, because \bgrm~introduces images with solid gray backgrounds, it induces a distribution shift between the unsupervised pre-training phase and the supervised downstream tasks which may limit performance improvements. Note that DiLo (\moco) is similar to \bgrm~applied \moco, but results only in a small gain of +0.2, while we obtain a 7$\times$ larger gain (and as we will show later, a 9$\times$ gain by developing an improved background augmentation method \bgswaps).

\subsection{Can we make Background Augmentations more In-Distribution?} 
\label{sec:bgrand}

In the previous section, we explored removing backgrounds and replacing them with uniform grayscale, which results in the data being out-of-distribution (OOD) relative to the downstream tasks. To mitigate this OOD issue, we instead replace backgrounds with a randomly chosen background from another instance in the same batch. We term this method \bgrand~(Figure \ref{fig:schematic}, middle). Interestingly, despite the fact that \bgrand~is more in-distribution than \bgrm, we found that performance was similar (e.g. 69.1\% for \bgrm~(Table \ref{tab:gain_summary_imgnet_deepusps}) \textit{vs}. 69.2\% for \bgrand~(Table \ref{tab:moco_bg_aug_strength_A}) with \moco). However, we note that these two settings are not necessarily directly comparable. For example, in the case of \moco, augmented positive keys are added to the queue to be used as negatives for subsequent mini-batches. As a result, \bgrm~might actually penalize background focus whereas \bgrand~may simply result in an uninformative background. This is because \bgrm~features a constant gray background which can be matched between the query and negatives that were used as positive keys in a previous mini-batch, whereas \bgrand~features distinct backgrounds for each augmented image.

\begin{table}
    \centering
    \begin{tabular}{@{}p{1.75em}@{}lccccc} \toprule
    & Method & \multicolumn{3}{@{}c@{}}{BG aug. in} & ImageNet acc. \\
    \cmidrule(lr){3-5}
    & & $q$       & $k^+$ &  $k^-$ &  \\ \midrule
    \rownumber{(a)} & baseline    &           &    &   & 67.7\\
    \multicolumn{2}{@{}l}{\small{\textit{Control Experiments}}}\\
    \rowcolor{lightgray}
    \rownumber{(b)}& \bgrm   & \checkmark &   &  & 67.8   \\
    \rowcolor{lightgray}
    \rownumber{(c)}& \bgrand     & \checkmark &   &  & \textbf{68.3}\\
   \multicolumn{2}{@{}l}{\small{\textit{``Full" Augmentations}}}\\
    \rownumber{(d)}& \bgrm       & \checkmark & \checkmark  & \checkmark & \textbf{69.3}   \\
    \rownumber{(e)}& \bgrand      & \checkmark & \checkmark  &  & 69.1\\ 
    \bottomrule
    \end{tabular}
    \caption{\textbf{\bgrm~\textit{vs}. \bgrand.} Comparing \bgrm~and \bgrand~in \moco~controlling for presence of negatives in the queue with similar background.}
    \label{tab:bg_rm_rand_control}
\end{table}

It is therefore unclear whether the similar performance of \bgrm~and \bgrand~stems from distributional shift or the matched gray backgrounds which can serve to make negatives more challenging. To disentangle these two factors, we performed a control experiment in which background augmented images were only included for the query (with $p=0.1$)---and thus \textit{not used} in subsequent mini-batches as the positive or the negative. This setting maintains the distribution shift of \bgrm, but removes the possibility of a teaching signal originating from matched gray backgrounds across the query and negative. To minimize confound stemming from mask quality, we use higher quality foreground masks generated by U$^2$Net, a state-of-the-art saliency detector trained with supervision instead of DeepUSPS$^2$.

This control reveals that, when only applied to the query but not the positive or negative, \bgrm~has similar performance (Table \ref{tab:bg_rm_rand_control}b) as \textit{no} \bgrm~(Table \ref{tab:bg_rm_rand_control}a), suggesting that \bgrm~benefits substantially from the teaching signal of negatives with matching (constant) backgrounds. In contrast, we found that, when only present in the query, \bgrand~still improves performance (Table \ref{tab:bg_rm_rand_control}c), but the improvement is decreased suggesting that having augmented images with randomized backgrounds in the positive keys provides additional benefit (Table \ref{tab:bg_rm_rand_control}e); here, for an apples-to-apples comparison with the control experiments (or ``partial" augmentations), we also report performance for the ``full" augmentations (Table \ref{tab:bg_rm_rand_control}d, e) using U$^2$Net masks and the same augmentation strength.

These analyses demonstrate both the importance of using background augmentations which remain close to the unaugmented input distribution and highlight the potential for methods which provide an additional teaching signal via negatives with query-matched backgrounds. Inspired by these results, we next investigate how to combine these approaches. 

\subsection{Exploiting the Structure of Contrastive Instance Discrimination via Background Matched Negatives} \label{sec:bgswap}

\begin{table}
    \centering
    \begin{tabular}{@{}p{1.7em}@{}lcccc} \toprule
         & Method & \multicolumn{3}{@{}c@{}}{BG aug. in} &  ImageNet acc. \\ 
        \cmidrule(lr){3-5}
         & & $q$ & $k^+$ & $k^-$ & \\ \midrule
         \rownumber{(a)}& baseline &  &   &   &  67.7\\
         \rownumber{(b)}& & \checkmark &  &   & 68.3\\
         \rownumber{(c)}& &  & \checkmark &  & 68.2\\
         \rownumber{(d)}& \bgrand & \checkmark &  \checkmark & & 69.1 \\
        \rowcolor{lightgray}  
         \rownumber{(e)}& \bgswaps & \checkmark & \checkmark & \checkmark & \textbf{69.7} \\
        \bottomrule
    \end{tabular}
    \caption{\textbf{Ablations of \bgswaps~for \moco.} Each component confers a performance improvement and the improvements stack on top of each other.}
    \label{tab:ablations}
\end{table}

Thus far, we have explored two background augmentations---\bgrm~ and \bgrand---both of which operate independently on the query and the positive and encourage semantic focus on foregrounds by simply removing backgrounds altogether or replacing them with randomized backgrounds so there's no effective signal in the background. This removes the incentive for models to focus on background information, but does nothing to directly penalize focus on backgrounds. However, contrastive instance discrimination (CID) methods (e.g., MoCo, SimCLR), use the query to  discriminate between the positive and \textit{negative instances} and thus feature structure that we can exploit to not only remove signal from backgrounds, but go further and provide \emph{explicitly misleading} signal in the backgrounds. Note that BYOL and SwAV are not CID methods, since they do not use negative instances.

We accomplish this through two modifications with a method we call \bgswaps~(Figure \ref{fig:schematic}, right). First, as in \bgrand, we ensure that the query and the positive feature \emph{distinct} random backgrounds. Models which focus on backgrounds would therefore place the positive and query further apart than they should since the semantic content is identical, but the background features differ. Second, we modify the negative set to include one additional negative\footnote{We explored using multiple background matched negatives, but found no improvement over a single matched negative. See Appendix \ref{app:ablations} for details.} whose background matches the query: a network which focuses on backgrounds would view the background-matched negative as highly similar to the query and receive strong negative supervision. As with \bgrm~and \bgrand, we introduce the background-matched negative with probability \pneg~(and include a randomly selected negative with probability $1-p_{\text{neg}}$, so that the total number of negatives is always $|Q| + 1$).

These background matched negatives can be considered an example of ``hard negatives", which have been explored recently in the context of SSL to improve learning \citep{kalantidis2020hard, wu2021conditional, robinson2021contrastive, cai_2020}. In this vein, one could consider the positive pair ($q$ and $k^+$) with different backgrounds as ``hard positives". For \moco, including a background matched negative further increases performance over \bgrm~by an additional 0.4\% (Table \ref{tab:gain_summary_imgnet_deepusps}). Consistent with our previous findings, we also found that it is important for the statistics of the augmentations to be similar for the positive and the negatives in order to achieve the best performance. In general, we found that the probability of an augmentation in the query and positive, \ppos, matching the probability of an augmentation in the negative, \pneg, so that \ppos $\simeq$\pneg, gives good performance.

\subsection{Ablating \bgswaps}
\label{sec:bgswap_ablations}

To characterize which components of the \bgswaps~augmentation matter and how much, we perform systematic ablations. As shown in Table \ref{tab:ablations}, we found that each independent component of \bgswaps~leads to a performance improvement (in contrast with \bgrm). In particular, we find benefits when employing each of the following: \bgrand~in the query ($p=0.1$); \bgrand~in the positive key ($p=0.1$); background matched negative ($p=0.2$). Notably, the improvements from randomized backgrounds in $q$ and $k^+$ stack superlinearly (Table \ref{tab:ablations}d), suggesting that incorporating both of these augmentations provides a greater advantage due to their interaction than either does independently; using background matched negatives further improves performance substantially (Table \ref{tab:ablations}e). As in the control experiments in Section \ref{sec:bgrand}, to minimize confound stemming from mask quality, we use higher quality foreground masks generated by U$^2$Net for these ablations.

There is significant design flexibility in how one could implement \bgswaps. For example, is it a better teaching signal to have independent  or correlated background augmentations in the query/positive and the negatives? Is it better to have a negative whose background matches the query or the positive? We find that \bgswaps~is robust to these specific choices (Appendix \ref{app:ablations}), making it a promising candidate for more general deployment in augmentation pipelines.

\subsection{Effect of Longer Training}

We also evaluate the impact of background augmentations on longer training ranging from 800 to 1000 epochs (Table \ref{tab:gain_summary_imgnet_deepusps}). As with the shorter training, we found that background augmentations consistently increased performance across models, e.g. enabling SwAV to reach 76.1\% with a ResNet-50 on ImageNet, only 0.3\% less than the standard supervised baseline. Interestingly, however, we found that the magnitude of the improvement decreased slightly in the longer training runs, which may be a saturation effect but also raises the more interesting possibility that SSL models initially learn representations that depend on backgrounds, but eventually learn some background invariance when trained for long enough. However, we later discuss (Section \ref{sec:bg_challenge}) evidence that does not find support for the latter possibility.

\subsection{Diagnosing and Improving SwAV + Background Augmentations}
\label{sec:diagnosis_swav}

\begin{figure}[t!]
\begin{minipage}{0.4\linewidth}
    \begin{tabular}{cccc}\toprule
        \multicolumn{4}{c}{\textbf{SwAV} w/ wider MLP (4096/256)} \\
        \rule{0pt}{3ex} Epochs & baseline & \bgrm & \bgrand\\ \midrule
        100  & 72.6 & 74.1 & 74.1 \\
        800  & 75.0 & 76.1 & 76.1 \\
        \bottomrule
    \end{tabular}
    \end{minipage}
    ~~~~~\begin{minipage}{0.5\linewidth}
    \centering
    \includegraphics[width=0.82\textwidth]{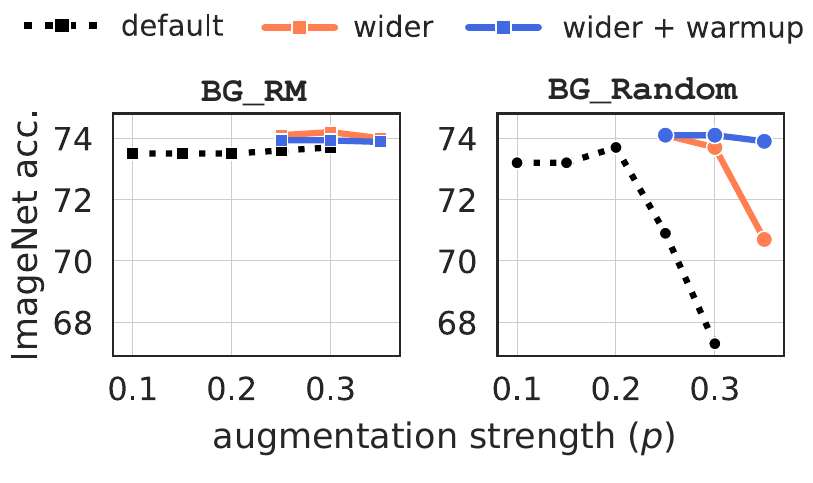} 
    \end{minipage}
    \caption{
    \textbf{Wider projection MLP and warmup alleviates early optimization difficulty.} (\textit{\textbf{left}})
    Wider projection MLP alleviates early optimization difficulty, improving performance and removes the gap between \bgrm~and \bgrand. Augmentation strength: $p=0.25$. 
    (\textit{\textbf{right}})
    The gap between \bgrm~and \bgrand~increases with stronger augmentation with the default (black dashed lines) MLP capacity. In addition to increasing MLP capacity, warming up background augmentations further adds  stability across a range of augmentation strengths. Notation: (MLP width/output dimension).
    }
    \label{tab:swav_wider_and_warmup}
\end{figure}

As previously discussed, due to \bgrm~being OOD, we might generally expect \bgrand, \bgswaps~to be on par or better than \bgrm. Our results in Table \ref{tab:gain_summary_imgnet_deepusps} show that while this is generally true across SSL methods and training durations, \bgrm~$>$~\bgrand~for SwAV trained for a short duration. Since  \bgrm~and \bgrand~result in the same final accuracy upon longer training (Table \ref{tab:gain_summary_imgnet_deepusps}), we hypothesized that there maybe early optimization difficulty arising from an interaction between SwAV's objective function and attempting to learning invariance to random \textit{natural} backgrounds (in contrast with solid grayscale backgrounds in \bgrm), at a stage in the pre-training when the representations are still quite poor. Consistent with this hypothesis, when \bgrand~is used, the loss lingers at chance early in pre-training, while the corresponding loss for \bgrm~falls rapidly. We reasoned that further increasing the augmentation strength of \bgrand~should result in higher optimization difficulty and consequently, worse performance. Consistent with this expectation, the performance of \bgrand~rapidly declines past a point, while the performance of \bgrm~remains stable, see Figure \ref{tab:swav_wider_and_warmup} (right, black dashed lines).

To alleviate this issue, we propose two solutions: \textit{a}) increasing the projection MLP capacity and \textit{b}) warming up background augmentations. We show results from (\textit{a}) in Figure \ref{tab:swav_wider_and_warmup} (left, Table), finding both improved performance (a baseline effect) and removing the gap between \bgrm~and \bgrand. Note that the default projection MLP capacity for SwAV is 2048/128. We report the results of (\textit{a}) and (\textit{b}) across a range of augmentation strengths in Figure \ref{tab:swav_wider_and_warmup} (right) \footnote{In setting of default MLP capacity (dashed lines), masks from U$^2$Net were used to control for influence of mask quality.}. In addition to increasing MLP capacity, warming up \bgrand~further stabilizes performance when stronger augmentation is used. More broadly, these analyses show that additional factors such as ease of optimization play an important role in determining performance apart from whether an augmentation induces a distribution shift. 

Our analyses here have broader implications. For instance, they shed new light on the role of the projection MLP and may help explain recent puzzling findings in literature; specifically, \citet{zbontar_barlow_2021} observed that their method, Barlow Twins, works best for large dimensionality of the projection MLP and noted that ``This result is quite surprising because the output...acts as a dimensionality bottleneck in our model and sets the limit of the intrinsic dimensionality of the representation". Our analyses suggest that it is important for the projection MLP to be of appropriate capacity for the pretext prediction task---more ``difficult" (e.g. due to stronger augmentation) prediction losses may benefit from a higher capacity MLP.

\begin{table}
    \centering
    \resizebox{1\columnwidth}{!}{
    \begin{tabular}{lccccc}\toprule
    Method &  Saliency  & \multicolumn{4}{c}{ImageNet acc.} \\
          &  Method &  \multicolumn{2}{c}{Original} & \multicolumn{2}{c}{ReaL} \\
            \cmidrule(lr){3-4} \cmidrule(lr){5-6}
          & & Top-1 & Top-5 & Top-1 & Top-5 \\\midrule
     \moco~(220)  &  \na & 67.7 & 88.1 & 74.7 & 91.7\\
    \multirow{2}{*}{+ \bgrm} & DeepUSPS$^2$  & 69.1\textcolor{gray}{{\scriptsize{$\pm$0.0}}} (\textbf{\textcolor{ForestGreen}{+1.4}}) & 88.8\textcolor{gray}{{\scriptsize{$\pm$0.0}}} & 76.2\textcolor{gray}{{\scriptsize{$\pm$0.0}}} (\textbf{\textcolor{ForestGreen}{+1.5}}) & 92.3\textcolor{gray}{{\scriptsize{$\pm$0.1}}} \\ 
     & U$^2$Net &  69.3\textcolor{gray}{{\scriptsize{$\pm$0.1}}} (\textbf{\textcolor{ForestGreen}{+1.6}}) & 88.6\textcolor{gray}{{\scriptsize{$\pm$0.1}}} & 76.3\textcolor{gray}{{\scriptsize{$\pm$0.1}}} (\textbf{\textcolor{ForestGreen}{+1.6}}) & 92.3\textcolor{gray}{{\scriptsize{$\pm$0.1}}} \\ 
  \multirow{2}{*}{+ \bgswaps} & DeepUSPS$^2$  & 69.5\textcolor{gray}{{\scriptsize{$\pm$0.1}}} (\textbf{\textcolor{ForestGreen}{+1.8}}) & 88.9\textcolor{gray}{{\scriptsize{$\pm$0.0}}} 
    & 76.6\textcolor{gray}{{\scriptsize{$\pm$0.1}}} (\textbf{\textcolor{ForestGreen}{+1.9}}) & 92.4\textcolor{gray}{{\scriptsize{$\pm$0.1}}} \\
     & U$^2$Net &  69.7\textcolor{gray}{{\scriptsize{$\pm$0.1}}} (\textbf{\textcolor{ForestGreen}{+2.0}}) & 88.9\textcolor{gray}{{\scriptsize{$\pm$0.0}}} 
    & 76.8\textcolor{gray}{{\scriptsize{$\pm$0.1}}} (\textbf{\textcolor{ForestGreen}{+2.1}}) & 92.3\textcolor{gray}{{\scriptsize{$\pm$0.1}}} \\
    \multicolumn{6}{@{}c@{}}{\makebox[\linewidth]{\dashrule[gray]}} \\
    BYOL (300)  & \na  & 72.7 & 90.9 & 79.6 & 94.0 \\
    \multirow{2}{*}{+ \bgrm} & DeepUSPS$^2$ & 73.3\textcolor{gray}{{\scriptsize{$\pm$0.2}}} (\textbf{\textcolor{ForestGreen}{+0.6}}) & 91.1\textcolor{gray}{{\scriptsize{$\pm$0.1}}} 
    & 80.4\textcolor{gray}{{\scriptsize{$\pm$0.3}}} (\textbf{\textcolor{ForestGreen}{+0.8}}) & 94.3\textcolor{gray}{{\scriptsize{$\pm$0.1}}} \\
     & U$^2$Net & 73.5\textcolor{gray}{{\scriptsize{$\pm$0.1}}} (\textbf{\textcolor{ForestGreen}{+0.8}}) & 91.2\textcolor{gray}{{\scriptsize{$\pm$0.1}}}
    & 80.5\textcolor{gray}{{\scriptsize{$\pm$0.1}}} (\textbf{\textcolor{ForestGreen}{+0.9}}) & 94.4\textcolor{gray}{{\scriptsize{$\pm$0.0}}} \\
    \multirow{2}{*}{+ \bgrand} & DeepUSPS$^2$  & 73.9\textcolor{gray}{{\scriptsize{$\pm$0.1}}} (\textbf{\textcolor{ForestGreen}{+1.2}}) & 91.6\textcolor{gray}{{\scriptsize{$\pm$0.0}}}
    & 81.0\textcolor{gray}{{\scriptsize{$\pm$0.0}}} (\textbf{\textcolor{ForestGreen}{+1.4}}) & 94.6\textcolor{gray}{{\scriptsize{$\pm$0.0}}} \\
     & U$^2$Net & 73.8{\textcolor{gray}{\scriptsize{$\pm$0.0}}} (\textbf{\textcolor{ForestGreen}{+1.1}}) & 91.7\textcolor{gray}{{\scriptsize{$\pm$0.0}}} 
    & 81.0\textcolor{gray}{{\scriptsize{$\pm$0.1}}} (\textbf{\textcolor{ForestGreen}{+1.4}}) & 94.7\textcolor{gray}{{\scriptsize{$\pm$0.0}}} \\
        \multicolumn{6}{@{}c@{}}{\makebox[\linewidth]{\dashrule[gray]}} \\
    SwAV (100)  & \na  & 72.2 & 91.0 & 79.1 & 94.0 \\
    \multirow{2}{*}{+ \bgrm} & DeepUSPS$^2$  & 73.6\textcolor{gray}{{\scriptsize{$\pm$0.1}}} (\textbf{\textcolor{ForestGreen}{+1.4}}) & 91.6\textcolor{gray}{{\scriptsize{$\pm$0.0}}} 
    & 80.6\textcolor{gray}{{\scriptsize{$\pm$0.1}}} (\textbf{\textcolor{ForestGreen}{+1.5}}) & 94.6\textcolor{gray}{{\scriptsize{$\pm$0.0}}} \\
     & U$^2$Net &  73.7\textcolor{gray}{{\scriptsize{$\pm$0.1}}} (\textbf{\textcolor{ForestGreen}{+1.5}}) & 91.6\textcolor{gray}{{\scriptsize{$\pm$0.0}}} 
    & 80.7\textcolor{gray}{{\scriptsize{$\pm$0.1}}} (\textbf{\textcolor{ForestGreen}{+1.6}}) & 94.6\textcolor{gray}{{\scriptsize{$\pm$0.0}}} \\
    \multirow{2}{*}{+ \bgrand} & DeepUSPS$^2$  & 73.4\textcolor{gray}{{\scriptsize{$\pm$0.0}}} (\textbf{\textcolor{ForestGreen}{+1.2}}) & 91.6\textcolor{gray}{{\scriptsize{$\pm$0.0}}} 
    & 80.4\textcolor{gray}{{\scriptsize{$\pm$0.1}}} (\textbf{\textcolor{ForestGreen}{+1.3}}) & 94.6\textcolor{gray}{{\scriptsize{$\pm$0.0}}} \\
     & U$^2$Net & 73.5\textcolor{gray}{{\scriptsize{$\pm$0.1}}} (\textbf{\textcolor{ForestGreen}{+1.3}}) & 91.6\textcolor{gray}{{\scriptsize{$\pm$0.0}}} 
    & 80.5\textcolor{gray}{{\scriptsize{$\pm$0.1}}} (\textbf{\textcolor{ForestGreen}{+1.4}}) & 94.6\textcolor{gray}{{\scriptsize{$\pm$0.0}}} \\
    \midrule
    \multicolumn{1}{@{}l}{\textit{Longer Training}}\\
    \rule{0pt}{2.5ex}\moco~(800)  & \na & 71.0 & 90.3 & 78.0 & 93.4 \\  
    \multirow{2}{*}{+ \bgrm} & DeepUSPS$^2$  & 71.9 (\textbf{\textcolor{ForestGreen}{+0.9}}) & 90.4 & 78.9 (\textbf{\textcolor{ForestGreen}{+0.9}}) & 93.5 \\
     & U$^2$Net &   72.0 (\textbf{\textcolor{ForestGreen}{+1.0}}) & 90.4 & 79.0 (\textbf{\textcolor{ForestGreen}{+1.0}}) & 93.6 \\ 
  \multirow{2}{*}{+ \bgswaps} & DeepUSPS$^2$ 
     & 72.2 (\textbf{\textcolor{ForestGreen}{+1.2}}) & 90.4 & 79.2 (\textbf{\textcolor{ForestGreen}{+1.2}}) & 93.6 \\ 
     & U$^2$Net 
    &  72.2 (\textbf{\textcolor{ForestGreen}{+1.2}}) & 90.5 & 79.0 (\textbf{\textcolor{ForestGreen}{+1.0}}) & 93.5 \\
    \multicolumn{6}{@{}c@{}}{\makebox[\linewidth]{\dashrule[gray]}} \\
    BYOL (1000)  & \na  & 73.8  & 91.5 & 80.5 &  94.3\\ 
    \multirow{2}{*}{+ \bgrm} & DeepUSPS$^2$  & 74.6 (\textbf{\textcolor{ForestGreen}{+0.8}}) & 91.8 & 81.3 (\textbf{\textcolor{ForestGreen}{+0.8}}) & 94.7 \\ 
     & U$^2$Net &  74.7 (\textbf{\textcolor{ForestGreen}{+0.9}}) & 91.9 & 81.5 (\textbf{\textcolor{ForestGreen}{+1.0}}) & 94.7 \\ 
    \multirow{2}{*}{+ \bgrand} & DeepUSPS$^2$  & 74.8 (\textbf{\textcolor{ForestGreen}{+1.0}}) & 92.0 & 81.7 (\textbf{\textcolor{ForestGreen}{+1.2}}) & 94.8 \\ 
     & U$^2$Net &  74.8 (\textbf{\textcolor{ForestGreen}{+1.0}}) & 92.1 & 81.6 (\textbf{\textcolor{ForestGreen}{+1.1}}) & 94.8 \\
     \multicolumn{6}{@{}c@{}}{\makebox[\linewidth]{\dashrule[gray]}} \\
    SwAV (800)  & \na & 74.9 & 92.1 & 81.4 & 95.1 \\ 
    \multirow{2}{*}{+ \bgrm} & DeepUSPS$^2$  & 76.1 (\textbf{\textcolor{ForestGreen}{+1.2}}) & 92.8 & 82.5 (\textbf{\textcolor{ForestGreen}{+1.1}}) & 95.4 \\
     & U$^2$Net &  76.2 (\textbf{\textcolor{ForestGreen}{+1.3}}) & 92.8 & 82.6 (\textbf{\textcolor{ForestGreen}{+1.2}}) & 95.4 \\ 
    \multirow{2}{*}{+ \bgrand} & DeepUSPS$^2$  & 76.1 (\textbf{\textcolor{ForestGreen}{+1.2}}) & 92.9 & 82.6 (\textbf{\textcolor{ForestGreen}{+1.2}}) & 95.5 \\
     & U$^2$Net &  76.0 (\textbf{\textcolor{ForestGreen}{+1.1}}) & 92.9 & 82.6 (\textbf{\textcolor{ForestGreen}{+1.2}}) & 95.5 \\
    \bottomrule
    \end{tabular}
    }
    \caption{
    \textbf{Ablating the impact of the saliency method used for foreground extraction.}
    We find nearly identical performance when we use U$^2$Net, a state-of-the art saliency detector that is trained with supervision. Foreground extraction using higher quality masks results in slightly better performance when trained for fewer epochs, but this benefit disappears with longer training. All numbers are based on our implementation. Notation: \moco~(800) indicates that \moco~was trained for 800 epochs.
    }
    \label{tab:u2net_vs_deepusps_ablation}
\end{table}

One important limitation of current SSL methods is the long training required for competitive performance, typically 800-1000 epochs, in contrast with supervised learning. Our results in Figure \ref{tab:swav_wider_and_warmup} (left) show that background augmentations enable a step forward in reducing the amount of training required for competitive performance in SSL. In these results, aside from diagnosing and fixing early optimization issues, we simply used the default settings for \swav. However, there remains much room for improvement in conjunction with background augmentations. We briefly explore one such improvement here.

Recall that \swav~uses \texttt{multi-crop} augmentation, where local crops covering small parts of the image are expected to be predictive of global crops. Here, we increase the area that the small crops may cover of the full image \footnote{Since we maintain the same resolution of 96$\times$96 for the smaller crops as in the default setting and simply modify the max scale in \texttt{RandomResizedCrop}, compute and memory requirements stay identical. Additional details in Appendix \ref{app:ablations}.}. While the small crops may feature more of the background with this change, background augmentations already prevent excessive focus on the background. This simple change improves the performance of \bgrm~(\bgrand)~from 74.1\% to 74.4\% (74.2\%). In only 100 epochs, performance exceeds many recent high performing SSL methods trained for 800-1000 epochs, e.g. Barlow Twins (73.2\%, 1000 epochs), MoCo-v3 \citep{chen_empirical_2021} (73.8\%, 800 epochs) and BYOL (74.3\%, 1000 epochs). In contrast, with the same change, the \swav~baseline fails to train and the loss at the end of pre-training is at chance. Note that our default setting for SwAV does not include the modifications discussed in this section unless otherwise indicated.

\subsection{What is the Impact of Mask Quality?}
\label{sec:u2net_mask_imnet_lin_eval}

While DeepUSPS$^2$ is better than or on par with weakly supervised saliency methods and even some recent supervised saliency methods, state-of-the-art supervised saliency methods like U$^2$Net achieve better performance on saliency benchmarks. We perform an ablation using foreground masks generated by U$^2$Net for background augmentations. While the resulting models are not truly self supervised, they can nevertheless help us understand if using better foreground masks can lead to larger performance improvements. We report the results of these experiments in Table \ref{tab:u2net_vs_deepusps_ablation}, finding that performance is nearly identical whether DeepUSPS$^2$ or U$^2$Net are used to extract foreground masks. Using higher quality masks leads to slightly better performance when trained for fewer epochs but this gap disappears with longer training. In later sections, we evaluate both sets of models on a range of downstream tasks to gain further insight. 

While these results suggest that there may be diminishing gains to using higher quality masks, some natural questions arise, e.g. which SSL methods and background augmentations are more robust to mask quality? How does performance vary as a function of mask quality? We systematically perturb mask quality in numerous ways (via mask rotation, shearing, translation, flips and replacing masks with bounding-box masks) to answer these questions in Appendix \ref{app:noisy_masks}. Overall, we find that there is substantial robustness to mask quality. Of the SSL methods and background augmentations considered, we find that \swav~and \bgswaps~are particularly robust.

\subsection{Limited-Label Setting}

While linear evaluation using 100\% of ImageNet labels is a standard evaluation metric, it is also somewhat impractical due to the large amount of labels involved - after all, one of the more important goals of SSL is good performance when labeled data is highly limited. Linear evaluation in limited label settings reveals a large improvement in performance from background augmentations. For 1\% and 10\% labels, we use the same fixed splits of ImageNet labeled training data as in \citet{chen2020big}. We similarly find large performance benefits in semi-supervised evaluation (fine-tuning the pre-trained backbone in addition to learning a linear classifier). We report Top-1 and Top-5 accuracies in Table \ref{tab:lim_lab_deepusps}.

\begin{table}
    \centering
    \begin{tabular}{llcccc}\toprule
    \multicolumn{2}{l}{Method}  &     \multicolumn{2}{c}{1\% Labels} & \multicolumn{2}{c}{10\% Labels}\\
           \cmidrule(lr){3-4}  \cmidrule(lr){5-6} \\
           && Top-1 & Top-5 & Top-1 & Top-5 \\ \midrule
    \multicolumn{2}{l}{\textcolor{gray}{Supervised \footnotesize{\citep{zhai_s4l_2019}}}} 
    & \textcolor{gray}{25.4} & \textcolor{gray}{48.4} & \textcolor{gray}{56.4} &  \textcolor{gray}{80.4} \\
    \midrule
    \parbox[t]{3mm}{\multirow{9}{*}{\rotatebox[origin=c]{90}{Linear}}}
    &\moco~{\scriptsize{(\textit{repro.})}}              & 52.0 & 77.7 & 63.9 & 85.8 \\
    &\moco~+ \bgrm      & 54.1 (\textbf{\textcolor{ForestGreen}{+2.1}}) & 78.6 & 65.1 (\textbf{\textcolor{ForestGreen}{+1.2}}) & 86.2 \\
    &\moco~+ \bgswaps   & 56.0 (\textbf{\textcolor{ForestGreen}{+4.0}}) & 79.5 & 65.9 (\textbf{\textcolor{ForestGreen}{+2.0}}) & 86.4 \\
    \vspace{-0.2mm}
    &\multicolumn{5}{@{}c@{}}{\makebox[0.76\linewidth]{\dashrule[gray]}} \\
    \vspace{-0.2mm}
    &BYOL {\scriptsize{(\textit{repro.})}}                & 57.5 & 80.8 & 68.6 & 88.6 \\
    &BYOL + \bgrm        & 60.1 (\textbf{\textcolor{ForestGreen}{+2.6}}) & 82.7 & 70.1 (\textbf{\textcolor{ForestGreen}{+1.5}})  & 89.2 \\
    &BYOL + \bgrand      & \textbf{60.9} (\textbf{\textcolor{ForestGreen}{+3.4}}) & \textbf{83.3} & 70.4 (\textbf{\textcolor{ForestGreen}{+1.8}}) & 89.5 \\
    \vspace{-0.2mm}
    &\multicolumn{5}{@{}c@{}}{\makebox[0.76\linewidth]{\dashrule[gray]}} \\
    \vspace{-0.2mm}
    &SwAV {\scriptsize{(\textit{repro.})}}                & 52.8 & 78.4 & 68.3 & 88.7 \\
    &SwAV + \bgrm        & 57.0 (\textbf{\textcolor{ForestGreen}{+4.2}}) & 81.3 & 70.4 (\textbf{\textcolor{ForestGreen}{+2.1}}) & 89.8 \\
    &SwAV + \bgrand      & 56.4 (\textbf{\textcolor{ForestGreen}{+3.6}}) & 81.1 & 70.2 (\textbf{\textcolor{ForestGreen}{+1.9}}) & 89.7 \\
    \midrule
    \parbox[t]{3mm}{\multirow{14}{*}{\rotatebox[origin=c]{90}{Finetune}}}
    &\moco~{\scriptsize{(\textit{repro.})}}              & 54.1 & 81.3 & 67.6 & 89.4 \\
    &\moco~+ \bgrm      & 55.2 (\textbf{\textcolor{ForestGreen}{+1.1}}) & 81.3 & 67.8 (\textbf{\textcolor{ForestGreen}{+0.2}}) & 89.2 \\
    &\moco~+ \bgswaps   & 57.3 (\textbf{\textcolor{ForestGreen}{+3.2}}) & 82.4 & 68.7 (\textbf{\textcolor{ForestGreen}{+1.1}}) & 89.5 \\
    \vspace{-0.2mm}
    &\multicolumn{5}{@{}c@{}}{\makebox[0.76\linewidth]{\dashrule[gray]}} \\
    \vspace{-0.2mm}
    &BYOL {\scriptsize{(\textit{repro.})}}                & 57.3 & 80.5 & 70.6 & 90.0 \\
    &BYOL + \bgrm        & 59.9 (\textbf{\textcolor{ForestGreen}{+2.6}}) & 82.4 & \underline{71.7} (\textbf{\textcolor{ForestGreen}{+1.1}}) & \underline{90.5} \\
    &BYOL + \bgrand      & \underline{60.7} (\textbf{\textcolor{ForestGreen}{+3.4}}) & \underline{82.8} & \textbf{72.0} (\textbf{\textcolor{ForestGreen}{+1.4}}) & \textbf{90.7} \\
    \vspace{-0.2mm}
    &\multicolumn{5}{@{}c@{}}{\makebox[0.76\linewidth]{\dashrule[gray]}} \\
    \vspace{-0.2mm}
    &SwAV {\scriptsize{(\textit{repro.})}}                & 54.0 & 78.5 & 70.1 & 89.9 \\
    &SwAV + \bgrm        & 55.2 (\textbf{\textcolor{ForestGreen}{+1.2}}) & 79.4 & 70.8 (\textbf{\textcolor{ForestGreen}{+0.7}}) & 90.2 \\
    &SwAV + \bgrand      & 55.9 (\textbf{\textcolor{ForestGreen}{+1.9}}) & 79.4 & 71.1 (\textbf{\textcolor{ForestGreen}{+1.0}}) & 90.4 \\
    \cmidrule(lr){2-6}
    & \multicolumn{5}{c}{\small{Published Baselines}} \\
    \cmidrule(lr){2-6}
    &PIRL               & - & 57.2 & - & 83.8 \\
    & SimCLR            & 48.3 & 75.5 & 65.6 & 87.8 \\
    & SwAV              & 53.9 & 78.5 & 70.2 & 89.9 \\
    & BYOL              & 53.2 & 78.4 & 68.8 & 89.0 \\
    & Barlow Twins      & 55.0 & 79.2 & 69.7 & 89.3\\
    \bottomrule
    \end{tabular}
    \caption{\textbf{Limited-Labels Setting.} Background augmentations improve performance in the limited-labels  setting. Linear evaluation using 100\% of ImageNet labels though a standard benchmark, is a somewhat unrealistic setting. Evaluation in the more practical setting of limited-labels reveals even larger improvement in performance. We highlight \textbf{\textcolor{ForestGreen}{performance gains}} due to background augmentations. Best (second best) results are in \textbf{bold} (\underline{underlined}).
    }
    \label{tab:lim_lab_deepusps}
\end{table}

Our first key finding is that the improvement in performance in limited label settings, for both linear and semi-supervised evaluation, is substantially larger than in 100\% linear evaluation, with improvements up to 4.2\%. Large gains in linear evaluation especially reflect a much better learned representation, since the backbone is frozen. Our second key finding is that \bgswaps~is especially effective in limited label settings. Indeed, in the 1\% setting, the gain from \bgswaps~is nearly $3\times$ the gain from \bgrm~in semi-supervised evaluation and $\sim2\times$ that of \bgrm~in linear evaluation, demonstrating the effectiveness of using negatives matched to the query's background.

Our third finding is that it is generally better to use \bgrand~or \bgswaps~over \bgrm, consistent with our previous results. Our findings here set new, stronger baselines: 60.9\% Top-1 in the 1\% labels setting and 72\% Top-1 in the 10\% labels setting. It is worth noting that $\sim$71\% is the linear evaluation baseline for \moco~ using 100\% of the labels. Note that our reproduction of BYOL's performance in limited label settings already improves upon the published baseline (by +4.1\%, +1.8\% in the 1\% and 10\% labels settings respectively) by adopting a much smaller learning rate for the pre-trained backbone than the classifier head---background augmentations \textit{further} improve on these stronger baselines.

Finally, we note that nearly identical findings hold when we instead use U$^2$Net for foreground extraction, see Table \ref{app:lim_lab_u2net}. All models receive full pre-training.

\subsection{Can Background Augmentations Improve Performance in the Supervised Setting?}

We have found that background augmentations provide a significant performance boost to a suite of high-performing SSL methods, and shrink the gap to the supervised baseline down to $0.3\%$. We note that most SSL methods utilize an augmentation suite that is inherited from supervised training. By designing augmentations specifically for SSL, we were able to induce a substantial increase in performance; this raises the question of whether a similar performance boost would be observed when applying background augmentations to supervised training.

\begin{table}
    \centering
    \begin{tabular}{l|cccc}\toprule
        baseline ($p=0.0$): 76.4 & $p=0.1$ & $p=0.2$ & $p=0.3$ & $p=0.5$\\\midrule
        \rowcolor{lightgray}
        \bgrm & 76.6 & 76.5 & 76.4 & 75.9 \\
        \bgrm~+ \small{retrain classifier} & 76.5 & 76.3 & 76.0 & 75.1 \\ 
        \bgrm~+ \small{finetune}  & 76.5 & 76.6 & 76.5 & 76.0 \\ \midrule
        \rowcolor{lightgray}
        \bgrand & 76.4 & 76.6 & 76.6 & 76.0 \\ 
        \bgrand~+ \small{retrain classifier}  & 76.4 & 76.2 & 75.9 & 73.5 \\
        \bgrand~+ \small{finetune}  & 76.6 & 76.6 & 76.6 & 76.4 \\ \bottomrule
    \end{tabular}
    \caption{\textbf{Supervised Setting:} Background augmentations do not improve over the baseline (76.4 in our setting).}
    \label{tab:supervised}
\end{table}

Interestingly, we find that background augmentations \emph{do not} confer a performance benefit in the supervised setting. In Table \ref{tab:supervised}, we report the performance of \bgrm~and \bgrand, sweeping over $p$, finding no setting that outperforms the supervised benchmark\footnote{Note that \bgswaps~is not applicable here since there is no concept of a negative to match.}.

\begin{table}
    \centering
    \begin{tabular}{lcc}\toprule
    Method & Epochs & Accuracy \\\midrule
    Supervised baseline & 90 & 76.4 \\ \midrule
    Supervised + \bgrm & 90 & 76.5 \\
    \rowcolor{lightgray}
    Supervised + \bgrm & 300 & 76.5 \\
    Supervised + \bgrand & 90 & 76.6 \\
    \rowcolor{lightgray}
    Supervised + \bgrand & 300 & 76.6 \\ \bottomrule
    \end{tabular}
    \caption{\textbf{Supervised Setting: Longer Training.} Longer training with background augmentation does not significantly improve performance in the supervised setting over the baseline. Augmentation strength: $p=0.2$.}
    \label{tab:supervised_longtrain}
\end{table}

One may wonder if this lack of improvement is an artifact of the evaluation protocol, which is  different from the SSL setting, where evaluation is either by training a linear classifier on top of the frozen trunk or by fine-tuning the whole network (trunk + linear classifier) \textit{without background augmentations}. We therefore, \textit{a}) re-train a linear classifier without background augmentations on top of the frozen trunk (of the supervised network trained with background augmentations) and (separately) \textit{b}) fine-tune the whole network without background augmentations, once again finding no performance benefit.

In the supervised setting, strong augmentations may require much longer training to be effective (e.g. as in the case of CutMix \citep{yun_cutmix_2019}). To account for a similar possibility in the case of background augmentations, we include background augmentations in supervised training and follow a much longer training schedule (see Appendix \ref{app:training_details} for details) for 300 epochs (following CutMix) and find no significant performance benefits, see Table \ref{tab:supervised_longtrain}.

\subsection{So, \textit{When} do Background Augmentations Help?}
\label{sec:rotnet}

Our results in the previous section suggest that the utility of background augmentations in SSL does not generalize to the supervised setting. Given the importance of augmentations to SSL (e.g., \citet{chen2020simple}), these results highlight the need to evaluate and explore augmentations tailor-made for the SSL setting and are consistent with
similar findings \citep{chen2020simple} for color distortion and blur augmentations. While the test bed of high performing SSL methods we have considered thus far is diverse, they share a commonality: they all use Siamese networks to compare or contrast views of images, raising the natural question of whether this is the only SSL setting where background augmentations confer an advantage.  

\begin{table}
    \centering
    \begin{tabular}{l|cccc}\toprule
        baseline $(p=0.0)$: 36.1 & $p=0.1$ & $p=0.2$ & $p=0.3$ & $p=0.5$\\\midrule
        \bgrm  & 36.0 & 35.5 & 35.5 & 34.8 \\
        \bgrand & 36.1 & 36.0 & 35.9 & 35.7 \\\bottomrule
    \end{tabular}
    \caption{\textbf{RotNet:} Background augmentations do not improve over the baseline.}
    \label{tab:rotnet}
\end{table}

To investigate this question, we turn to RotNet \citep{gidaris2018unsupervised}---a simple, yet surprisingly effective SSL method that is not based on a Siamese architecture nor on comparisons between images. Training a RotNet involves augmenting the data with rotated images and training a network to categorize the orientation of an image, thereby forcing the network to learn a meaningful representation to accomplish this task. We implemented background augmentations in RotNet, i.e. we either perform \bgrm~or~\bgrand~followed by rotating the image (and training the network to classify the orientation). Interestingly, we found that background augmentations confer no performance benefits, see Table \ref{tab:rotnet}. Here, \bgrand~and \bgrm~decorrelate the foreground and the background, while \bgrm~additionally reduces the incentive to encode background information, since a grayscale background is not informative for the pretext task of categorizing image orientation. Thus, merely decorrelating the foreground and background or disincentivizing focus on the background are not sufficient to improve semantic focus.

Based on our findings, we speculate that background augmentations are most helpful when there is a similarity comparison between images, and can help prevent the model from using the background as a shortcut to place images nearby (or far away) in embedding space which can hinder learning about the semantic content present in an image.

\begin{figure}
    \centering
    \includegraphics[width=0.75\textwidth]{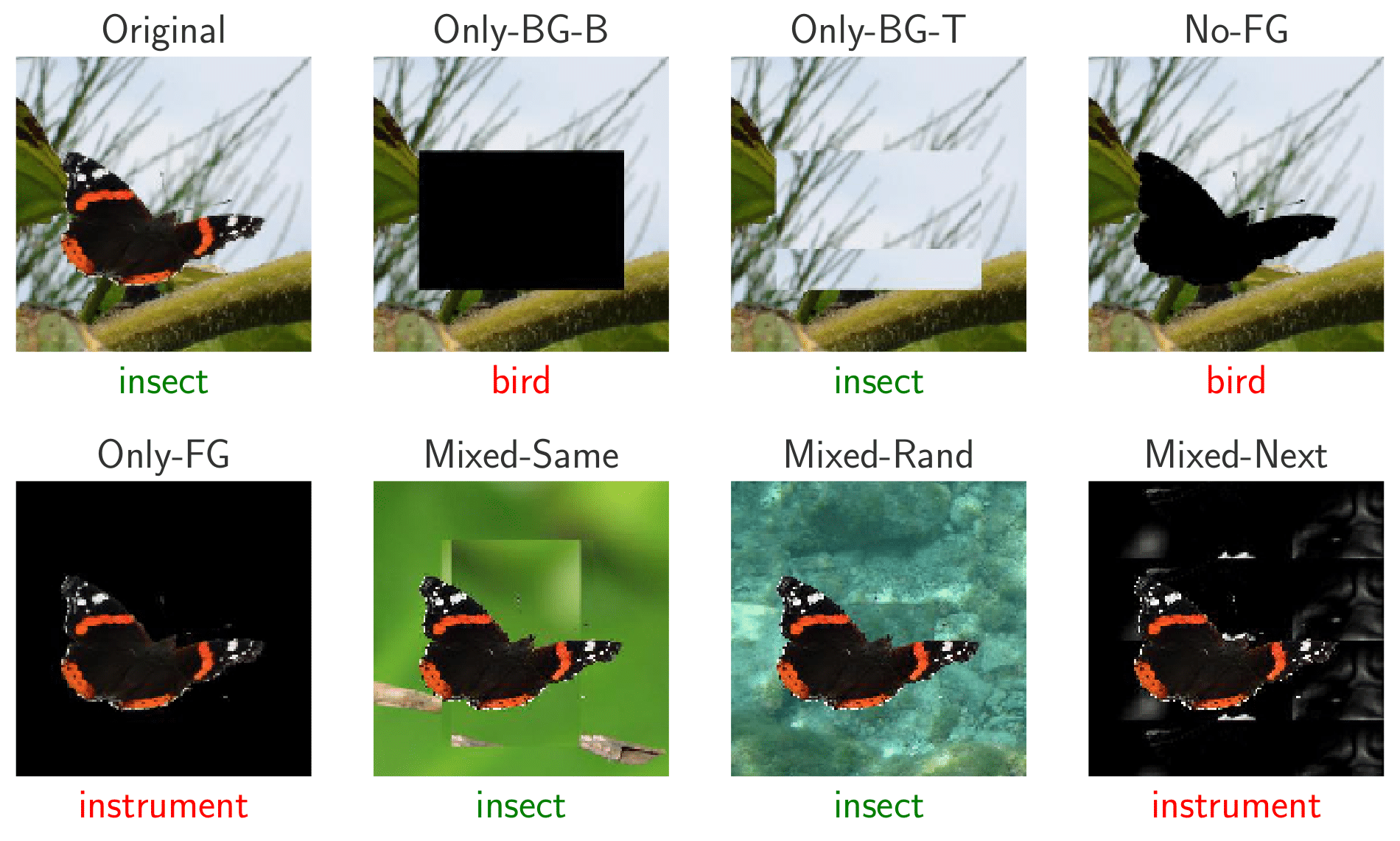}
    \caption{\textbf{Examples of variations of ImageNet-9.} Also shown are classification decisions from the supervised baseline. Image from \citet{xiao2020noise}. }
    \label{fig:bg_chall_example}
\end{figure}

\section{Generality of Representations Induced by Background Augmentations}
If background augmentations lead to increased focus on semantic content and decreased focus on non-robust predictors for classification (e.g., \citet{ilyas_neurips_2020_features_not_bugs}), we expect that these augmentations would also lead to improved performance on out-of-distribution downstream tasks. In particular, we expect gains on those tasks which have proven especially challenging for supervised networks. Here, we discuss several such tasks, including ImageNet-9 \citep{xiao2020noise}, adversarial attacks \citep{goodfellow_fgsm, kurakin_adversarial_2016, madry2018towards}, natural adversarial examples \citep{hendrycks2019nae}, ImageNet-Renditions \citep{hendrycks_many_2021} and ReaL ImageNet \citep{beyer2020imagenetreal}, finding improved performance across the board.

\subsection{Improved Robustness to Shift in Foreground-Background Statistics}
\label{sec:bg_challenge}

ImageNet-9 (IN-9), introduced in \citet{xiao2020noise}, consists of out-of-distribution data sets that are different variations of a 9-class subset of ImageNet. The variants are designed to have different amounts of foreground and background signal, see Figure \ref{fig:bg_chall_example} for examples. In the Only-BG-B and Only-BG-T variants, the foreground is removed and replaced either with a black box (Only-BG-B) or a tiled version of the background (Only-BG-T); No-FG features images with the foreground shape cut out (and discernible), while Only-FG features the foreground alone on a black background (similar to our \bgrm); Mixed-Same, Mixed-Rand, and Mixed-Next, feature foregrounds pasted onto backgrounds from different images of the same class (Mixed-Same), random images (Mixed-Rand), and deterministically from the next class such that backgrounds provide systematically misleading information (Mixed-Next). If models learn to focus on the semantically meaningful foreground and ignore the background, we should expect classification performance to decrease for Only-BG-B and Only-BG-T, and to increase for Only-FG, Mixed-Same, Mixed-Rand, and Mixed-Next\footnote{It is more difficult to determine whether performance should increase or decrease for the No-FG variant, since this manipulation leaves a perfectly shaped cutout of the foreground on the background, which provides substantial information about the structure of the foreground even though it has been removed.}. 

\begin{table}
    \centering
    \resizebox{1\columnwidth}{!}{
    \begin{tabular}{lcccccccccc}\toprule
Data Set  & Supervised & \multicolumn{3}{c}{\textbf{\moco}}  & \multicolumn{3}{c}{\textbf{BYOL}} & \multicolumn{3}{c}{\textbf{SwAV}}
    \\\cmidrule(lr){3-5} \cmidrule(lr){6-8} \cmidrule(lr){9-11} 
     &      & baseline & \bgrm & \bgswaps & baseline & \bgrm & \bgrand & baseline & \bgrm & \bgrand \\ \midrule
Original    & 95.6 &  92.7 & 93.8  & 94.2  & 94.9  & 95.6  &   96.0  &   94.1 & 95.0 & 94.9 \\  
Only-BG-B $\downarrow$  & 11.4 &  6.1 &  6.1   & 3.6   & 5.4   & 4.9   & 6.0  &   10.9 & 8.8  & 8.3  \\
Only-BG-T $\downarrow$  & 16.3 & 14.8 &  12.9  & 9.3  & 12.7  & 11.8   &  11.5 &  15.8 & 16.7 & 17.6 \\
No-FG      & 45.9 & 37.8 &  42.3  & 39.6  & 43.9  & 45.9   &   46.2 &    41.3 & 44.2 & 45.2 \\
\rowcolor{lightgray}
Only-FG $\uparrow$     & 86.8 & 74.4 &  81.9 (\textbf{\textcolor{ForestGreen}{+7.5}})  & 86.1 (\textbf{\textcolor{ForestGreen}{+11.7}})  & 83.5  & 88.8 (\textbf{\textcolor{ForestGreen}{+5.3}})  &   87.7  (\textbf{\textcolor{ForestGreen}{+4.2}}) &    79.4 & 85.3  (\textbf{\textcolor{ForestGreen}{+5.9}}) & 84.3  (\textbf{\textcolor{ForestGreen}{+4.9}}) \\
\rowcolor{lightgray}
Mixed-Same $\uparrow$ & 86.2 & 81.8 &  84.0 (\textbf{\textcolor{ForestGreen}{+2.2}})  & 87.9 (\textbf{\textcolor{ForestGreen}{+6.1}})  & 86.2  & 88.6 (\textbf{\textcolor{ForestGreen}{+2.4}})   &   90.1 (\textbf{\textcolor{ForestGreen}{+3.9}}) &    82.2 & 86.1 (\textbf{\textcolor{ForestGreen}{+3.9}}) & 86.3 (\textbf{\textcolor{ForestGreen}{+4.1}}) \\
\rowcolor{lightgray}
Mixed-Rand $\uparrow$ & 78.9 & 70.7 &  76.3 (\textbf{\textcolor{ForestGreen}{+5.6}}) & 84.1 (\textbf{\textcolor{ForestGreen}{+13.4}})  & 79.6  & 83.2 (\textbf{\textcolor{ForestGreen}{+3.6}})   &   85.5 (\textbf{\textcolor{ForestGreen}{+5.9}}) &    71.3 & 77.1 (\textbf{\textcolor{ForestGreen}{+5.8}}) & 77.0 (\textbf{\textcolor{ForestGreen}{+5.7}}) \\
\rowcolor{lightgray}
Mixed-Next $\uparrow$ & 77.2 & 67.0 &  73.0 (\textbf{\textcolor{ForestGreen}{+6.0}})  & 82.2 (\textbf{\textcolor{ForestGreen}{+15.2}})  & 77.6  & 80.7 (\textbf{\textcolor{ForestGreen}{+3.1}})   &   84.0 (\textbf{\textcolor{ForestGreen}{+6.4}}) &    69.0 & 74.3 (\textbf{\textcolor{ForestGreen}{+5.3}}) & 74.4 (\textbf{\textcolor{ForestGreen}{+5.4}}) \\\bottomrule
    \end{tabular}
    }
    \caption{\textbf{Robustness: Foreground-Background Shifts.} Background augmentations result in large performance gains on ImageNet-9 across all SSL methods, with \bgswaps~generally exhibiting similar or better performance than \bgrm. We highlight the performance benefit on the variants of ImageNet-9 especially relevant to our work. All accuracies reported for background augmented SSL methods are averages of 3 independent runs (we exclude SEM to avoid clutter, see Table \ref{tab:background_challenge_deepusps_expand} for an expanded table that includes SEM). All pre-training durations correspond to respective medium settings. Note that ImageNet-9 uses only 9 classes, so chance is $\sim$11.1\%.}
    \label{tab:background_challenge_deepusps}
\end{table}

We evaluate the baseline SSL methods as well as models with background augmentations on all variants of IN-9 in Table \ref{tab:background_challenge_deepusps}. As in the supervised setting  \citep[see][]{xiao2020noise}, we found that models which perform better on the Original IN-9 also perform better across other IN-9 variants. Critically, we also found that background augmentations consistently improved performance on IN-9, especially on the images with misleading backgrounds (Mixed-X), and in some cases, enable outperforming the supervised baseline. We also found that \bgswaps~consistently improved performance over \bgrm. For example, on Mixed-Next, the \moco~baseline has an accuracy of 67.0\%, worse than the supervised baseline's performance of 77.2\%, but incorporating \bgrm~and \bgswaps~increases this to 73.0\% and 82.2\%, respectively. These results demonstrate that background augmentations do indeed encourage semantic focus on the foreground, and that explicitly discouraging background focus (as in \bgswaps) is beneficial over simply removing positive signal in the background. We also note that \bgrand~generally confers larger improvements over \bgrm.

\begin{table}
    \centering
    \resizebox{1\columnwidth}{!}{
    \begin{tabular}{cccccccccl}\toprule
    \shortstack{Pre-Train\\Duration} & \multicolumn{3}{c}{\textbf{\moco}}  & \multicolumn{3}{c}{\textbf{BYOL}} & \multicolumn{3}{c}{\textbf{SwAV}}
    \\\cmidrule(lr){2-4} \cmidrule(lr){5-7} \cmidrule(lr){8-10} 
 & baseline & \bgrm & \bgswaps & baseline & \bgrm & \bgrand & baseline & \bgrm & \bgrand \\ \midrule
Med. 
& 11.1 & 7.7 (\textbf{\textcolor{ForestGreen}{-3.4}})  & 3.8 (\textbf{\textcolor{ForestGreen}{-7.3}})
& 6.6 & 5.4 (\textbf{\textcolor{ForestGreen}{-1.2}})   & 4.6 (\textbf{\textcolor{ForestGreen}{-2.0}}) 
& 10.9 & 9.0 (\textbf{\textcolor{ForestGreen}{-1.9}})  & 9.3 (\textbf{\textcolor{ForestGreen}{-1.6}})  \\
Full 
& 10.0  & 6.8 (\textbf{\textcolor{ForestGreen}{-3.2}})  & 4.4 (\textbf{\textcolor{ForestGreen}{-5.6}})
& 9.1  &  5.3 (\textbf{\textcolor{ForestGreen}{-3.8}})  & 4.4 (\textbf{\textcolor{ForestGreen}{-4.7}})  
& 11.4 &  9.3 (\textbf{\textcolor{ForestGreen}{-2.1}})  & 9.0 (\textbf{\textcolor{ForestGreen}{-2.4}})  \\  
\bottomrule
    \end{tabular}
    }
    \caption{\textbf{BG-Gap:} Background augmentations decrease BG-Gap of SSL Methods.}
    \label{tab:bg_gap}
\end{table}

To quantify the impact of foreground-background correlations in the learned representations, we compute the \textit{BG-Gap} \citep{xiao2020noise} as the difference between accuracies in the Mixed-Same and Mixed-Rand settings and find that background augmentations decrease the BG-Gap in the SSL methods considered, relative to the baselines. For the baselines, we also find that the BG-Gap slightly increases when trained for longer (Table \ref{tab:bg_gap}) for BYOL and SwAV, while it slightly decreases for \moco. We speculate that this is due to the use of a large number ($|Q|=65536$) of negative instances in \moco---it is possible some of the negative instances have backgrounds similar to the query $q$, thereby implicitly discouraging background focus. As such, SSL models do not seem to learn much background invariance when trained for longer duration. When background augmentations are used, the BG-Gap is roughly the same for shorter or longer training duration---in other words, background augmentations do not require long training to be effective. Additional results: Appendix \ref{app:add_results} (Tables \ref{tab:background_challenge_deepusps_expand}, \ref{tab:background_challenge_deepusps_full}, \ref{tab:background_challenge_u2net}, \ref{tab:background_challenge_u2net_full}, \ref{tab:bg_gap_u2net}). 

\subsection{ReaL Imagenet Confirms Improvement of Semantic Focus}
Next, we evaluate performance using Reassessed Labels (ReaL, \citet{beyer2020imagenetreal}) for ImageNet, which relabel ImageNet to better represent the semantic content of the images. Using ReaL, \citet{beyer2020imagenetreal} found that the gains due to many recent methods were smaller than when the original labels are used. As with the original ImageNet labels, we found that background augmentations substantially improve performance on ImageNet ReaL (Table \ref{tab:gain_summary_imgnet_deepusps}), confirming that background augmentations do induce increased semantic focus rather than simply facilitating overfitting to the original ImageNet labels. In fact, the improvement on ReaL is \textit{slightly larger} when trained for fewer epochs.

\begin{table}
    \centering
    \resizebox{1\columnwidth}{!}{
    \begin{tabular}{cccccccccl}\toprule
  \shortstack{Pre-Train\\Duration}  & \multicolumn{3}{c}{\textbf{\moco}}  & \multicolumn{3}{c}{\textbf{BYOL}} & \multicolumn{3}{c}{\textbf{SwAV}}
    \\\cmidrule(lr){2-4} \cmidrule(lr){5-7} \cmidrule(lr){8-10} 
 & baseline & \bgrm & \bgswaps & baseline & \bgrm & \bgrand & baseline & \bgrm & \bgrand \\ \midrule
Med.
& 54.7 & 56.7\gsem{0.1} (\textbf{\textcolor{ForestGreen}{+2.0}})  & 57.2\gsem{0.1} (\textbf{\textcolor{ForestGreen}{+2.5}})
& 60.7 & 61.7\gsem{0.2} (\textbf{\textcolor{ForestGreen}{+1.0}})  & 62.1\gsem{0.1} (\textbf{\textcolor{ForestGreen}{+1.4}}) 
& 59.3 & 61.2\gsem{0.3} (\textbf{\textcolor{ForestGreen}{+1.9}})  & 60.7\gsem{0.0} (\textbf{\textcolor{ForestGreen}{+1.4}})  \\
Full
& 58.9 & 59.6 (\textbf{\textcolor{ForestGreen}{+0.7}})  & 60.3 (\textbf{\textcolor{ForestGreen}{+1.4}})
& 61.9 & 63.4 (\textbf{\textcolor{ForestGreen}{+1.5}})  & 62.8 (\textbf{\textcolor{ForestGreen}{+0.9}}) 
& 61.7 & \textbf{63.8} (\textbf{\textcolor{ForestGreen}{+2.1}}) & 63.4 (\textbf{\textcolor{ForestGreen}{+1.7}})  \\
\bottomrule
    \end{tabular}
    }
    \caption{\textbf{Robustness: Natural Distribution Shift.} Background  augmentations  improve  performance on ImageNet-v2, a test set for ImageNet. Notably, background augmentations enable \swav~to perform \textit{on par} with the standard supervised baseline (63.8\%).}
    \label{tab:imagenet_v2_deepusps_sq}
\end{table}

\subsection{Improvement on ImageNet-v2 and ObjectNet}
\label{sec:imnet_v2_and_objectnet}
We next evaluate performance on ImageNet-v2 \citep{imagenetv2} and ObjectNet \citep{objectnet}. ImageNet-v2 is a test set for ImageNet and can be considered a “natural" distribution shift setting. ObjectNet is a challenging test set where the object orientation, viewpoint and background are varied in a controlled manner. We find that background augmentations confer sizeable performance benefits in both of these settings, see Tables \ref{tab:imagenet_v2_deepusps_sq} and \ref{tab:objectnet_deepusps_sq}. 

Notably, on ImageNet-v2, background augmentations enable \swav~to perform \textit{on par} with the supervised baselines. Specifically, the torchvision ResNet50 baseline has an accuracy of 63.3\% on ImageNet-v2, while our re-implementation of the standard, stronger baseline \citep{goyal_accurate_2018} has an accuracy of 63.8\%. Additional results: Appendix \ref{app:add_results} (Tables \ref{tab:expanded_imnetv2_deepusps_sq}, \ref{tab:expanded_imnetv2_u2net}, \ref{tab:objectnet_u2net}).

\begin{table}
    \centering
    \resizebox{1\columnwidth}{!}{
    \begin{tabular}{cccccccccl}\toprule
  \shortstack{Pre-Train\\Duration}  & \multicolumn{3}{c}{\textbf{\moco}}  & \multicolumn{3}{c}{\textbf{BYOL}} & \multicolumn{3}{c}{\textbf{SwAV}}
    \\\cmidrule(lr){2-4} \cmidrule(lr){5-7} \cmidrule(lr){8-10} 
 & baseline & \bgrm & \bgswaps & baseline & \bgrm & \bgrand & baseline & \bgrm & \bgrand \\ \midrule
Med.
& 14.4 & 16.8\gsem{0.2} (\textbf{\textcolor{ForestGreen}{+2.4}})  & 18.2\gsem{0.1} (\textbf{\textcolor{ForestGreen}{+3.8}})
& 20.4 & 22.1\gsem{0.3} (\textbf{\textcolor{ForestGreen}{+1.7}})  & 22.3\gsem{0.1} (\textbf{\textcolor{ForestGreen}{+1.9}}) 
& 16.1 & 19.3\gsem{0.1} (\textbf{\textcolor{ForestGreen}{+3.2}})  & 18.1\gsem{0.1} (\textbf{\textcolor{ForestGreen}{+2.0}})  \\
Full
& 17.4 & 19.9 (\textbf{\textcolor{ForestGreen}{+2.5}})  & 20.8 (\textbf{\textcolor{ForestGreen}{+3.4}})
& 20.8 & 23.9 (\textbf{\textcolor{ForestGreen}{+3.1}})  & 23.4 (\textbf{\textcolor{ForestGreen}{+2.6}}) 
& 19.3 & 21.9 (\textbf{\textcolor{ForestGreen}{+2.6}}) & 21.3 (\textbf{\textcolor{ForestGreen}{+2.0}})  \\
\bottomrule
    \end{tabular}
    }
    \caption{\textbf{Robustness: Rotation, Viewpoint, Background Shift.} Background augmentations improve  performance on ObjectNet, a challenging test set that controls object orientation, viewpoint and background.}
    \label{tab:objectnet_deepusps_sq}
\end{table}

\subsection{Natural Adversarial Examples}
\label{sec:nat_adv_examples}

\begin{table}
    \centering
    \resizebox{1\columnwidth}{!}{
    \begin{tabular}{cccccccccl}\toprule
     \shortstack{Pre-Train\\Duration} & \multicolumn{3}{c}{\textbf{\moco}}  & \multicolumn{3}{c}{\textbf{BYOL}} & \multicolumn{3}{c}{\textbf{SwAV}}
    \\\cmidrule(lr){2-4} \cmidrule(lr){5-7} \cmidrule(lr){8-10} 
 & baseline & \bgrm & \bgswaps & baseline & \bgrm & \bgrand & baseline & \bgrm & \bgrand \\ \midrule
Med.
& 3.1 & 3.3\gsem{0.1} (\textbf{\textcolor{ForestGreen}{+0.2}})  & 3.6\gsem{0.1} (\textbf{\textcolor{ForestGreen}{+0.5}})
& 4.4 & 5.8\gsem{0.3} (\textbf{\textcolor{ForestGreen}{+1.4}})  & 6.1\gsem{0.1} (\textbf{\textcolor{ForestGreen}{+1.7}}) 
& 3.7 & 4.2\gsem{0.1} (\textbf{\textcolor{ForestGreen}{+0.5}}) & 4.1\gsem{0.1} (\textbf{\textcolor{ForestGreen}{+0.4}})  \\
Full 
& 4.2  & 4.7 (\textbf{\textcolor{ForestGreen}{+0.5}})  & 5.3 (\textbf{\textcolor{ForestGreen}{+1.1}})
& 5.3  & 7.2 (\textbf{\textcolor{ForestGreen}{+1.9}})  & 7.2 (\textbf{\textcolor{ForestGreen}{+1.9}})  
& 5.2  & 6.0 (\textbf{\textcolor{ForestGreen}{+0.8}})  & 5.7 (\textbf{\textcolor{ForestGreen}{+0.5}})  \\  
\bottomrule
    \end{tabular}
    }
    \caption{\textbf{Robustness: Natural Adversarial Examples.} Background augmentations improve performance on ImageNet-A, a data set of natural adversarial examples.}
    \label{tab:nat_adv_examples_deepusps}
\end{table}

We next evaluate classification performance on a particularly difficult distribution shift data set: ImageNet-A, a data set of natural adversarial examples that were found to be consistently mis-classified across models. These are extremely challenging for even supervised methods with ResNet-50 accuracy at only $\sim$2.2\% \citep{hendrycks2019nae}. As a first experiment, we investigate whether the difficulty of natural adversarial examples partially stems from misleading signal in the background. To test this, we modify the ImageNet-A data set by removing backgrounds such that only the foreground is present (Only-FG ImageNet-A). Indeed, we find that performance of supervised ResNet-50 improves by +2.8\%\footnote{We use the same pre-trained torchvision ResNet-50 model which was used in the construction of the data set. Since images mis-classified by this particular pre-trained model comprise the data set, the ImageNet-A (Only-FG ImageNet-A) accuracy for this specific model is 0\% (2.8\%), though a model trained from scratch has an ImageNet-A accuracy of $\sim$2.2\%.}, suggesting that some amount of the difficulty of natural adversarial examples stems from misleading information in the background. We note that the magnitude of this number must be interpreted with some caution, since this data set is also challenging for saliency detection.

We next investigate the performance of standard SSL methods on this task, finding substantively improved performance relative to the supervised methods (Table \ref{tab:nat_adv_examples_deepusps}). Despite this improvement, comparing the performance of SSL methods for the unmodified ImageNet-A \textit{vs}. Only-FG ImageNet-A (see Appendix \ref{app: only_fg_imagenet_a}) demonstrates that SSL models perform \textit{worse} on the version of ImageNet-A with only foregrounds, suggesting that SSL methods still may be overly focused on backgrounds. Together with the supervised results, this suggests that background augmentations in SSL should prove helpful. Indeed, we find that they are, with all versions of background augmentations resulting in substantially improved performance on ImageNet-A. In particular, we found \bgswaps~ to be more effective than \bgrm, suggesting the importance of using background matched negatives. These results demonstrate that part of the challenge of ImageNet-A stems from images with misleading backgrounds and that background augmentations can substantially improve robustness to these natural adversarial examples. Additional results: Appendix \ref{app:add_results} (Tables \ref{app:nat_adv_examples_deepusps_extended}, \ref{tab:nat_adv_examples_u2net_extended}).

\subsection{Improvement on ImageNet-Renditions}

We next investigate the performance on ImageNet-R \citep{hendrycks_many_2021}, a data set curated to measure generalization to various abstract visual renditions (e.g. paintings, embroidery, cartoons etc., see Figure \ref{fig:imnet_r_examples} for examples) of ImageNet classes. This is a challenging OOD data set for classifiers trained on ImageNet, since they often rely heavily on natural texture cues. Indeed, the supervised baseline accuracy for ResNet-50 is only 36.1\%. We find that background augmentations confer significant performance benefits of $\sim$2-6\%, suggesting that they help with generalizing to abstract visual renditions. Additional results: Appendix \ref{app:add_results} (Table \ref{tab:imagenet_r_u2net}).

\begin{table}
    \centering
    \resizebox{1\columnwidth}{!}{
    \begin{tabular}{cccccccccl}\toprule
  \shortstack{Pre-Train\\Duration}  & \multicolumn{3}{c}{\textbf{\moco}}  & \multicolumn{3}{c}{\textbf{BYOL}} & \multicolumn{3}{c}{\textbf{SwAV}}
    \\\cmidrule(lr){2-4} \cmidrule(lr){5-7} \cmidrule(lr){8-10} 
 & baseline & \bgrm & \bgswaps & baseline & \bgrm & \bgrand & baseline & \bgrm & \bgrand \\ \midrule
Med.
& 27.7 & 31.3\gsem{0.0} (\textbf{\textcolor{ForestGreen}{+3.6}})  & 32.3\gsem{0.1} (\textbf{\textcolor{ForestGreen}{+4.6}})
& 36.3 & 39.4\gsem{0.3} (\textbf{\textcolor{ForestGreen}{+3.1}})  & 38.4\gsem{0.0} (\textbf{\textcolor{ForestGreen}{+2.1}}) 
& 27.9 & 32.1\gsem{0.1} (\textbf{\textcolor{ForestGreen}{+4.2}})  & 31.2\gsem{0.3} (\textbf{\textcolor{ForestGreen}{+3.3}})  \\
Full
& 30.4 & 33.4 (\textbf{\textcolor{ForestGreen}{+3.0}})  & 33.5 (\textbf{\textcolor{ForestGreen}{+3.1}})
& 34.4 & 40.2 (\textbf{\textcolor{ForestGreen}{+5.8}})  & 39.2 (\textbf{\textcolor{ForestGreen}{+4.8}}) 
& 29.4 & 32.7 (\textbf{\textcolor{ForestGreen}{+3.3}}) & 32.5 (\textbf{\textcolor{ForestGreen}{+3.1}})  \\
\bottomrule
    \end{tabular}
    }
    \caption{\textbf{Robustness: Renditions.} Background augmentations improve performance on ImageNet-R, a data set of ImageNet-Renditions (e.g. paintings, sculpture).}
    \label{tab:imagenet_r_deepusps_sq}
\end{table}

\begin{table}[t]
    \centering
    \resizebox{1\columnwidth}{!}{
    \begin{tabular}{cccccccccl}\toprule
    \shortstack{Pre-Train\\Duration} & \multicolumn{3}{c}{\textbf{\moco}}  & \multicolumn{3}{c}{\textbf{BYOL}} & \multicolumn{3}{c}{\textbf{SwAV}}
    \\\cmidrule(lr){2-4} \cmidrule(lr){5-7} \cmidrule(lr){8-10} 
 & baseline & \bgrm & \bgswaps & baseline & \bgrm & \bgrand & baseline & \bgrm & \bgrand \\ \midrule
Med. & 4.5 & 6.4\gsem{0.0} (\textbf{\textcolor{ForestGreen}{+1.9}})  & 8.4\gsem{0.2} (\textbf{\textcolor{ForestGreen}{+3.9}})  
& 10.6  & 11.9\gsem{0.4} (\textbf{\textcolor{ForestGreen}{+1.3}})   & 11.4\gsem{0.1} (\textbf{\textcolor{ForestGreen}{+0.8}})  
& 6.0  &  6.6\gsem{0.1} (\textbf{\textcolor{ForestGreen}{+0.6}})    &  6.7\gsem{0.1} (\textbf{\textcolor{ForestGreen}{+0.7}})   \\
Full & 7.8 & 10.6 (\textbf{\textcolor{ForestGreen}{+2.8}})  & 13.1 (\textbf{\textcolor{ForestGreen}{+5.3}})  & 10.4  & 13.2 (\textbf{\textcolor{ForestGreen}{+2.8}})   & 13.4 (\textbf{\textcolor{ForestGreen}{+3.0}})  & 9.1  &  10.1 (\textbf{\textcolor{ForestGreen}{+1.0}})    &  10.4 (\textbf{\textcolor{ForestGreen}{+1.3}})    \\  
\bottomrule
    \end{tabular}
    }
    \caption{\textbf{Robustness: Adversarial Attack.} Background augmentations increase robustness to FGSM adversarial attacks.}
    \label{tab:adv_attacks_deepusps}
\end{table}

\subsection{Background Augmentations Improve Robustness to Adversarial Perturbations}

\citet{ilyas_neurips_2020_features_not_bugs} demonstrated that adversarial examples are partially driven by the learning of non-robust, high frequency features which can be predictive of ground-truth classification labels, but which are also highly susceptible to adversarial attacks. Since background augmentations encourage focus on semantically meaningful content in images, a natural question is whether these augmentations also confer increased robustness to adversarial perturbations. To test this, we use a popular adversarial attack: FGSM \citep{goodfellow_fgsm}. We found that background augmentations did indeed result in increased robustness, with \bgswaps~consistently conferring a greater benefit than \bgrm~(Table \ref{tab:adv_attacks_deepusps}), once again emphasizing the importance of penalizing focus on backgrounds. Additional results: Appendix \ref{app:add_results} (Table \ref{app:adv_attacks_u2net}).

\subsection{Evaluation on CIFAR-10 and 100}

We find that the performance benefits of including background augmentations extends to CIFAR-10 and 100, see Table \ref{tab:cifar_deepusps}. All methods used the same protocol to be directly comparable. All models receive full pre-training. Additional results: Appendix \ref{app:add_results} (Table \ref{app:cifar_u2net}).

\begin{table}[ht]
    \centering
    \resizebox{1\columnwidth}{!}{
    \begin{tabular}{cccccccccl}\toprule
    Data Set & \multicolumn{3}{c}{\textbf{\moco}}  & \multicolumn{3}{c}{\textbf{BYOL}} & \multicolumn{3}{c}{\textbf{SwAV}}
    \\\cmidrule(lr){2-4} \cmidrule(lr){5-7} \cmidrule(lr){8-10} 
 & baseline & \bgrm & \bgswaps & baseline & \bgrm & \bgrand & baseline & \bgrm & \bgrand \\ \midrule
CIFAR-10 & 73.9 & 80.7 (\textbf{\textcolor{ForestGreen}{+6.8}}) & 76.0 (\textbf{\textcolor{ForestGreen}{+2.1}})  & 86.7  & 87.7 (\textbf{\textcolor{ForestGreen}{+1.0}}) & 88.1 (\textbf{\textcolor{ForestGreen}{+1.4}})  &  92.7 &   92.7 (+0.0)  &   92.9 (\textbf{\textcolor{ForestGreen}{+0.2}})    \\
CIFAR-100 & 40.8 & 51.6 (\textbf{\textcolor{ForestGreen}{+10.8}}) & 44.9 (\textbf{\textcolor{ForestGreen}{+4.1}})  & 67.6  & 66.5 (\textbf{\textcolor{BrickRed}{-1.1}}) & 67.0 (\textbf{\textcolor{BrickRed}{-0.6}})  & 76.0  &  76.4   (\textbf{\textcolor{ForestGreen}{+0.4}})  &   76.4 (\textbf{\textcolor{ForestGreen}{+0.4}})    \\\bottomrule
    \end{tabular}
    }
    \caption{\textbf{CIFAR-10, 100.} Background augmentations improve performance on linear evaluation on CIFAR-10 and 100.}
    \label{tab:cifar_deepusps}
\end{table}

\subsection{A Limitation of Learning Background Invariance} 
\label{sec:places205} 
We have characterized the impact of background augmentations in view-invariant SSL, finding improved generalization, robustness, label and training efficiency. Here, we discuss an important \textit{limitation} of our work. As previously discussed, \textit{by design} SSL augmentations are meant to induce ``desirable" invariances---what is desirable depends on the downstream tasks (e.g. \citet{purushwalkam2020demystifying, xiao_multihead_2021, tian_what_2020}). Consequently, when \textit{background is informative} to the task at hand, we expect poorer performance. We demonstrate this by linear evaluation on Places-205, finding that this is indeed the case, see Table \ref{tab:places205}. Note that this limitation is not specific to background augmentations. Indeed, ``aggressive" cropping is an integral part of the augmentation pipeline in nearly all high performing SSL methods but can be detrimental \citep{purushwalkam2020demystifying} like background augmentations, in similar situations. 

This limitation of background augmentations on domains different from intended application may be overcome by training a multi-head network with a shared backbone (as in \citet{xiao_multihead_2021}), so that one head is trained to be background invariant, while one head is not. All models receive full pre-training; foreground masks used for background augmentations were based on U$^2$Net to control for mask quality.

\begin{table}
    \centering
    \resizebox{1\columnwidth}{!}{
    \begin{tabular}{ccccccccl}\toprule
    \multicolumn{3}{c}{\textbf{\moco}}  & \multicolumn{3}{c}{\textbf{BYOL}} & \multicolumn{3}{c}{\textbf{SwAV}}
    \\\cmidrule(lr){1-3} \cmidrule(lr){4-6} \cmidrule(lr){7-9} 
 baseline & \bgrm & \bgswaps & baseline & \bgrm & \bgrand & baseline & \bgrm & \bgrand \\ \midrule
 28.7 & 27.3 (\textbf{\textcolor{BrickRed}{-1.4}}) & 25.8 (\textbf{\textcolor{BrickRed}{-2.9}})   &   
 44.5 & 40.0 (\textbf{\textcolor{BrickRed}{-4.5}})   & 42.1 (\textbf{\textcolor{BrickRed}{-2.4}}) &   
 49.6 & 48.1 (\textbf{\textcolor{BrickRed}{-1.5}}) & 48.1 (\textbf{\textcolor{BrickRed}{-1.5}})   \\ \bottomrule
    \end{tabular}
    }
    \caption{{\bf When background is relevant: Places-205.} When background information is important, background augmentations can reduce downstream performance.}
    \label{tab:places205}
\end{table}

\subsection{Object Detection and Instance Segmentation}
We report evaluation on the downstream tasks of object detection and instance segmentation, since these are common evaluations for SSL methods. However, a priori we expect background augmentations to yield only small gains in these tasks, since the models receive extensive supervised information about object identities and locations during finetuning. Indeed, identity information alone can induce strong localization ability \citep{simonyan_deep_2013}. Consistent with our expectations, we see only small gains in these tasks in Table \ref{tab:voc_coco_deepusp}. We note that it is possible that background augmentations may yield larger gains in these tasks with less training or by incorporating the augmentations into the finetuning pipeline \citep{ghiasi2020simple}. Additional results: Appendix \ref{app:add_results} (Table \ref{tab:voc_coco_u2net}).

\begin{table}
    \centering
    \resizebox{1\columnwidth}{!}{
    \begin{tabular}{cccccccccc}\toprule
    & \multicolumn{3}{c}{VOC 07+12 detection}  & \multicolumn{3}{c}{COCO detection} & \multicolumn{3}{c}{COCO instance seg.}
    \\\cmidrule(lr){2-4} \cmidrule(lr){5-7} \cmidrule(lr){8-10} 
Method & AP$_{50}$ & AP & AP$_{75}$ & AP$_{50}$ & AP & AP$_{75}$ & AP$_{50}^{m}$ & AP$^{m}$ & AP$_{75}^{m}$ \\ \midrule
\moco~{\scriptsize{(\textit{repro.})}}            
& 82.7\gsem{0.0} & 57.9\gsem{0.0} & 64.5\gsem{0.1} &
61.0 & 41.1 & \textbf{44.8} 
& 57.7 & 35.8 & 38.4\\
\rowcolor{lightgray}
\moco~+ \bgrm    
& 82.9\gsem{0.1} & \textbf{58.1}\gsem{0.1} & \textbf{65.2}\gsem{0.2}
& 61.2 & 41.2 & 44.7 
& 58.0 & \textbf{36.0} & \textbf{38.6}\\
\rowcolor{lightgray}
\moco~+ \bgswaps 
& 82.7\gsem{0.0} & 57.5\gsem{0.0} & 63.9\gsem{0.1} 
& 61.1 & 41.1 & 44.3 
& 57.6 & 35.8 & 38.3\\ \midrule
BYOL {\scriptsize{(\textit{repro.})}}       
& 82.7\gsem{0.1} & 56.7\gsem{0.1} & 63.0\gsem{0.3}
& 61.1 & 40.9 & 44.5 
& 57.6 & 35.5 & 37.8\\
\rowcolor{lightgray}
BYOL + \bgrm      
& 83.0\gsem{0.1} & 57.0\gsem{0.0} & 64.0\gsem{0.1}
& 61.5 & 41.1 & 44.4 
& 57.9 & 35.6 & 38.0\\
\rowcolor{lightgray}
BYOL + \bgrand   
& \textbf{83.1}{\scriptsize{$\pm$0.2}} & 57.6{\scriptsize{$\pm$0.1}} & 64.7{\scriptsize{$\pm$0.1}} 
& \textbf{61.7} & \textbf{41.4} & 44.7 
& \textbf{58.4} & \textbf{36.0} & 38.3\\ \midrule
SwAV {\scriptsize{(\textit{repro.})}}       
& 82.3\gsem{0.1} & 55.6\gsem{0.0} & 61.9\gsem{0.2} 
& 61.4 & 40.7 & 43.7 
& 57.6 & 35.4 & 37.4\\
\rowcolor{lightgray}
SwAV + \bgrm      
& 82.4\gsem{0.0} & 55.9\gsem{0.1} & 62.2\gsem{0.2}
& 61.2 & 40.6 & 44.0 
& 57.6 & 35.4 & 37.4\\
\rowcolor{lightgray}
SwAV + \bgrand   
& 82.4\gsem{0.1} & 55.9\gsem{0.1} & 62.4\gsem{0.2}
& 61.2 & \textbf{41.4} & \textbf{44.8} 
& 58.0 & \textbf{36.0} & 38.3\\
\bottomrule
    \end{tabular}
    }
    \caption{\textbf{Detection and Instance Segmentation}. Background Augmentations result in small improvements in detection and instance segmentation tasks, likely due to extensive supervision involved in subsequent training. All VOC metrics reported are average of 3 independent runs.}
    \label{tab:voc_coco_deepusp}
\end{table}

\subsection{Background Augmentations Increase the Shape Bias of SSL Methods}
\label{sec:shape_bias}

Supervised Convolutional Neural Networks (CNNs) have been found to be biased toward texture, i.e. they tend to classify based on the texture information in an image over shape, whereas humans are more shape biased; increasing the shape bias of supervised CNNs has been found to increase accuracy and robustness \citep{geirhos2019imagenettrained}. Recent work \citep{geirhos2020surprising} has also found that many SSL methods are heavily texture biased like their supervised counterparts. We use the shape bias measure \citep{geirhos2019imagenettrained} to probe the pre-trained SSL models to gain some insight. The shape bias of a model is computed using texture-shape cue conflict stimuli (the shape and texture cues in the image correspond to different ImageNet classes, e.g. see Figure \ref{fig:cue_conflict_stim}) as the fraction of classification decisions that correspond to shape information.

We find that (see Table \ref{tab:shape_bias_deepusps}) while the SSL methods considered are heavily texture biased, they are less so than their supervised counterpart, with the exception of SwAV. However, the default setting of SwAV uses \texttt{multi-crop} with 2 global views and 6 local views; the local views may be expected to push the model to be biased toward local texture features. Consistent with this hypothesis, SwAV trained without \texttt{multi-crop}\footnote{We evaluated the shape-bias of an official SwAV model trained for 400 epochs without \texttt{multi-crop} from \url{https://github.com/facebookresearch/swav.}} has a shape bias of 27.4. Our second finding is that across all SSL methods, \textit{background augmentations increase shape bias} (Tables \ref{tab:shape_bias_deepusps}, \ref{app:shape_bias_u2net}). We note that our improvements on the ImageNet-R data set, whose texture cues are OOD relative to ImageNet, may have been driven in-part by the increased shape bias of the models trained using background augmentations. Our findings raise the intriguing possibility that background augmentations induce representations that are (slightly) more brain-like. All models receive full pre-training.

\begin{table}
    \centering
    \resizebox{1\columnwidth}{!}{
    \begin{tabular}{cccccccccc}\toprule
    Supervised & \multicolumn{3}{c}{\textbf{\moco}}  & \multicolumn{3}{c}{\textbf{BYOL}} & \multicolumn{3}{c}{\textbf{SwAV}}
    \\\cmidrule(lr){2-4} \cmidrule(lr){5-7} \cmidrule(lr){8-10} 
& baseline & \bgrm & \bgswaps & baseline & \bgrm & \bgrand & baseline & \bgrm & \bgrand \\ \midrule
 22.1 &  28.8 & 31.7  & 33.4  & 27.6  & 29.8  &   31.0  &   17.0 & 17.7 & 19.4 \\\bottomrule
    \end{tabular}
    }
    \caption{\textbf{Background augmentations increase shape bias.} SSL methods considered generally have a higher shape bias than the supervised baseline. SwAV deviates from this pattern due to \texttt{multi-crop} (SwAV w/o \texttt{multi-crop} shape bias: 27.4).}
    \label{tab:shape_bias_deepusps}
\end{table}

\section{Related Work}
\label{sec:related_work}

\paragraph{Semantic Focus and Robustness.} A number of recent works have investigated whether non-semantic features are exploited by models in supervised learning. We draw heavy inspiration from this literature, especially \citet{xiao2020noise}, \citet{sehwag2020time} and \citet{beery_recognition_2018} who demonstrate the importance of backgrounds for image classification. Other works have demonstrated the importance of high-frequency information for classification, both in traditional image classification \citep{jo2017measuring} and in the context of adversarial robustness \citep{ilyas_neurips_2020_features_not_bugs}. There have also been a number of works investigating the importance of shape \textit{vs}. texture for classification decisions, both in supervised \citep{geirhos2019imagenettrained, hermann2019origins} and self-supervised learning \citep{geirhos2020surprising}. Similar to the findings in the supervised setting in \citet{geirhos2019imagenettrained}---that increasing the shape-bias increases robustness and accuracy, we found that background augmentations increase shape-bias and also improve robustness and accuracy. 

While there has been much work investigating robustness properties in the supervised setting (e.g. \citet{xiao2020noise, hendrycks2019nae, hendrycks_many_2021, goodfellow_fgsm}), the self-supervised setting has received relatively less attention. \citet{geirhos2020surprising} characterize the robustness of several SSL models to low-level noise distortions but do not investigate other aspects of robustness nor approaches to improve semantic focus and performance. We evaluate a diverse spectrum of high performing SSL methods in 17 distribution shift settings, in addition to investigating approaches to improve robustness. Thus, our work is complementary to existing work.

\paragraph{Self-Supervised Learning.} We do not make a formal distinction between self-/un-supervised learning (but see \citet{ssl_review}) and broadly discuss  related work. Generally, representation learning without human-annotated labels involves solving ``pretext" prediction tasks. We coarsely organize the literature as follows.

\textit{ Hand-crafted pretext tasks.} Early work used hand-crafted pretext tasks such as predicting image orientation (RotNet, \citet{gidaris2018unsupervised}), image inpainting \citep{inpainting_ssl}, solving image jigsaw puzzles \citep{jigsaw_ssl}, denoising \citep{vincent_extracting_2008} and cross-channel \citep{colorization_ssl, split-brain_2017} auto-encoding for representation learning. Combining multiple pretext tasks \citep{doersch_multi-task_2017} and using larger networks \citep{kolesnikov_revisiting_2019} can improve performance.

 \textit{Learning view invariance.} While hand-crafted pretext tasks have been shown to be useful for learning representations useful for downstream tasks, their performance has been far from their supervised counterparts. Learning \textit{view-invariant} representations has recently been a fruitful direction in SSL; such approaches date back to \citet{becker_self-organizing_1992}. We coarsely group such works based on how trivial representations are avoided. 

\begin{itemize}[leftmargin=2em]
    \itemsep0em 
    \item  \textit{Contrastive learning.} Contrastive learning \citep{hadsell_dimensionality_2006} is a framework for learning representations from data organized into similar/dissimilar pairs. Contrastive learning prevents trivial representations through use of dissimilar pairs and has been a popular design choice in SSL \citep{he2019moco, chen2020mocov2, chen2020simple, chen2020big, wu_unsupervised_2018, infonce_2018, hjelm_learning_2019, ye_unsupervised_2019, henaff_data-efficient_2020, bachman_learning_2019, tian_contrastive_2020, misra2020pirl, dosovitskiy_discriminative_2014, fnc}. 
    
     \item \textit{Clustering.} A number of SSL methods have avoided trivial representations through clustering \citep{asano_self-labelling_2020, caron_deep_2018, caron_unsupervised_2019, swav, ji_invariant_2019}. There has also been work \citep{li_prototypical_2021, zhuang_local_2019} that bridges clustering and contrastive learning approaches.
     
     \item \textit{Other.} Methods such as BYOL \citep{grill2020bootstrap}, SimSiam \citep{simsiam} and Barlow Twins \citep{zbontar_barlow_2021} are not explicitly contrastive nor based on clustering and prevent trivial representations in other ways.
\end{itemize}

While we compare performance with respect to numerous SSL methods to situate our work in literature, we note that we do not propose any new SSL methods. Rather, we improve upon the core ingredient of the best performing methods: the augmentation pipeline. We choose one SSL method from each coarse grouping of the literature to form a diverse test bed of SSL methods, so as to characterize when background augmentations can or cannot confer benefits, as well as to demonstrate the generality of our results. We show that learning background invariance improves performance, robustness and label efficiency across a diverse spectrum of high-performing SSL methods. Importantly, our extensive analyses led to insights that allowed us to improve performance beyond a plug-and-play approach. While we focus on view-invariant SSL approaches that differently augment the same image to generate views, background augmentations can also be applied to approaches that use different frames from video to generate views (e.g. \citet{zhuang_unsupervised_2020, sermanet_time-contrastive_2018, gordon_watching_2020, han_video_2019}).

\paragraph{Analyzing and Improving SSL Augmentation Pipelines.} The augmentation pipeline for most high-performing SSL methods is similar. A number of recent studies have focused on analyzing and improving this pipeline, e.g. \citet{tamkin2021viewmaker} learn the augmentations jointly with the contrastive learning objective; \citet{tian_what_2020} use labeled data to learn color spaces which are then split to generate views and also characterize ImageNet acc. \textit{vs}. augmentation strength for many augmentations. \citet{purushwalkam2020demystifying} investigate invariance to occlusion, viewpoint, and category instance and show that common SSL pipelines encourage occlusion invariance---a useful property for object recognition tasks.  \citet{tian_what_2020} observe on a synthetic toy dataset that the background can overwhelm the foreground, but do not investigate further nor propose a solution.

 We show that background augmentations improve semantic focus in representations, leading to better generalization and robustness. Of particular note, \citet{selvaraju2020casting} also aims to improve focus on semantically meaningful content in images; they do so by constraining the crops used in the SSL augmentation pipeline to contain the object of interest as determined by a saliency method. We also investigate the impact of constraining crops to contain the salient object, and find that it does not improve performance (Appendix \ref{app:rand_crops}) on top of background augmentations. Critically, in contrast with our work, \citet{selvaraju2020casting} relies on a saliency detector trained using ImageNet class labels making this method not truly self-supervised. Note that \citet{selvaraju2020casting} also investigate using an ``attention" loss computed using Grad-CAM \citep{grad-cam_2017}, which we discuss below.

\paragraph{``Copy-Paste" Augmentations.} There is a long history of work, largely in the supervised setting, that has investigated the use of ``copy-paste" augmentations in which ``copied" foregrounds are ``pasted" onto backgrounds, generally with the aim of generating more \textit{labeled} data. We draw heavy inspiration from this literature. Copy-paste has been used for learning optical flow \citep{dosovitskiy_flownet_2015}, instance detection \citep{Dwibedi_2017_ICCV}, object detection \citep{georgakis_synthesizing_2017, Dvornik_2018_ECCV}, text localization \citep{gupta_synthetic_2016} and instance segmentation \citep{remez_learning_2018, fang_instaboost_2019, ghiasi2020simple}. In these works, the segmentation masks required for copying are obtained from human annotation or using networks trained in a supervised manner to generate them. This implicit or explicit reliance on human annotation has been an obstacle limiting the application of copy-paste outside of the supervised setting where it is widely used. 

Recent work \citep{ZhaoAAAI2021} took a first step in the SSL setting by using a heuristic saliency method to generate a mask, and applied a ``copy-paste" augmentation similar to \bgrm, finding improved performance on ImageNet linear classification accuracy. However, the gain achieved in their \textit{best} performing SSL method (\moco+DiLo-RBD) is only +0.2\% (see Table \ref{tab:gain_summary_imgnet_deepusps}). Thus, it remains unclear if such augmentations can significantly benefit SSL, especially high performing SSL methods---indeed, in \citet{ZhaoAAAI2021}, the gains rapidly and monotonically decline with the baseline SSL method's performance. Thus, not only is it unclear if copy-paste can significantly benefit SSL, it is unclear when and how such augmentations confer benefits. Further, the impact of such augmentations on downstream tasks is also unclear, though \citet{ZhaoAAAI2021} report small gains ($\sim$0.4-0.6 AP \%) on object detection and instance segmentation tasks.

In contrast, in our work, \textit{a}) we develop a completely unsupervised method to generate high quality masks and demonstrate the utility of background augmentations in conferring large performance benefits ($\sim$1-2\%) across a spectrum of high performing SSL methods (e.g. we obtain a 7$\times$ larger gain of +1.4 on \moco~using \bgrm), \textit{b}) we systematically characterize \textit{when} and \textit{how} background augmentations confer benefits: it is not sufficient to merely decorrelate the foreground and background nor to disincentive focus on the background; rather, background augmentations confer benefits when there is a similarity comparison between images in view-invariant SSL, where the background maybe used as a shortcut, \textit{c}) \textit{Contrary} to \citet{ZhaoAAAI2021}, we show that using natural random backgrounds (\bgrand) can result in \textit{better performance}, \textit{d}) further, our insights on how background augmentations work enable us to develop a novel, more effective background augmentation method (\bgswaps), leading to even larger performance gains (+1.8) (a 9$\times$ larger gain on \moco), \textit{e}) we show how benefits from background augmentations may be hidden by implementation choices, \textit{f}) Perhaps most critically, we focus on generalization in OOD settings, robustness, label and training efficiency, \textit{g}) we also characterize the limitations of background augmentations. 

 We note that in our work, we use the term ``background augmentation" rather than ``copy-paste" augmentation, since our purpose is to discourage focus on the backgrounds, rather than creating more \textit{labeled} data for the foreground (e.g. \citet{ghiasi2020simple,Dwibedi_2017_ICCV}). However, more broadly, even in absence of labels our work enables \textit{foreground} augmentations for SSL, making it possible to create images with multiple objects in a controlled manner.

\paragraph{Mixing Augmentations.}
Background augmentations have some resemblance to mixing augmentations used in the supervised setting, e.g. mixup \citep{mixup}, CutMix \citep{yun_cutmix_2019}, which mix information contained in images. In contrast with background augmentations, these methods \textit{a}) mix images ignoring the semantic relevance of parts of an image and \textit{b}) also mix the corresponding labels. Extending such mixing augmentations to SSL is an orthogonal improvement to our work and may be a fruitful line of inquiry for future work. Relatedly, but conversely, mixing augmentations that consider semantic relevance of parts of an image in the supervised setting, could also be a fruitful direction.

\paragraph{``Attention" Loss.} \citet{selvaraju2020casting} investigate the impact of using an additional ``attention" loss that encourages similarity between the Grad-CAM heatmap of the query and its saliency mask. The Grad-CAM heatmap is computed by additionally encoding the masked (background removed) key using the teacher network and computing the spatial regions in the query that the network relies on to map the masked key to the query, by back-propagation on the activations in the last convolutional layer. In contrast, our work, besides not relying on a saliency detector trained using supervised information, is much simpler---simply adding an augmentation to the data augmentation pipeline and thereby more agnostic to the specific method and yet, is highly effective.

\paragraph{``Hard" Instances.} 
Our work is also related to the literature on ``hard negatives", which have been explored recently in the context of contrastive SSL to improve learning \citep{kalantidis2020hard, wu2021conditional, robinson2021contrastive, cai_2020}. In this literature, hard negative instances are ``mined" using distances in the embedding space. In a broader sense, creating negatives whose background matches the query (as we do in \bgswaps) can also be considered ``hard negatives" and in a similar vein, one could consider the positive pair ($q$ and $k^+$) with different backgrounds as ``hard positives". \textit{Mining} for hard negatives (or even hard positives) is an orthogonal improvement to our work.

\section{Discussion} \label{sec:discussion}

We investigated the potential of background augmentations for self-supervised learning. We explored several variants of background augmentations, including those with constant gray backgrounds (\bgrm), randomized natural backgrounds (\bgrand), and a novel method that uses matched backgrounds between queries and negatives (\bgswaps). Across view-invariant SSL methods, we found that background augmentations result in sizeable performance improvements $\sim$1-2\% on ImageNet linear classification, enabling performance on par with the supervised baseline. In the limited-labels setting, we found \textit{even larger} performance benefits. Across SSL methods, we found that \bgrand~and \bgswaps~often lead to larger performance improvements than \bgrm~due to being more in-distribution. However, other factors such as ease of optimization (Section \ref{sec:diagnosis_swav}) also play a role. Background augmentations take a large step forward in reducing the amount of training required for competitive performance in SSL, e.g. enabling performance on par with or better than many recent SSL methods trained for 800-1000 epochs in only 100 epochs.

Interestingly, we found that background augmentations conferred no benefit in supervised training nor in RotNet, an SSL method not based on view-invariance. These results demonstrate the importance of designing augmentations tailored to the SSL setting, especially view-invariant SSL. These findings are timely and relevant to the community, since view-invariant SSL methods are currently the best performing methods across a range of architectures and downstream tasks.

As SSL methods shrink the gap to their supervised counterparts, it has become increasingly important to characterize the limitations of performant SSL methods as well as to understand their robustness and generalization properties in a more comprehensive manner. Across state-of-the-art SSL methods, we found that background augmentations enable increased model focus on semantically meaningful content and lead to improved robustness to numerous distribution shifts including ImageNet-9, natural adversarial examples, ImageNet-R, adversarial attacks, as well as natural distribution shift. Our analyses revealed an increased shape bias for SSL models trained with background augmentations, which may have driven some of the improvement, especially on the ImageNet-R data set where texture cues are OOD relative to ImageNet, requiring representations to better encode shape cues in an image for improved performance. All of these results raise the intriguing possibility that background augmentations induce representations that are (slightly) more brain-like. Future work could investigate this idea further, potentially by comparing representations with neuronal recordings \citep{yamins_pnas_2014}. Relatedly, neural networks are known to be easily fooled by objects in unusual poses \citep{alcorn_strike_2019}, unlike humans. An interesting line of investigation for future work could be to learn robustness to unusual poses by \textit{foreground} augmentations (e.g. using foreground masks to augment data with rotated objects). Indeed, such approaches have been adopted in the supervised setting \citep[e.g.][]{Dwibedi_2017_ICCV, ghiasi2020simple} when segmentation masks for foreground objects are available. Our work enables such approaches in the \textit{absence of human-annotation}, opening up new possibilities beyond our application here in background augmentations.

It is worth noting that background augmentations as implemented here are specific instantiations of encouraging background invariance---we focused on extensively evaluating \textit{simple} instantiations. Specifically, using a saliency method to separate foreground and background could be problematic in more complex scenes. Remarkably, though ImageNet has a large share \citep[e.g.][]{stock_convnets_2018,beyer2020imagenetreal} of multi-object multi-class images, the simple approach we have taken here works well. More sophisticated approaches hold the potential for further improvement, e.g. one straightforward approach could be to copy foregrounds from simple images using saliency detection and paste multiple objects into images to create more complex scenes in controlled manner, either offline or on-the-fly.

There has been increasing recent interest \citep{chen_empirical_2021,caron_emerging_2021,li_efficient_2021} in self-supervised learning for Vision Transformers (ViTs, \citet{vit_dosovitskiy_2021}). Future work could investigate whether background (or foreground) augmentations can benefit SSL for ViTs. Yet another interesting line of inquiry could be to investigate the impact of background augmentations in high performing semi-supervised learning methods (e.g. FixMatch \citep{fixmatch_sohn_2020}).

\acks{We thank Xinlei Chen, Mathilde Caron, and Saining Xie for helpful discussions, and Ting-Wu Rudy Chin and Matthew Leavitt for feedback on an early draft of the manuscript. }

\appendix
\newpage

\clearpage
\onecolumn
\renewcommand\thesection{\Alph{section}}
\setcounter{section}{0}
\renewcommand\thefigure{A\arabic{figure}}
\setcounter{figure}{0}
\renewcommand\thetable{A\arabic{table}}
\setcounter{table}{0}
\phantomsection

\label{sec:appendix}

\paragraph{Organization of Supplementary Information.} Appendix \ref{app:training_details} contains implementation details and evaluation protocols. In Appendix \ref{app:noisy_masks}, we characterize the impact of foreground mask quality by systematically distorting the masks in numerous ways. Appendix \ref{app:ablations} contains additional ablations. Appendix \ref{app:add_results} contains additional results, including characterization of robustness to image corruptions and evaluations of models where background augmentations used masks generated by a supervised saliency method, U$^2$Net. Table captions throughout the appendices have additional redundancy to increase ease of reference.

\section{Implementation Details} 
\label{app:training_details}
In this Appendix, we discuss pre-training details for each of the SSL methods in our test bed and the protocols followed for downstream evaluations. We also discuss the implementation details of the unsupervised saliency detection method, DeepUSPS$^2$. 

\paragraph{General Settings.} All experiments use the ResNet-50 \citep{resnet} architecture unless otherwise indicated. All specified learning rates are \textit{base learning rates} for a batch size of 256 unless otherwise indicated. Learning rates are obtained by linearly scaling the base learning rates as base learning rate $\times$ batch size$/256$. We closely follow the implementation details of the original works where possible. Settings not mentioned here are identical to respective original works. All models were implemented using PyTorch \citep{pytorch_cite}, torchvision \citep{marcel_torchvision_2010} and NumPy \citep{numpy}. All cosine schedules \citep{loshchilov_sgdr_2016} are half-period without restarts. Binary foreground masks used in background augmentations are obtained by thresholding saliency predictions between 0-1 using a threshold value of 0.9; by default, we use our unsupervised saliency detector DeepUSPS$^2$.

\subsection{Pre-Training of SSL Methods}

\paragraph{\moco.} We use a larger (than the standard 256) batch size of 1024 (distributed across 32 GPUs) with a 20 epoch warmup for 220 (800) epochs in the medium (full) setting. These setting were chosen to speed up pre-training while matching (or improving upon) the reported performance at a similar number of epochs in \citet{chen2020mocov2}. 

{\it Background Augmentations.} In \bgrm,~backgrounds were removed in $q$ and $k^+$ independently with $p=0.2$. In \bgswaps,~$p_{\text{pos}}=p_{\text{neg}}=0.2$ and crops in \texttt{RandomResizedCrop} (\texttt{RRC}) were constrained to include $\text{FG}_{\text{min}}=0.1$ fraction of the foreground, by rejection sampling. Concretely, \texttt{RRC} parameters were sampled up to 10 times to satisfy constraints, defaulting to a \texttt{CenterCrop} if no satisfactory parameters are sampled. See Appendix \ref{app:rand_crops} for additional discussion and ablations. In \bgswaps, the background matched negatives are batched together with the positive keys during forward pass through the teacher/key network; only the positives keys are subsequently placed in the queue $Q$.

\paragraph{\byol.} We used a batch size of 4096 (distributed across 64 GPUs). Our implementation used synchronized batch-normalization layers (synced per group of 8 GPUs) using the apex\footnote{\url{https://github.com/NVIDIA/apex}} library. In \texttt{RRC}, we used a scale setting of $(0.2, 1.0)$. We obtained similar results in the medium setting (300 epochs) when we instead used  \textit{a}) synchronized batch-normalization layers across all GPUs (global sync) or \textit{b}) used a scale setting of $(0.08, 1.0)$ as in \citep{grill2020bootstrap}, but this may be different in the full setting (1000 epochs), potentially further improving on the performance we obtained in the full setting using background augmentations. 

{\it Background Augmentations.} In \bgrm,~$p=0.15$ and $\text{FG}_{\text{min}}=0.05$, while in \bgrand,\\~$p=0.05$ and $\text{FG}_{\text{min}}=0.05$.

\paragraph{SwAV.} Pre-training was identical to original implementation \citep{swav}. 
{\it Background Augmentations.} In \bgrm,~$p=0.25$ and $\text{FG}_{\text{min}}=0.15$, while in \bgrand,~$p=0.2$ and $\text{FG}_{\text{min}}=0.15$. We only apply background augmentation to the global views in \texttt{multi-crop}.

\paragraph{RotNet.} Pre-training procedure was largely similar to the original paper \citep{gidaris2018unsupervised}.  When background augmentations were used, \bgrm~and \bgrand~were applied before the default augmentations of \texttt{RandomResizedCrop}, \texttt{RandomHorizontalFlip} and \texttt{Rotation}. Models were pre-trained for 30 epochs with a learning rate of 0.01, with a step schedule (10, 20) and a decay factor of 0.1 using SGD with momentum=0.9, a batchsize of 192 and weight decay of $5\times10^{-4}$.   

\begin{figure}
    \centering
    \includegraphics[width=1\textwidth]{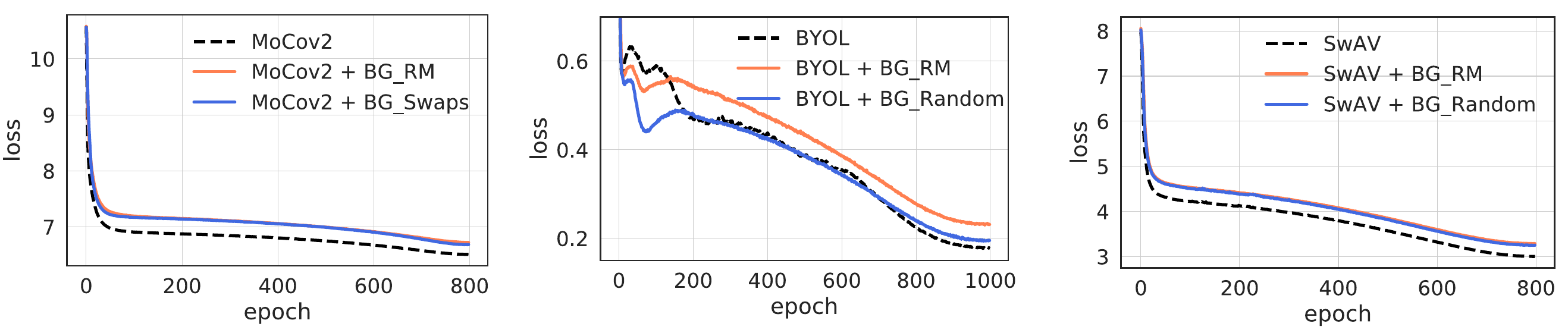}
    \caption{\textbf{Higher pre-training loss but better generalization.} Background augmentations generally lead to a higher final pre-training loss as they make the objective function more “difficult", but can lead to better generalization.}
    \label{fig:loss_curves}
\end{figure}

\paragraph{Discussion: Loss and Background Augmentations.}
Background augmentations make the objective function more “difficult", leading to a higher final pre-training loss but can lead to better generalization, see Figure \ref{fig:loss_curves}. This is consonant with previous work \citep{he2019moco,kolesnikov_revisiting_2019} finding that the loss of the pretext task is not necessarily monotonically related to generalization performance.

\subsection{Training DeepUSPS\texorpdfstring{$^2$}{2}}
\label{app:train_saliency_impl_details}
We trained a DRN-D-105 \citep{drn_yu_2017} network via BYOL for 500 epochs to an accuracy of 73.9\% with the following settings: base learning rate of $0.3$, weight decay of $1\times10^{-6}$ and momentum coefficient of $0.99$ for the teacher network. All other settings use the defaults for training BYOL. We then use this network as an initialization to train a saliency detector. Note that previous work instead initialized with a network trained using ImageNet class labels as well as CityScapes \citep{cordts_cityscapes_2016} segmentation labels. We train this network following the procedure from DeepUSPS, but in phase 1 of training, we use a learning rate of $6\times10^{-6}$ instead of the default value of $1\times10^{-6}$.

\subsection{Linear Evaluation on ImageNet}
\label{app:imagenet_lin_eval}
Linear evaluation protocol was largely similar to original work. For \moco~and SwAV, we evaluate with a larger batch size to speed up evaluation. We train the linear classifier for more epochs in the case of \moco~and BYOL to reduce variability in the results. Note that warmup is not required, but for simplicity we opted to keep the training procedure close to standard supervised training.

\paragraph{\moco.} The linear classifier was trained for 120 epochs, with a step schedule of 60, 80, 100 and a decay factor of 0.1, with a warmup of 10 epochs, a batch size of 2048 (distributed across 32 GPUs). Parameters in the backbone were frozen to the pre-trained values.

\paragraph{BYOL.} The linear classifier was trained for 140 epochs, with a step schedule of 80, 100, 120 and a decay factor of 0.1. We used a base learning rate of 0.2 and a batch size of 1024 distributed across 16 GPUs. Parameters in the backbone were frozen to the pre-trained values.

\paragraph{SwAV.} The linear classifier was trained for 100 epochs with 5 warmup epochs using a batch size of 2048 (distributed across 32 GPUs) and a cosine schedule. Parameters and buffers in the backbone were frozen to the pre-trained values.

\paragraph{RotNet.} The linear evaluation procedure  was largely similar to the original paper \citep{gidaris2018unsupervised}. A linear classifier was trained on top of Conv5 layer using SGD with Nesterov momentum over 35 epochs using a batchsize of 256 and a momentum of 0.9 with a learning rate of 0.1, a step schedule of (5, 15, 25) and weight decay of $5\times10^{-4}$.  

\paragraph{General Settings.} Following common protocol, pre-processing during training consists of \texttt{RandomResizedCrop} and \texttt{RandomHorizontalFlip} followed by normalization. The pre-processing on validation images consists of \texttt{Resize} to size 256 along the shorter edge of the image, followed by \texttt{CenterCrop} to size 224$\times$224 and normalization.

\subsection{Background Augmentations in the Supervised Setting}

By default, training followed the settings specified in Methods (Section \ref{sec:methods}).  Here, we discuss implementation details for \textit{a}) re-training a classifier without background augmentations and \textit{b}) finetuning the whole network without background augmentations and \textit{c}) longer training.

\subsubsection{Re-training a linear classifier w/o background augmentations}
We trained a linear classifier from scratch on top of the frozen trunk \textit{without} background augmentations using standard pre-processing and data augmentation. We used SGD with momentum of 0.9 , a batchsize of 4096 with a base learning rate of 0.01 and a cosine schedule over 40 epochs and a weight decay of $1\times10^{-4}$. 

\subsubsection{Finetuning w/o background augmentations}
We finetuned the network \textit{without} background augmentations, using standard preprocessing and data augmentation. We used SGD with momentum of 0.9 , a batchsize of 4096 with a base learning rate of 0.0005 and a cosine schedule  over 20 epochs and a weight decay of $1\times10^{-4}$.

\subsubsection{Longer Training}
Training followed the settings specified in Methods (section \ref{sec:methods}), with the following changes: following CutMix \citep{yun_cutmix_2019}, training was for 300 epochs with a step schedule of (75, 150, 225).

\subsection{Evaluation in Limited-Labels Setting}

For consistency with previous work, we use the same fixed splits\footnote{\url{https://github.com/google-research/simclr/tree/master/imagenet_subsets}} as in \citet{chen2020big} for the 1\% and 10\% labels settings of ImageNet training data. 

\subsubsection{Linear Evaluation} For simplicity, we train a linear classifier using the same settings as in the corresponding 100\% labels linear evaluation, except that we use the training data from the corresponding split (1\% or 10\%). 

\subsubsection{Semi-Supervised Evaluation}
\moco/BYOL: We finetuned the network for 50 epochs with a step schedule (30, 40) with a decay factor of 0.1, with a batch size of 256 and no weight decay. For \moco, in the 1\% (10\%) label setting, we used a learning rate of 1.0 (0.3) for the classifier head and a learning rate of $1\times10^{-4}$ ($1\times10^{-4}$) for the trunk. For BYOL, in the 1\% (10\%) label setting, we used a learning rate of 1.0 (0.1) for the classifier head and a learning rate of $1\times10^{-3}$ ($1\times10^{-3}$) for the trunk.
SwAV: we use the same settings as in the original implementation \citep{swav}.

\begin{figure}
    \centering
    \includegraphics[width=0.65\textwidth]{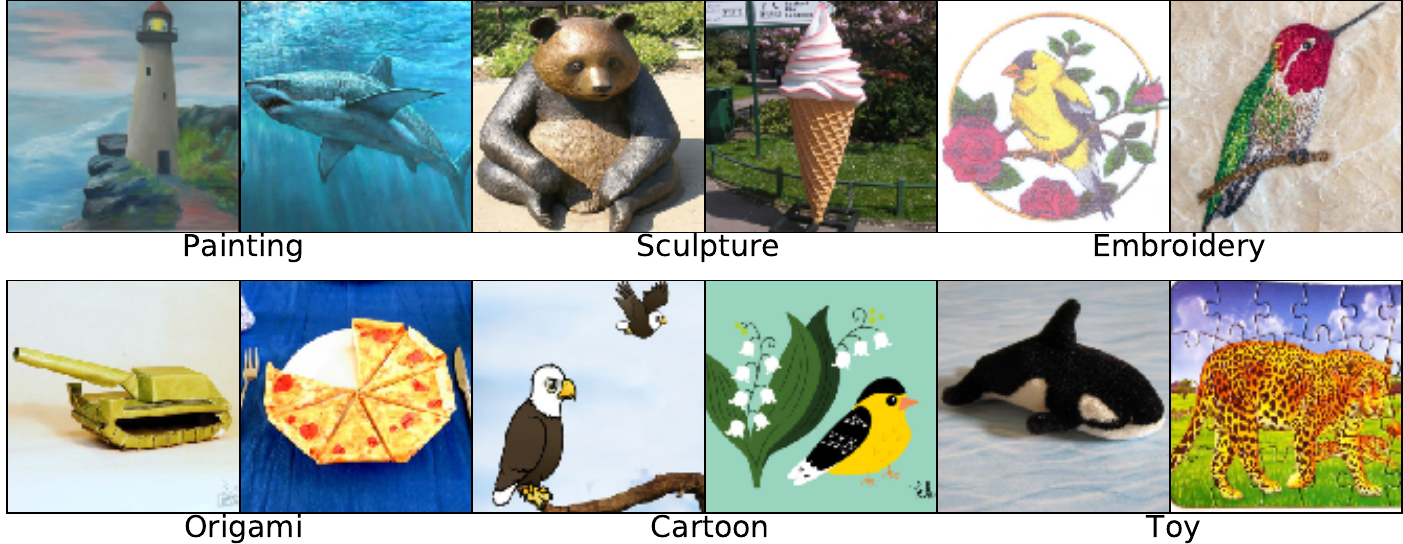}
    \caption{Examples from ImageNet-Renditions. Image from \citet{hendrycks_many_2021}.}
    \label{fig:imnet_r_examples}
\end{figure}

\subsection{Robustness Evaluations}
Robustness evaluations do not involve any additional training---they use the same network (backbone and linear classifier) used for linear evaluation on ImageNet in the 100\% labels setting. Note that it is common for robustness benchmarks to use the pre-trained model from torchvision\footnote{\url{https://github.com/pytorch/vision/tree/master/torchvision/models}} as the supervised baseline. We additionally report metrics using our re-implementation of the standard, stronger supervised baseline in \citet{goyal_accurate_2018}. We follow the pre-processing protocols from respective original works; unless otherwise mentioned, this is simply the standard way that ImageNet validation images are pre-processed (Appendix \ref{app:imagenet_lin_eval}).

\paragraph{ImageNet-9.}
 ImageNet-9 \citep{xiao2020noise} (IN-9) consists of data sets with varying amount of foreground-background signal. Variations of IN-9 (excluding the “Original") involve a distribution shift in foreground-background statistics. We use the data and code \footnote{\url{https://github.com/MadryLab/backgrounds_challenge}} from the original work for evaluation.
 
 \paragraph{ImageNet-ReaL.} \citet{beyer2020imagenetreal} relabel the ImageNet validation images to better represent the semantic content present in the images; we use the Reassessed Labels\footnote{\url{https://github.com/google-research/reassessed-imagenet}} (ReaL) to evaluate ImageNet-ReaL accuracy. Our supervised (torchvision) baseline has an ImageNet-ReaL accuracy of 82.7\% (82.9\%).

\paragraph{ImageNet-v2.} ImageNet-v2 \citep{imagenetv2} consists of three test data sets for ImageNet, with 10,000 images each. The three variations are \textit{a}) MatchedFrequency, \textit{b}) Threshold0.7 and \textit{c}) TopImages. Accuracies reported in the main text (Section \ref{sec:imnet_v2_and_objectnet}) correspond to MatchedFrequency. Accuracies for other variations are reported in Appendix \ref{app:additional_results_imnet_v2}. 

\paragraph{ObjectNet.} ObjectNet \citep{objectnet} is a test data set that controls for rotation, background, and viewpoint. It contains 50,000 images with 313 object classes. 113 of ObjectNet's classes overlap with ImageNet---we evaluate on this subset\footnote{We used code from \url{https://github.com/lucaslie/torchprune} to map images to ImageNet classes.}. Following the original work\footnote{\url{https://github.com/abarbu/objectnet-template-pytorch}}, after removing the one-pixel red border, images are resized to size 224 along the shorter edge, followed by  pre-processing steps of \texttt{CenterCrop} and normalization as in Appendix \ref{app:imagenet_lin_eval}). Our supervised (torchvision) baseline has an ObjectNet accuracy of 24.4\% (24.7\%).

\paragraph{ImageNet-A.} ImageNet-A \citep{hendrycks2019nae} is a data set of “natural" adversarial examples---images obtained by a process of adversarial filtering of natural images. It consists of 7,500 images mis-classified by the torchvision ResNet-50 pre-trained model\footnote{See \citet{hendrycks2019nae} for additional filtering criteria used.}. \citet{hendrycks2019nae} report that a corresponding model re-trained from scratch has an accuracy of 2.2\%; our supervised baseline has an accuracy of 2.4\%.

\paragraph{ImageNet-R.} ImageNet-R is a data set curated to measure generalization to various abstract visual  renditions (e.g. paintings, embroidery, cartoons etc., see Figure \ref{fig:imnet_r_examples} for examples) of  ImageNet classes. ImageNet-R involves a shift in texture statistics and contains 30,000 images. Our supervised (torchvision) baseline has an ImageNet-R accuracy of 36.0\% (36.1\%).

\paragraph{Adversarial Attack.}
We used foolbox \citep{foolbox_rauber2017} for $\ell_{\infty}$ FGSM adversarial attack with $\epsilon=8/255$ applied to ImageNet validation images.

\paragraph{ImageNet-C.} ImageNet-C \citep{hendrycks_benchmarking_2019} consists of 75 test data sets; 15 types of corruptions from four main categories (noise, blur, weather, digital) are applied to ImageNet validation images to generate the test images. Each corruption type has five levels of severity. We report the average performance on the four main categories of corruptions in Appendix \ref{app:additional_results_imnet_c}. Pre-processing steps are \texttt{CenterCrop} and normalization as in Appendix \ref{app:imagenet_lin_eval}.

\begin{figure}
    \centering
    \includegraphics[width=0.65\textwidth]{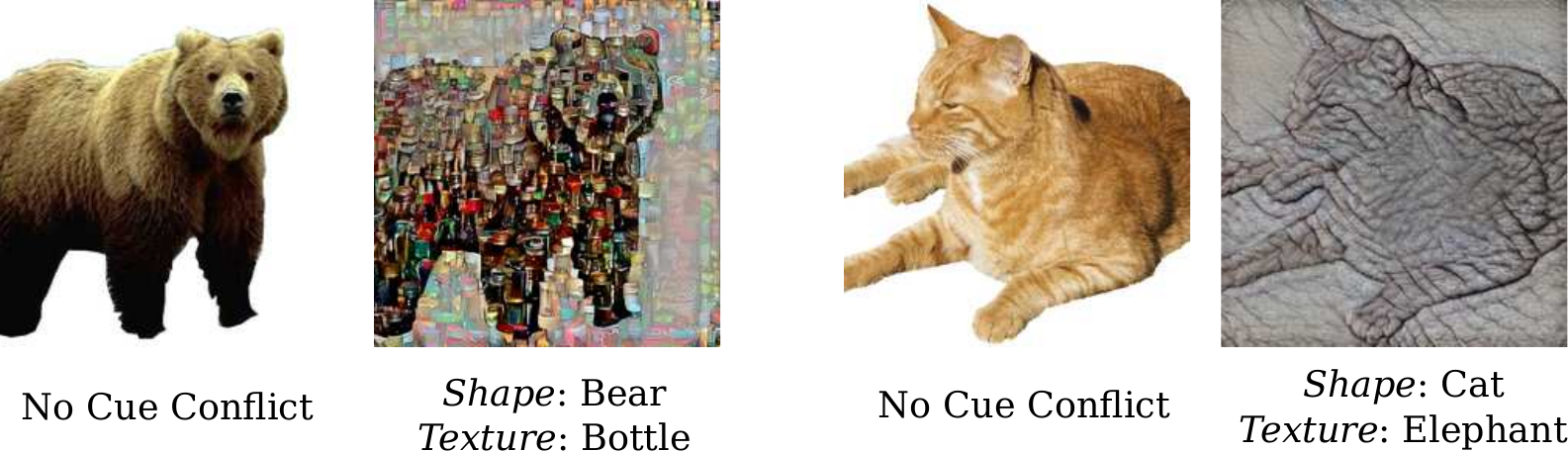}
    \caption{Examples with and without texture-shape cue conflict.}
    \label{fig:cue_conflict_stim}
\end{figure}

\subsection{Shape-Bias Evaluation}
The shape-bias \citep{geirhos2019imagenettrained} of a model is computed using texture-shape cue conflict stimuli\footnote{We use the cue conflict stimuli released at \url{https://github.com/rgeirhos/texture-vs-shape}.} (the shape and texture cues in the image correspond to different ImageNet classes, e.g. see Figure \ref{fig:cue_conflict_stim}) as the fraction of classification decisions that correspond to shape information; this computation only considers the subset of “correctly" classified images---either the shape or texture category are correctly classified. Images are pre-processed as in Appendix \ref{app:imagenet_lin_eval}.

\subsection{Linear Evaluation on CIFAR-10, 100}
All methods were evaluated using the same protocol for fair comparison. A linear classifier was trained with SGD with momentum 0.9 for 100 epochs with a base learning rate of 0.08, a batch size of 32 and a cosine learning rate schedule.

\subsection{Linear Evaluation on Places-205}
All methods were evaluated using the same protocol for fair comparison. A linear classifier was trained with SGD with momentum 0.9 for 28 epochs with a base learning rate of 1.0, a batch size of 256 and a step schedule of 7, 14, 21 and a decay factor of 0.1. Weight decay was set to 0.0001.

\subsection{Object Detection and Instance Segmentation}

We follow standard protocol across all SSL methods: \textit{a}) VOC detection: We finetuned a Faster R-CNN \citep{ren_faster_2015} in VOC 2007 + 2012 trainval for 24k iterations and evaluated in VOC 2007 test, \textit{b}) COCO detection and COCO instance segmentation: We fine-tuned a Mask R-CNN \citep{he_mask_2017} (2× schedule) in COCO 2017 train, evaluated in COCO 2017 val. All Faster/Mask R-CNN models are with the C4-backbone. We use \texttt{Detectron2} \citep{wu2019detectron2} for all experiments\footnote{We use the code provided at \url{https://github.com/facebookresearch/moco/tree/main/detection}.}.

VOC: For \moco~and SwAV, we followed the settings from their corresponding original papers. In the case of BYOL, we followed the settings of \moco~with one deviation: we used a base learning rate of $0.08$. COCO: we used the default settings of \moco~for all methods.

\section{Characterizing the Impact of Mask Quality} 
\label{app:noisy_masks}
While the results in Section \ref{sec:u2net_mask_imnet_lin_eval} show that there may be diminishing gains to using \textit{higher} quality masks than those provided by DeepUSPS$^2$, it does not give us a clear picture of the impact of mask quality. For example, one may wonder,
\begin{itemize}
    \item How does performance vary as a function of mask quality?
    \item Which SSL methods and background augmentations are more robust to mask quality?
\end{itemize}

To answer these questions, and in an attempt to gain further insight into the mechanism by which background augmentations improve representations, we systematically perturb the quality of the foreground masks. In these experiments, we use U$^2$Net to generate the initial foreground masks (that we then perturb) to minimize the possibility of starting with a poor mask and maintain greater control over the quality of a perturbed mask. 

\begin{figure}
    \centering
    \includegraphics[width=0.7\textwidth]{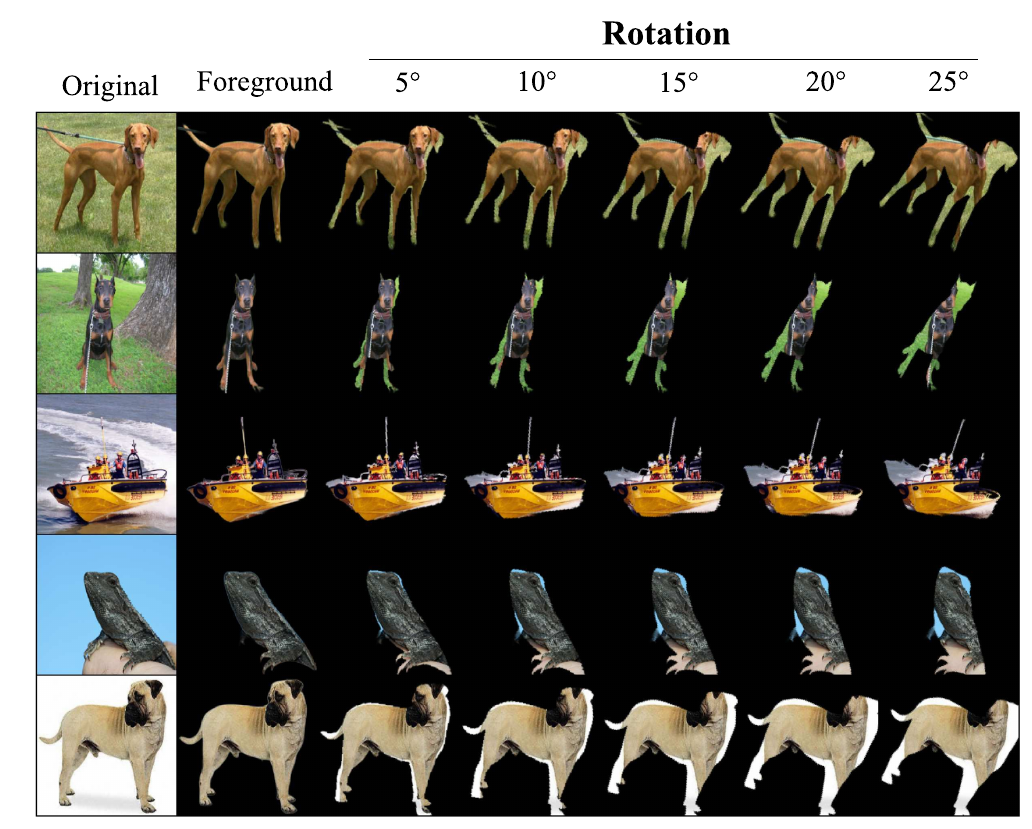}
    \caption{{\bf Mask Distortion: Rotation}. Examples of varying distortion strength.}
    \label{fig: rotation}
\end{figure}

\begin{figure}
    \centering
    \includegraphics[width=0.7\textwidth]{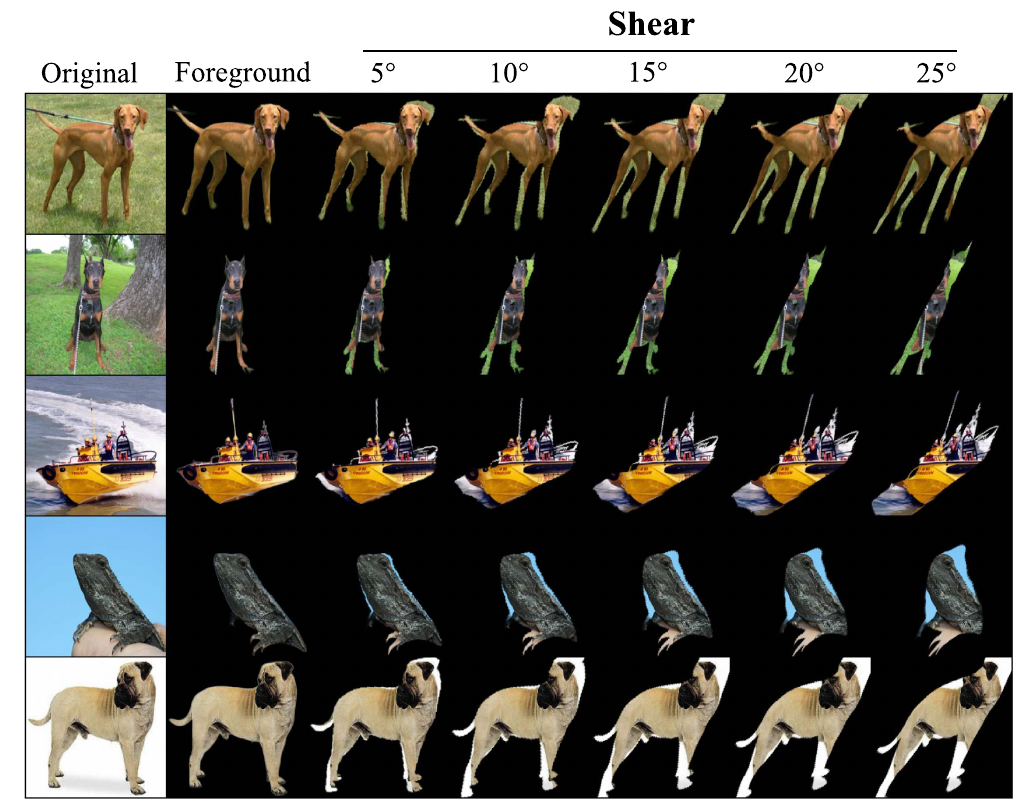}
    \caption{{\bf Mask Distortion: Shearing}. Examples of varying distortion strength.}
    \label{fig: shear}
\end{figure}

\begin{figure}
    \centering
    \includegraphics[width=0.7\textwidth]{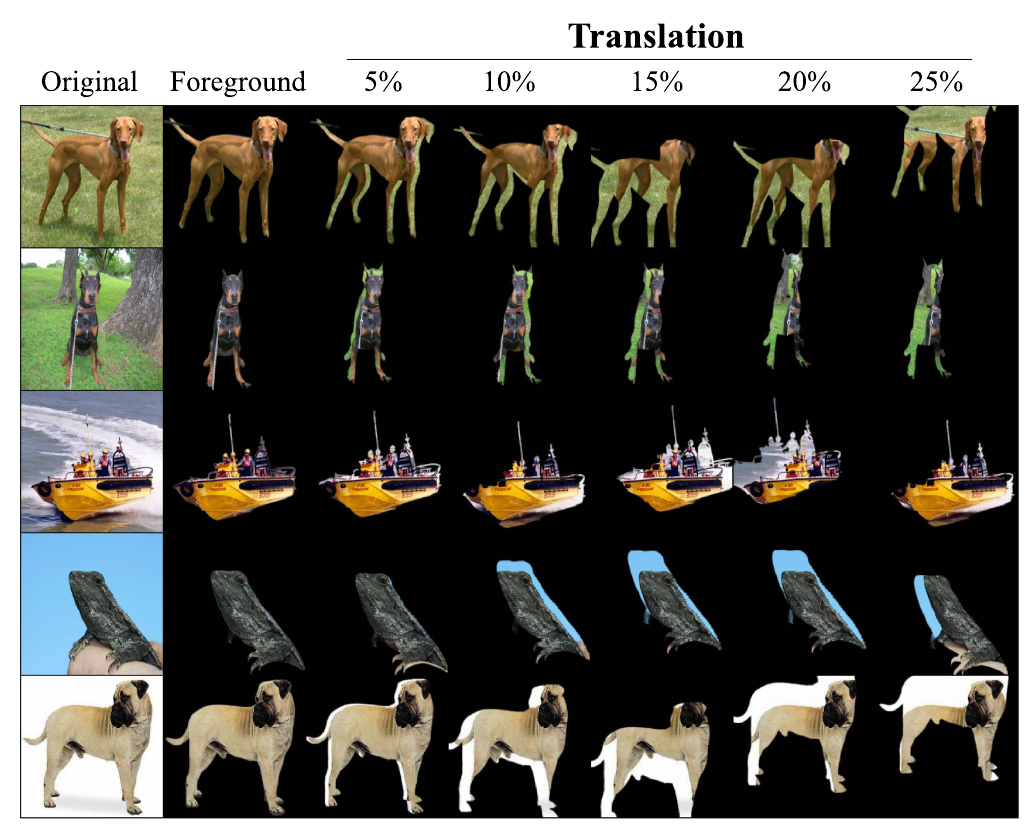}
    \caption{{\bf Mask Distortion: Translation}. Examples of varying distortion strength.}
    \label{fig: traslation}
\end{figure}

\begin{figure}
    \centering
    \includegraphics[width=0.5\textwidth]{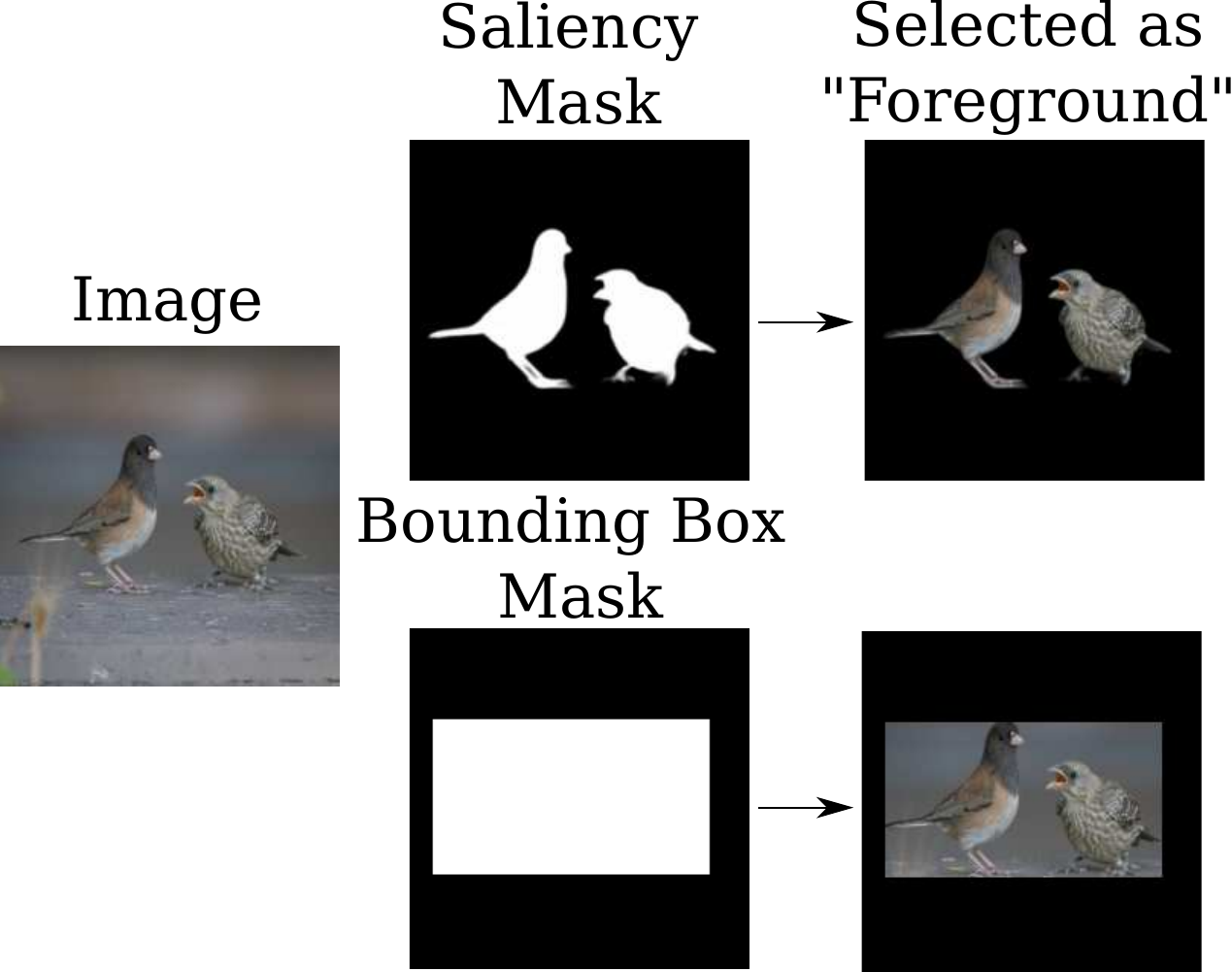}
    \caption{{\bf Mask Distortion: Bounding-Box Mask}. An example of a bounding-box mask.}
    \label{fig: bbox_mask}
\end{figure}

 We perturb the masks in numerous ways across a range of distortion strengths/levels. The distortions we consider are: \textit{a}) \textit{rotation} (see Figure \ref{fig: rotation} for examples), \textit{b}) \textit{shearing} (Figure \ref{fig: shear}), \textit{c}) \textit{translation} (Figure \ref{fig: traslation}), \textit{d}) \textit{horizontal flips} and \textit{e}) using \textit{bounding-box masks} instead of the original mask (Figure \ref{fig: bbox_mask}). We expect mask translation and using bounding-box masks to be particularly challenging. 
 
\begin{table}
    \centering
    \resizebox{1\columnwidth}{!}{
    \begin{tabular}{ccccccccc}\toprule
      & \multicolumn{3}{c}{\textbf{\moco}}  & \multicolumn{2}{c}{\textbf{BYOL}} & \multicolumn{2}{c}{\textbf{SwAV}} \\
    \cmidrule(lr){2-4} \cmidrule(lr){5-6} \cmidrule(lr){7-8} 
 \shortstack{Max \\ \textbf{Rotation} } &   baseline & \bgrm & \bgswaps & baseline & \bgrm & baseline & \bgrm  \\ \midrule
  0$^{\circ}$ & 67.7 & 69.3 (\textbf{\textcolor{ForestGreen}{+1.6}}) & 69.7 (\textbf{\textcolor{ForestGreen}{+2.0}}) & 72.7 & 73.5 (\textbf{\textcolor{ForestGreen}{+0.8}}) & 72.2 & 73.7 (\textbf{\textcolor{ForestGreen}{+1.5}}) \\ 
  5$^{\circ}$ & - & 69.3 (\textbf{\textcolor{ForestGreen}{+1.6}})   & 69.3  (\textbf{\textcolor{ForestGreen}{+1.6}}) & -  & 73.1  (\textbf{\textcolor{ForestGreen}{+0.4}}) & - & 73.7 (\textbf{\textcolor{ForestGreen}{+1.5}})    \\
  10$^{\circ}$ & - & 69.0 (\textbf{\textcolor{ForestGreen}{+1.3}})   & 69.3 (\textbf{\textcolor{ForestGreen}{+1.6}}) &  -   & 73.3  (\textbf{\textcolor{ForestGreen}{+0.6}}) &-& 73.6 (\textbf{\textcolor{ForestGreen}{+1.4}})   \\
  15$^{\circ}$ & - & 68.7  (\textbf{\textcolor{ForestGreen}{+1.0}})  & 69.4 (\textbf{\textcolor{ForestGreen}{+1.7}}) &  -   & 73.3  (\textbf{\textcolor{ForestGreen}{+0.6}}) &-&  73.3 (\textbf{\textcolor{ForestGreen}{+1.1}})    \\
  20$^{\circ}$ & - & 68.7 (\textbf{\textcolor{ForestGreen}{+1.0}})  & 69.1 (\textbf{\textcolor{ForestGreen}{+1.4}}) &  -   & 73.1  (\textbf{\textcolor{ForestGreen}{+0.4}})  & -& 73.5 (\textbf{\textcolor{ForestGreen}{+1.3}})    \\
  25$^{\circ}$ & - & 68.4  (\textbf{\textcolor{ForestGreen}{+0.7}})  & 69.0 (\textbf{\textcolor{ForestGreen}{+1.3}}) &  -   & 73.1  (\textbf{\textcolor{ForestGreen}{+0.4}})  &- & 73.5 (\textbf{\textcolor{ForestGreen}{+1.3}})  \\\bottomrule
    \end{tabular}
    }
    \caption{ {\bf Mask Distortion: Rotation. Background augmentations are robust to substantial mask noise induced via \textit{rotations} of the foreground mask.} Of note, \bgswaps~maintains a large performance benefit even for strong distortions. We highlight the $\Delta$ relative to the baselines with no background augmentations.}
    \label{tab: mask_rotation}
\end{table} 

 We characterize the dependence on mask quality for \bgrm~across all the view-invariant SSL methods in our test bed; in the case of \moco, we additionally include \bgswaps. We apply the respective distortion to \textit{each mask}, \textit{every time} it is used in a background augmentation. All experiments are in the respective medium duration settings; we report ImageNet accuracy. We make the following observations:

\paragraph{Low quality masks can still be beneficial.} Overall, we find that there is \textit{substantial} robustness to mask quality. In many instances, only a little of the foreground remains in view, yet background augmentations maintain improved performance over the baseline. We find that mask translation and using bounding-box masks are particularly challenging distortions as expected---surprisingly, some performance benefits persist (but quickly disappear with higher distortion strength). 

\paragraph{\swav~and \bgswaps~are quite robust to mask quality.} 
We find that \bgswaps~is far more robust to mask quality compared to \bgrm~across all variants and degrees of distortions, showcasing the robustness of this augmentation method. For example, across rotation, shearing, translation and flip distortions, the improvement due to \bgswaps~is $\sim$2-3$\times$ the improvement due to \bgrm~in the respective strongest distortion levels; further, when the foreground mask is replaced with a bounding-box mask, the benefit due to \bgrm~disappears, while \bgswaps~retains a significant performance benefit.

Among the SSL methods in our test bed, \swav~appears to be most robust to mask quality, managing to maintain significant performance benefits even with strong mask distortion. We speculate that this may be linked to use of  \texttt{multi-crop} augmentation wherein local crops are expected to be predictive of global crops. When background augmentations are applied using distorted masks, only a small part of the foreground may be featured in a view---much like a local crop in \texttt{multi-crop} augmentation.

\begin{table}
    \centering
        \resizebox{1\columnwidth}{!}{
    \begin{tabular}{ccccccccc}\toprule
      & \multicolumn{3}{c}{\textbf{\moco}}  & \multicolumn{2}{c}{\textbf{BYOL}} & \multicolumn{2}{c}{\textbf{SwAV}} \\
    \cmidrule(lr){2-4} \cmidrule(lr){5-6} \cmidrule(lr){7-8} 
 \shortstack{Max \\ \textbf{Shear} } &   baseline & \bgrm & \bgswaps & baseline & \bgrm & baseline & \bgrm  \\ \midrule
  0$^{\circ}$ & 67.7 & 69.3 (\textbf{\textcolor{ForestGreen}{+1.6}}) & 69.7 (\textbf{\textcolor{ForestGreen}{+2.0}}) & 72.7 & 73.5 (\textbf{\textcolor{ForestGreen}{+0.8}}) & 72.2 & 73.7 (\textbf{\textcolor{ForestGreen}{+1.5}}) \\
  5$^{\circ}$ & - & 69.2 (\textbf{\textcolor{ForestGreen}{+1.5}}) & 69.3 (\textbf{\textcolor{ForestGreen}{+1.6}})  & -& 73.2 (\textbf{\textcolor{ForestGreen}{+0.5}})  &-& 73.5  (\textbf{\textcolor{ForestGreen}{+1.3}})    \\
  10$^{\circ}$ & - & 68.8 (\textbf{\textcolor{ForestGreen}{+1.1}}) & 69.2 (\textbf{\textcolor{ForestGreen}{+1.5}})  & - & 73.5 (\textbf{\textcolor{ForestGreen}{+0.8}})&-& 73.6 (\textbf{\textcolor{ForestGreen}{+1.4}})  \\
  15$^{\circ}$ & - & 68.6 (\textbf{\textcolor{ForestGreen}{+0.9}}) & 69.4 (\textbf{\textcolor{ForestGreen}{+1.7}}) & - &73.0 (\textbf{\textcolor{ForestGreen}{+0.3}})&-& 73.5 (\textbf{\textcolor{ForestGreen}{+1.3}})   \\
  20$^{\circ}$ & - & 68.3 (\textbf{\textcolor{ForestGreen}{+0.6}}) & 69.0 (\textbf{\textcolor{ForestGreen}{+1.3}})  & - &73.2 (\textbf{\textcolor{ForestGreen}{+0.5}})&-& 73.4 (\textbf{\textcolor{ForestGreen}{+1.2}}) \\
  25$^{\circ}$ & - & 68.1 (\textbf{\textcolor{ForestGreen}{+0.4}}) & 68.9 (\textbf{\textcolor{ForestGreen}{+1.2}}) & - &72.9 (\textbf{\textcolor{ForestGreen}{+0.2}})&-& 73.5 (\textbf{\textcolor{ForestGreen}{+1.3}})   \\\bottomrule
    \end{tabular}
    }
    \caption{ {\bf Mask Distortion: Shearing. Background augmentations are robust to substantial mask noise induced via \textit{shearing} of the foreground mask.} Of note, \bgswaps~maintains a large performance benefit even for strong distortions. We highlight the $\Delta$ relative to the baselines with no background augmentations.}
    \label{tab: mask_shear}
\end{table}

\subsection{Implementation Details.} Every mask was perturbed prior to background augmentations in every epoch. {\it Rotation.} Each mask was subject to random rotation sampled between $(-\text{max rot.},+\text{max rot.})$. {\it Shearing.} Each mask was subject to shearing independently along $x$ and $y$ coordinates with a value uniformly sampled from $(-\text{max shear},+\text{max shear})$. 
{\it Translation.} Each mask was subject to translation independently along the width and height with values uniformly sampled from $(-\text{max trans.},+\text{max trans.}) \times \text{max FG width}$ and $(-\text{max trans.},+\text{max trans.}) \times \text{max FG height}$ respectively. {\it Horizontal Flip.} Each mask was horizontally flipped with a probability 0.5 (on each use of the mask). {\it Bounding-Box Masks.} Binary foreground masks were used to generate rectangular bounding-box masks of size max FG width $\times$ max FG height.

\begin{table}
    \centering
    \resizebox{1\columnwidth}{!}{
    \begin{tabular}{ccccccccc}\toprule
      & \multicolumn{3}{c}{\textbf{\moco}}  & \multicolumn{2}{c}{\textbf{BYOL}} & \multicolumn{2}{c}{\textbf{SwAV}} \\
    \cmidrule(lr){2-4} \cmidrule(lr){5-6} \cmidrule(lr){7-8} 
 \shortstack{Max \\ \textbf{Translation} } &   baseline & \bgrm & \bgswaps & baseline & \bgrm & baseline & \bgrm  \\ \midrule
  0\% &  67.7 & 69.3 (\textbf{\textcolor{ForestGreen}{+1.6}}) & 69.7 (\textbf{\textcolor{ForestGreen}{+2.0}}) & 72.7 & 73.5 (\textbf{\textcolor{ForestGreen}{+0.8}}) & 72.2 & 73.7 (\textbf{\textcolor{ForestGreen}{+1.5}}) \\ 
  5\%  & -& 68.9 (\textbf{\textcolor{ForestGreen}{+1.2}}) & 69.5 (\textbf{\textcolor{ForestGreen}{+1.8}}) &  -   & 73.3 (\textbf{\textcolor{ForestGreen}{+0.6}}) & -& 73.3  (\textbf{\textcolor{ForestGreen}{+1.1}})   \\
  10\%  & -& 68.7 (\textbf{\textcolor{ForestGreen}{+1.0}})   & 69.2 (\textbf{\textcolor{ForestGreen}{+1.5}})  &  -& 73.3 (\textbf{\textcolor{ForestGreen}{+0.6}}) &-&73.5 (\textbf{\textcolor{ForestGreen}{+1.3}})  \\
  15\%  & -& 68.2 (\textbf{\textcolor{ForestGreen}{+0.5}}) & 68.7 (\textbf{\textcolor{ForestGreen}{+1.0}}) & - & 73.1 (\textbf{\textcolor{ForestGreen}{+0.4}}) &-&73.2 (\textbf{\textcolor{ForestGreen}{+1.0}})   \\
  20\%  & -& 68.0 (\textbf{\textcolor{ForestGreen}{+0.3}}) & 68.7 (\textbf{\textcolor{ForestGreen}{+1.0}}) &-  & 73.1 (\textbf{\textcolor{ForestGreen}{+0.4}}) &-& 73.3 (\textbf{\textcolor{ForestGreen}{+1.1}}) \\
  25\%  &- &67.9 (\textbf{\textcolor{ForestGreen}{+0.2}}) & 68.3 (\textbf{\textcolor{ForestGreen}{+0.6}}) & - & 72.9 (\textbf{\textcolor{ForestGreen}{+0.2}}) &-& 73.0 (\textbf{\textcolor{ForestGreen}{+0.8}})  \\\bottomrule
    \end{tabular}
    }
    \caption{ {\bf Mask Distortion: Translation. Background augmentations are robust to substantial mask noise induced via \textit{translation} of the foreground mask.} Of note, \bgswaps~maintains a large performance benefit even for strong distortions. We highlight the $\Delta$ relative to the baselines with no background augmentations.}
    \label{tab: mask_translate}
\end{table}

\begin{table}
    \centering
    \resizebox{1\columnwidth}{!}{
    \begin{tabular}{ccccccccc}\toprule
      & \multicolumn{3}{c}{\textbf{\moco}}  & \multicolumn{2}{c}{\textbf{BYOL}} & \multicolumn{2}{c}{\textbf{SwAV}} \\
    \cmidrule(lr){2-4} \cmidrule(lr){5-6} \cmidrule(lr){7-8} 
 \shortstack{Random \\ \textbf{Horizontal Flip} } &   baseline & \bgrm & \bgswaps & baseline & \bgrm & baseline & \bgrm  \\ \midrule
    \xmark    & 67.7 & 69.3 (\textbf{\textcolor{ForestGreen}{+1.6}}) & 69.7 (\textbf{\textcolor{ForestGreen}{+2.0}}) & 72.7 & 73.5 (\textbf{\textcolor{ForestGreen}{+0.8}}) & 72.2 & 73.7 (\textbf{\textcolor{ForestGreen}{+1.5}}) \\
  \checkmark & - & 68.7 (\textbf{\textcolor{ForestGreen}{+1.0}}) & 69.4 (\textbf{\textcolor{ForestGreen}{+1.7}})  & - & 73.3 (\textbf{\textcolor{ForestGreen}{+0.6}})  & - & 73.5  (\textbf{\textcolor{ForestGreen}{+1.3}}) \\\bottomrule
    \end{tabular}
    }
    \caption{ {\bf Mask Distortion: Horizontal Flip. Background augmentations are robust to substantial mask noise induced via \textit{horizontal flip} of the foreground mask.} Of note, \bgswaps~maintains a large performance benefit even for strong distortions. We highlight the $\Delta$ relative to the baselines with no background augmentations.}
    \label{tab: mask_hflip}
\end{table}

\begin{table}
    \centering
    \resizebox{1\columnwidth}{!}{
    \begin{tabular}{ccccccccc}\toprule
      & \multicolumn{3}{c}{\textbf{\moco}}  & \multicolumn{2}{c}{\textbf{BYOL}} & \multicolumn{2}{c}{\textbf{SwAV}} \\
    \cmidrule(lr){2-4} \cmidrule(lr){5-6} \cmidrule(lr){7-8} 
 &   baseline & \bgrm & \bgswaps & baseline & \bgrm & baseline & \bgrm  \\ \midrule
    Saliency mask & 67.7 & 69.3 (\textbf{\textcolor{ForestGreen}{+1.6}}) & 69.7 (\textbf{\textcolor{ForestGreen}{+2.0}}) & 72.7 & 73.5 (\textbf{\textcolor{ForestGreen}{+0.8}}) & 72.2 & 73.7 (\textbf{\textcolor{ForestGreen}{+1.5}}) \\
  \textbf{Bounding-box mask} & - & 67.8 (\textbf{\textcolor{ForestGreen}{+0.1}}) & 68.7 (\textbf{\textcolor{ForestGreen}{+1.0}})  & - & 73.0 (\textbf{\textcolor{ForestGreen}{+0.3}})  & - & 73.0  (\textbf{\textcolor{ForestGreen}{+0.8}}) \\\bottomrule
    \end{tabular}
    }
    \caption{ {\bf Mask Distortion: Bounding-Box Mask. \bgswaps~is robust even to replacing the mask with a bounding-box mask.} Notably, SwAV also manages to retain some performance benefit even in this extreme setting.}
    \label{tab: mask_bbox}
\end{table}

\clearpage
\section{Ablations} 
\label{app:ablations}
Here, we report additional ablations of design choices and hyperparameter settings. We also provide additional information regarding ablations in the main text. To control for mask quality, we use U$^2$Net to generate foreground masks unless otherwise indicated. We report ImageNet accuracy. 

\subsection{Augmentation Strength}
\label{app:aug_strength_ablation}
We ablate the strength of background augmentations and find that $p\in[0.1,0.3]$ is generally a good setting.
For \moco, as previously discussed, though \bgrm~is more (Out-of-Distribution) OOD than \bgrand, the presence of negatives in the queue $Q$ with (gray) backgrounds similar to $q$ offsets this and results in better performance than \bgrand~across all augmentation strengths, see Table \ref{tab:moco_bg_aug_strength_A}.

\begin{table}
    \centering
    \begin{tabular}{lccccc}\toprule
         Baseline (67.7) & $p=0.1$ & $p=0.2$ & $p=0.3$ &  $p=0.4$ & $p=0.5$ \\\midrule
         \bgrm &  \textbf{69.3} & \textbf{69.3} & \textbf{69.3} & 68.8 & 68.4 \\ 
         \bgrand &  69.1 & \textbf{69.2} & 68.6 & 67.8 & 66.8\\\bottomrule
    \end{tabular}
    \caption{\textbf{\moco}: Ablating the strength of background augmentations \bgrm~and \bgrand.}
    \label{tab:moco_bg_aug_strength_A}
\end{table}

\begin{table}
    \centering
    \begin{tabular}{lccccc}\toprule
         Baseline (67.7) & $p_{\text{neg}}=0.1$ & $p_{\text{neg}}=0.2$ & $p_{\text{neg}}=0.3$ &  $p_{\text{neg}}=0.4$ & $p_{\text{neg}}=0.5$ \\\midrule
         $p_{\text{pos}}=0.1$ &  69.2 & \textbf{69.6} &\textbf{ 69.7} & \textbf{69.6} & \textbf{69.7}  \\ 
        $p_{\text{pos}}=0.2$ &  69.5 & \textbf{69.7} & 69.5 & 69.5 & \textbf{69.7} \\
        $p_{\text{pos}}=0.3$ &  68.9 & 69.3 & 69.3 & 69.0 & 69.3 \\ \bottomrule
    \end{tabular}
    \caption{\textbf{\moco}: Ablating the strength of background augmentation \bgswaps.}
    \label{tab: moco_bg_aug_strength_B}
\end{table}

As discussed in Section \ref{sec:bgswap_ablations}, \bgswaps~overcomes the OOD issue of \bgrm~by using cached random natural backgrounds in $q$ and $k^+$ with a probability $p_{\text{pos}}$ and additionally includes an extra negative whose background matches $q$ with a probability $p_{\text{neg}}$. As can be seen in Table \ref{tab: moco_bg_aug_strength_B}, this results in substantially improved performance at each level of $p_{\text{pos}}$ over the corresponding level of \bgrand. Performance is more sensitive to $p_{\text{pos}}$ than to $p_{\text{neg}}$ and \ppos$\simeq$\pneg~is generally a reasonable setting. Performance as a function of augmentation strength for BYOL and SwAV are shown in Tables \ref{tab: byol_aug_strength} and \ref{tab: swav_aug_strength}. In all of the above ablations, $\text{FG}_{\text{min}}=0$, i.e. no constraints were imposed on \texttt{RRC} regarding the foreground.

\subsection{Do Multiple Matched Negatives Help?}
\label{sec: multi_neg}
We investigated using multiple background matched negatives in \bgswaps~but found that it did not confer further improvements in performance. We obtained 68.9\% ImageNet accuracy using 5 matched negatives \textit{vs.} 68.8\% using 1 matched negative, suggesting that there may be little benefit to increasing the number of background matched negatives. In these experiments, there were no background augmentations in $q$, only in $k^+$ and $k^-$; $p_{\text{pos}}=p_{\text{neg}}=0.2$ and background augmented negatives matched the background of $k^+$. 

\subsection{Is it a Better Teaching Signal for Background Augmentations in the Positive and Negative to be Independent or Coupled? }
To answer this question, we contrasted \textit{independent} background augmentations in \bgswaps~in $k^+$ and $k^-$ with \textit{coupled }background augmentations in $k^+$ and $k^-$. We observed identical performance of 68.8\% in each case. Other experiment settings as in \ref{sec: multi_neg}.

\subsection{Is it Better for a Negative's Background to Match \texorpdfstring{$q$}{q} or \texorpdfstring{$k^+$}{k+}?}

We observed similar results in both cases \textit{a}) background matches $q$ (68.9\%) and \textit{b}) background matches $k^+$ (68.8\%). Other experiment settings as in \ref{sec: multi_neg}.

\subsection{Order of Augmentations}
\label{app:aug_order}
As noted in Section \ref{sec:bg_augs_methods}, by default, we apply background augmentations \textit{after} all other augmentations in the respective pipeline. Here, we show that background augmentation \textit{before} all other augmentations produces similar results. We apply \bgrand~in \swav~with $p=0.1$ and find corresponding accuracies of 73.4\% (\textit{before}) \textit{vs}. 73.2\% (\textit{after}). Similarly, applying \bgrm~with $p=0.1$~in \moco, we find corresponding accuracies of 69.6\%  (\textit{before}) \textit{vs}. 69.3\% (\textit{after}).

\begin{table}
    \centering
    \begin{tabular}{lccccc}\toprule
         Baseline (72.7) & $p=0.05$ & $p=0.1$ & $p=0.15$ &  $p=0.2$ & $p=0.25$ \\\midrule
         \bgrm &  73.7 & 73.6 & \textbf{73.8} & 73.5 & 73.1 \\ 
         \bgrand & 73.7 & \textbf{73.9} & 73.5 & 73.4 & 72.7 \\ \bottomrule
    \end{tabular}
    \caption{\textbf{BYOL}: Ablating the strength of background augmentations \bgrm~and \bgrand.}
    \label{tab: byol_aug_strength}
\end{table}

\begin{table}
    \centering
    \begin{tabular}{lcccccc}\toprule
         Baseline (72.2) & $p=0.05$ & $p=0.1$ & $p=0.15$ & $p=0.2$ &  $p=0.25$ & $p=0.3$ \\\midrule
         \bgrm & 73.1 & 73.5 & 73.5 & 73.5 & \textbf{73.6} & \textbf{73.7} \\ 
         \bgrand &  73.0 & 73.2 & 73.2 & \textbf{73.7 }& 70.9 & 67.2 \\ \bottomrule
    \end{tabular}
    \caption{\textbf{SwAV}: Ablating the strength of background augmentations \bgrm~and \bgrand.}
    \label{tab: swav_aug_strength}
\end{table}

\subsection{Influence of Random Crop}
\label{app:rand_crops}

\texttt{RandomResizedCrop} (\texttt{RRC}) is a critical part of current SSL pipelines. Indeed, replacing \texttt{RRC} with \texttt{CenterCrop} results in very poor accuracy, 26.8\% for \moco, see Table \ref{tab:no_rrc} (\rownumber{b}). Concretely, instead of \texttt{RRC}, we first \texttt{Resize} to size 256 along the shorter edge of the image, followed by a \texttt{CenterCrop} to size 224$\times$224. Interestingly, using \bgswaps~with \texttt{CenterCrop} can substantially improve performance (\rownumber{c}) over \texttt{CenterCrop} alone, though the performance is still far reduced from using \texttt{RRC}. This raises an intriguing possibility: perhaps one role of \texttt{RRC} is to bootstrap learning background invariance. Future work could potentially investigate this hypothesis.

One effect of \texttt{RRC} is that only a little of the foreground may be present in a view. We investigate the effect of including a lower bound on the amount of foreground included in a view. Concretely, we generate the parameters for \texttt{RRC} imposing one additional constraint: that $\text{FG}_{\text{min}}$ fraction of the foreground be present in the resulting crop. The foreground area is obtained from the corresponding binary foreground mask.  Thus, no constraint corresponds to $\text{FG}_{\text{min}}=0$. Too large a value of $\text{FG}_{\text{min}}$ can be expected to hurt performance, since it overly constrains \texttt{RRC} which is useful for inducing desirable invariances (e.g. occlusion or scale invariance). Crop parameters are sampled in the standard way, defaulting to a \texttt{CenterCrop} if the sampled \texttt{RRC} parameters are rejected 10 times.

 We find that this strategy of imposing a foreground constraint can help when the setting of background augmentation strength is lower than optimal, but it adds little benefit (if any) on top of optimal settings, see Table \ref{tab: moco_min_fg}. As another example, consider \swav, where \bgrm~(\bgrand) results in an accuracy of 73.6\% (73.7\%) with $\text{FG}_{\text{min}}=0$, see Table \ref{tab: swav_aug_strength}. We find no benefit when we impose a constraint of $\text{FG}_{\text{min}}=0.15$, with corresponding accuracies for \bgrm~(\bgrand) of 73.7\% (73.5\%), see Table \ref{tab:u2net_vs_deepusps_ablation}.

\begin{table}
    \centering
    \begin{tabular}{lcccc} \toprule
                                & \texttt{RRC}       & \texttt{CenterCrop}  & \bgswaps  & ImageNet acc. \\ \midrule
        baseline                & \checkmark &            &             &   67.7 \\
        \rownumber{(a)}         & \checkmark &            & \checkmark  &   69.2 \\
        \rownumber{(b)}         &            & \checkmark &             &   26.8 \\
        \rownumber{(c)}         &            & \checkmark & \checkmark   &   49.6 \\ \bottomrule
    \end{tabular}
    \caption{{\bf Impact of \texttt{RandomResizedCrop}.} \texttt{RRC} is critical for good performance in current SSL pipelines. In \moco, replacing \texttt{RRC} with \texttt{CenterCrop} significantly hurts performance, but application of \bgswaps~somewhat helps compensate.  Aug. Strength of \bgswaps:~$p_{\text{pos}}=p_{\text{neg}}=0.1$.}
    \label{tab:no_rrc}
\end{table}

\begin{table}
    \centering
    \begin{tabular}{lccc}\toprule
         Baseline (67.7) & $\text{FG}_{\text{min}}=0$ & $\text{FG}_{\text{min}}=0.1$ & $\text{FG}_{\text{min}}=0.2$  \\\midrule
         $p_{\text{pos}}=p_{\text{neg}}=0.1$ &  69.2 & \textbf{69.7} & 69.4 \\ 
         $p_{\text{pos}}=p_{\text{neg}}=0.2$ &  \textbf{69.7} & \textbf{69.8} & 69.5 \\\bottomrule
    \end{tabular}
    \caption{ \textbf{Impact of including more foreground in views.} Constraining \texttt{RRC} to include more of the foreground in \moco, we see that it can be beneficial when the setting of background augmentation (\bgswaps) strength is lower than optimal, but confers little (if any) benefit on top of the optimal setting of background augmentation strength.}
    \label{tab: moco_min_fg}
\end{table}

\subsection{Sensitivity of SwAV} 
\label{app:swav_sensitive}

\paragraph{Ease of Optimization.} 
We observe that SwAV is more sensitive to high amount of background augmentations when using using \bgrand, with performance degrading less gracefully than BYOL or \moco~when the strength of background augmentation is high. We linked this behavior to the difficulty of the optimization task of learning invariance to random natural backgrounds in conjunction with SwAV's objective function. As discussed in Section \ref{sec:diagnosis_swav}, 
this can be alleviated by increasing the capacity of the projection MLP or by warming up the background augmentations, resulting in stable performance even at very strong augmentation strength, see Figure \ref{tab:swav_wider_and_warmup}.  We linearly warmed up background augmentation strength over 10 epochs.

\paragraph{Crop Scale Ablation.} 
Parametrizing the scale setting for local and global crops as $(0.05, s)$ and $(s, 1)$, the default setting uses $s=0.14$. As discussed in Section \ref{sec:diagnosis_swav}, we find that increasing $s$ can help improve performance when used in conjunction with background augmentations. For \bgrm~(\bgrand), we used $s=0.26$ ($s=0.2$). We used a 5 epoch linear warmup of background augmentation strength. We found that increasing $s$ to 0.2 for the \swav~baseline, i.e. without background augmentations, results in failure---the loss at the end of pre-training is at chance. The  projection MLP capacity was set to 4096/256 in all cases.

\paragraph{Temperature Sensitivity.} 
We also note that we observed high sensitivity to the temperature setting in SwAV, in contrast with \moco. Future work could investigate the source of this sensitivity, potentially further improving performance.

\paragraph{General Settings.} All ablations discussed here (Appendix \ref{app:swav_sensitive}, Section \ref{sec:diagnosis_swav}) use defaults for remaining settings except the experiments with the default projection MLP capacity (2048/128)---these numbers are from the ablation of augmentation strength in Appendix \ref{app:aug_strength_ablation}.

\subsection{Does BN Adaptation Help?}

Due to strong augmentation during SSL pre-training, the statistics of images during pre-training may be different from those in downstream application, e.g. linear evaluation on ImageNet. Intuitively, one might expect that adapting BN statistics might result in improved performance. Indeed, in the supervised setting, it has been shown \citep{schneider_improving_2020, nado_evaluating_2020} that adapting the batch normalization (BN) statistics under distribution shift can result in improved performance.  Here, we consider linear evaluation on ImageNet with and without adaptation of BN statistics in the backbone. Specifically, for adaptation (no adaptation), we use train (eval) mode for BN layers while training the linear classifier and eval (eval) model during evaluation. Intriguingly, we find that adaptation does not necessarily result in improved performance. 

Note that no supervised information is used for BN adaptation. All parameters in the backbone remain frozen to their pre-trained values. All models received full pre-training. Background augmentations used DeepUSPS$^2$.

\begin{table}
    \centering
     \begin{tabular}{lcc}\toprule
    Method  &  \multicolumn{2}{c}{ImageNet acc.}\\
          &  w/o adapt.  & w/ adapt. \\\midrule
    \moco~{\scriptsize{(\textit{repro.})}}  & 71.0 & 71.0 \\
    \moco~+ \bgrm  & 71.9 & 71.9 \\
    \moco~+ \bgswaps  & 72.1 & 72.2 \\
    BYOL {\scriptsize{(\textit{repro.})}} & 74.1 & 73.8 \\
    BYOL + \bgrm & 75.1 & 74.6 \\
    BYOL + \bgrand &  75.2 & 74.8 \\
    SwAV {\scriptsize{(\textit{repro.})}} & 74.9 & 74.7 \\
    SwAV + \bgrm  & 76.1 &  75.1 \\
    SwAV + \bgrand & 76.1 &  75.1 \\
    \bottomrule
    \end{tabular}
    \caption{
    \textbf{Impact of Adapting BN Statistics.} Adaptation does not necessarily result in improved performance.
    }
    \label{tab:adapt_bn_imnet_lin_eval}
\end{table}

\subsection{Additional Information for Ablations in Main Text}

Ablations in Tables \ref{tab:bg_rm_rand_control}, \ref{tab:ablations} follow the settings in Appendix \ref{app:aug_strength_ablation} and use $\text{FG}_{\text{min}}=0$.

\section{Additional Results}
\label{app:add_results}

Here we report additional results, including downstream evaluations of corresponding models which used U$^2$Net for generating foreground masks---evaluations of these models are consistent with evaluations of corresponding DeepUSPS$^2$ models; we therefore skip detailed discussions of these specific results. 

\subsection{ImageNet-9}
\label{app: expand_bg_challenge}
We expand on results in Table \ref{tab:background_challenge_deepusps} in Table \ref{tab:background_challenge_deepusps_expand} to include SEM, which was excluded in the main text to avoid clutter.  These results correspond to \textit{medium} duration pre-trained models. We report corresponding results for the \textit{full} duration pre-trained models in Table \ref{tab:background_challenge_deepusps_full}. Corresponding tables for U$^2$Net are Tables \ref{tab:background_challenge_u2net} and \ref{tab:background_challenge_u2net_full}; for convenience, we report also report corresponding BG-Gap in Table \ref{tab:bg_gap_u2net}.

\begin{table}
    \centering
    \begin{tabular}{llcccc}\toprule
    \multicolumn{2}{l}{Method}  &     \multicolumn{2}{c}{1\% Labels} & \multicolumn{2}{c}{10\% Labels}\\
           \cmidrule(lr){3-4}  \cmidrule(lr){5-6} \\
           && Top-1 & Top-5 & Top-1 & Top-5 \\ \midrule
    \multicolumn{2}{l}{\textcolor{gray}{Supervised}} 
    & \textcolor{gray}{25.4} & \textcolor{gray}{48.4} & \textcolor{gray}{56.4} &  \textcolor{gray}{80.4} \\
    \midrule
    \parbox[t]{3mm}{\multirow{9}{*}{\rotatebox[origin=c]{90}{Linear}}}
    &\moco~{\scriptsize{(\textit{repro.})}}              & 52.0 & 77.7 & 63.9 & 85.8 \\
    &\moco~+ \bgrm      & 54.4 (\textbf{\textcolor{ForestGreen}{+2.4}}) & 78.7 & 65.2 (\textbf{\textcolor{ForestGreen}{+1.3}}) & 86.3 \\
    &\moco~+ \bgswaps   & 56.4 (\textbf{\textcolor{ForestGreen}{+4.4}}) & 79.8 & 65.8 (\textbf{\textcolor{ForestGreen}{+1.9}}) & 86.5 \\
    \vspace{-0.2mm}
    &\multicolumn{5}{@{}c@{}}{\makebox[0.76\linewidth]{\dashrule[gray]}} \\
    \vspace{-0.2mm}
    &BYOL {\scriptsize{(\textit{repro.})}}                & 57.5 & 80.8 & 68.6 & 88.6 \\
    &BYOL + \bgrm        & \underline{60.8} (\textbf{\textcolor{ForestGreen}{+3.3}}) & 82.9 & 70.2 (\textbf{\textcolor{ForestGreen}{+1.6}})  & 89.3 \\
    &BYOL + \bgrand      & \textbf{61.0} (\textbf{\textcolor{ForestGreen}{+3.5}}) & \textbf{83.5} & 70.6 (\textbf{\textcolor{ForestGreen}{+2.0}}) & 89.6 \\
    \vspace{-0.2mm}
    &\multicolumn{5}{@{}c@{}}{\makebox[0.76\linewidth]{\dashrule[gray]}} \\
    \vspace{-0.2mm}
    &SwAV {\scriptsize{(\textit{repro.})}}                & 52.8 & 78.4 & 68.3 & 88.7 \\
    &SwAV + \bgrm        & 57.6 (\textbf{\textcolor{ForestGreen}{+4.8}}) & 81.8 & 70.3 (\textbf{\textcolor{ForestGreen}{+2.0}}) & 89.8 \\
    &SwAV + \bgrand      & 56.4 (\textbf{\textcolor{ForestGreen}{+3.6}}) & 80.8 & 70.3 (\textbf{\textcolor{ForestGreen}{+2.0}}) & 89.8 \\
    \midrule
    \parbox[t]{3mm}{\multirow{9}{*}{\rotatebox[origin=c]{90}{Finetune}}}
    &\moco~{\scriptsize{(\textit{repro.})}}              & 54.1 & 81.3 & 67.6 & 89.4 \\
    &\moco~+ \bgrm      & 55.3 (\textbf{\textcolor{ForestGreen}{+1.2}}) & 81.4 & 68.0 (\textbf{\textcolor{ForestGreen}{+0.4}}) & 89.3 \\
    &\moco~+ \bgswaps   & 57.7 (\textbf{\textcolor{ForestGreen}{+3.6}}) & 82.7 & 68.8 (\textbf{\textcolor{ForestGreen}{+1.2}}) & 89.6 \\
    \vspace{-0.2mm}
    &\multicolumn{5}{@{}c@{}}{\makebox[0.76\linewidth]{\dashrule[gray]}} \\
    \vspace{-0.2mm}
    &BYOL {\scriptsize{(\textit{repro.})}}                & 57.3 & 80.5 & 70.6 & 90.0 \\
    &BYOL + \bgrm        & 60.5 (\textbf{\textcolor{ForestGreen}{+3.2}}) & 82.6 & \underline{71.7} (\textbf{\textcolor{ForestGreen}{+1.1}}) & \underline{90.6} \\
    &BYOL + \bgrand      & 60.7 (\textbf{\textcolor{ForestGreen}{+3.4}}) & \underline{83.1} & \textbf{71.9} (\textbf{\textcolor{ForestGreen}{+1.3}}) & \textbf{90.8} \\
    \vspace{-0.2mm}
    &\multicolumn{5}{@{}c@{}}{\makebox[0.76\linewidth]{\dashrule[gray]}} \\
    \vspace{-0.2mm}
    &SwAV {\scriptsize{(\textit{repro.})}}                & 54.0 & 78.5 & 70.1 & 89.9 \\
    &SwAV + \bgrm        & 54.7 (\textbf{\textcolor{ForestGreen}{+0.7}}) & 78.9 & 70.7 (\textbf{\textcolor{ForestGreen}{+0.6}}) & 90.2 \\
    &SwAV + \bgrand      & 55.7 (\textbf{\textcolor{ForestGreen}{+1.7}}) & 79.3 & 70.8 (\textbf{\textcolor{ForestGreen}{+0.7}}) & 90.2 \\
    \bottomrule
    \end{tabular}
    \caption{\textbf{Limited-Labels Setting.} Background augmentations improve performance in the limited-labels  setting. Linear evaluation using 100\% of ImageNet labels though a standard benchmark, is a somewhat unrealistic setting. Evaluation in the more practical setting of limited-labels reveals even larger improvement in performance. We highlight \textbf{\textcolor{ForestGreen}{performance gains}} due to background augmentations. Similar to Table \ref{tab:lim_lab_deepusps}, but \underline{U$^2$Net} was used for foreground extraction. Best (second best) results are in \textbf{bold} (\underline{underlined}). 
    }
    \label{app:lim_lab_u2net}
\end{table}
 
\begin{table}
    \centering
    \resizebox{1\columnwidth}{!}{
    \begin{tabular}{lcccccccccc}\toprule
Data Set  & Supervised & \multicolumn{3}{c}{\textbf{\moco}}  & \multicolumn{3}{c}{\textbf{BYOL}} & \multicolumn{3}{c}{\textbf{SwAV}}
    \\\cmidrule(lr){3-5} \cmidrule(lr){6-8} \cmidrule(lr){9-11} 
     &      & baseline & \bgrm & \bgswaps & baseline & \bgrm & \bgrand & baseline & \bgrm & \bgrand \\ \midrule
Original    & 95.6 &  92.7 & 93.8  & 94.2  & 94.9  & 95.6  &   96.0  &   94.1 & 95.0 & 94.9 \\  
Only-BG-B $\downarrow$  & 11.4 &  6.1 &  6.1   & 3.6   & 5.4   & 4.9   & 6.0  &   10.9 & 8.8  & 8.3  \\
Only-BG-T $\downarrow$  & 16.3 & 14.8 &  12.9  & 9.3  & 12.7  & 11.8   &  11.5 &  15.8 & 16.7 & 17.6 \\
No-FG      & 45.9 & 37.8 &  42.3  & 39.6  & 43.9  & 45.9   &   46.2 &    41.3 & 44.2 & 45.2 \\
\rowcolor{lightgray}
Only-FG $\uparrow$     & 86.8 & 74.4 &  81.9{\scriptsize{$\pm$0.1}} (\textbf{\textcolor{ForestGreen}{+7.5}})  & 86.1{\scriptsize{$\pm$0.4}} (\textbf{\textcolor{ForestGreen}{+11.7}})  & 83.5  & 88.8{\scriptsize{$\pm$0.1}} (\textbf{\textcolor{ForestGreen}{+5.3}})  &   87.7{\scriptsize{$\pm$0.6}}  (\textbf{\textcolor{ForestGreen}{+4.2}}) &    79.4 & 85.3{\scriptsize{$\pm$0.1}}  (\textbf{\textcolor{ForestGreen}{+5.9}}) & 84.3{\scriptsize{$\pm$0.2}}  (\textbf{\textcolor{ForestGreen}{+4.9}}) \\
\rowcolor{lightgray}
Mixed-Same $\uparrow$ & 86.2 & 81.8 &  84.0{\scriptsize{$\pm$0.1}} (\textbf{\textcolor{ForestGreen}{+2.2}})  & 87.9{\scriptsize{$\pm$0.3}} (\textbf{\textcolor{ForestGreen}{+6.1}})  & 86.2  & 88.6{\scriptsize{$\pm$0.2}} (\textbf{\textcolor{ForestGreen}{+2.4}})   &   90.1{\scriptsize{$\pm$0.1}} (\textbf{\textcolor{ForestGreen}{+3.9}}) &    82.2 & 86.1{\scriptsize{$\pm$0.3}} (\textbf{\textcolor{ForestGreen}{+3.9}}) & 86.3{\scriptsize{$\pm$0.2}} (\textbf{\textcolor{ForestGreen}{+4.1}}) \\
\rowcolor{lightgray}
Mixed-Rand $\uparrow$ & 78.9 & 70.7 &  76.3{\scriptsize{$\pm$0.2}} (\textbf{\textcolor{ForestGreen}{+5.6}}) & 84.1{\scriptsize{$\pm$0.3}} (\textbf{\textcolor{ForestGreen}{+13.4}})  & 79.6  & 83.2{\scriptsize{$\pm$0.1}} (\textbf{\textcolor{ForestGreen}{+3.6}})   &   85.5{\scriptsize{$\pm$0.3}} (\textbf{\textcolor{ForestGreen}{+5.9}}) &    71.3 & 77.1{\scriptsize{$\pm$0.3}} (\textbf{\textcolor{ForestGreen}{+5.8}}) & 77.0{\scriptsize{$\pm$0.3}} (\textbf{\textcolor{ForestGreen}{+5.7}}) \\
\rowcolor{lightgray}
Mixed-Next $\uparrow$ & 77.2 & 67.0 &  73.0{\scriptsize{$\pm$0.1}} (\textbf{\textcolor{ForestGreen}{+6.0}})  & 82.2{\scriptsize{$\pm$0.4}} (\textbf{\textcolor{ForestGreen}{+15.2}})  & 77.6  & 80.7{\scriptsize{$\pm$0.1}} (\textbf{\textcolor{ForestGreen}{+3.1}})   &   84.0{\scriptsize{$\pm$0.1}} (\textbf{\textcolor{ForestGreen}{+6.4}}) &    69.0 & 74.3{\scriptsize{$\pm$0.2}} (\textbf{\textcolor{ForestGreen}{+5.3}}) & 74.4{\scriptsize{$\pm$0.2}} (\textbf{\textcolor{ForestGreen}{+5.4}}) \\ \bottomrule
    \end{tabular}
    }
    \caption{\textbf{Robustness: Foreground-Background Shifts.} Background augmentations result in large performance gains on ImageNet-9 (IN-9) across all SSL methods, with \bgswaps~generally exhibiting similar or better performance than \bgrm. We highlight the performance benefit on the variants of IN-9 especially relevant to our work. All pre-training durations correspond to respective \underline{\textit{medium}} settings. Note that IN-9 uses only 9 classes, so chance is $\sim$11.1\%. This table is an expanded version of Table \ref{tab:background_challenge_deepusps}, to include SEM which were excluded in the main text to avoid clutter.}
    \label{tab:background_challenge_deepusps_expand}
\end{table}

\begin{table}
    \centering
    \resizebox{1\columnwidth}{!}{
    \begin{tabular}{lcccccccccc}\toprule
Data Set    & Supervised & \multicolumn{3}{c}{\textbf{\moco}}  & \multicolumn{3}{c}{\textbf{BYOL}} & \multicolumn{3}{c}{\textbf{SwAV}}
    \\\cmidrule(lr){3-5} \cmidrule(lr){6-8} \cmidrule(lr){9-11} 
     &      & baseline & \bgrm & \bgswaps & baseline & \bgrm & \bgrand & baseline & \bgrm & \bgrand \\ \midrule
Original    & 95.6 &  94.7 & 94.9  & 95.3  & 95.2  & 95.8  &  95.7  &   94.6 & 95.4 & 95.1 \\  
Only-BG-B $\downarrow$  & 11.4 &  7.9 &  6.3   & 5.1   & 7.1   & 5.8   & 6.0  &   11.4 & 10.5  & 10.9  \\
Only-BG-T $\downarrow$  & 16.3 & 14.7 &  13.9  & 11.1  & 16.5  & 13.0  & 14.1 &  19.2 & 18.3 & 18.0 \\
No-FG      & 45.9 & 42.3 &  43.5  & 42.6  & 42.7  & 46.5   &   47.5 &    46.0 & 47.4 & 43.5 \\
\rowcolor{lightgray}
Only-FG $\uparrow$     & 86.8 & 79.7 &  85.2 (\textbf{\textcolor{ForestGreen}{+5.5}})  & 86.9 (\textbf{\textcolor{ForestGreen}{+7.2}})  & 81.4  & 88.5 (\textbf{\textcolor{ForestGreen}{+7.1}})  &  87.3  (\textbf{\textcolor{ForestGreen}{+5.9}}) &    81.9 & 84.1  (\textbf{\textcolor{ForestGreen}{+2.2}}) & 83.2  (\textbf{\textcolor{ForestGreen}{+1.3}}) \\
\rowcolor{lightgray}
Mixed-Same $\uparrow$ & 86.2 & 84.9 &  85.8 (\textbf{\textcolor{ForestGreen}{+0.9}})  & 89.7 (\textbf{\textcolor{ForestGreen}{+4.8}})  & 86.7  & 89.2 (\textbf{\textcolor{ForestGreen}{+2.5}})   &   90.2 (\textbf{\textcolor{ForestGreen}{+3.5}}) &  84.3 & 86.0 (\textbf{\textcolor{ForestGreen}{+1.7}}) & 85.5 (\textbf{\textcolor{ForestGreen}{+1.2}}) \\
\rowcolor{lightgray}
Mixed-Rand $\uparrow$ & 78.9 & 74.9 &  79.0 (\textbf{\textcolor{ForestGreen}{+4.1}}) & 85.3 (\textbf{\textcolor{ForestGreen}{+10.4}})  & 77.6  & 83.9 (\textbf{\textcolor{ForestGreen}{+6.3}}) &  85.8 (\textbf{\textcolor{ForestGreen}{+8.2}}) & 72.9 & 76.7 (\textbf{\textcolor{ForestGreen}{+3.8}}) & 76.5 (\textbf{\textcolor{ForestGreen}{+3.9}}) \\
\rowcolor{lightgray}
Mixed-Next $\uparrow$ & 77.2 & 72.9 &  76.0 (\textbf{\textcolor{ForestGreen}{+3.1}})  & 82.7 (\textbf{\textcolor{ForestGreen}{+9.8}})  & 75.7  & 82.0 (\textbf{\textcolor{ForestGreen}{+6.3}})   &   84.1 (\textbf{\textcolor{ForestGreen}{+8.4}}) &    70.2 & 74.5 (\textbf{\textcolor{ForestGreen}{+4.3}}) & 73.1 (\textbf{\textcolor{ForestGreen}{+2.9}}) \\\bottomrule
    \end{tabular}
    }
    \caption{\textbf{Robustness: Foreground-Background Shifts.} Background augmentations result in large performance gains on ImageNet-9 (IN-9) across all SSL methods, with \bgswaps~generally exhibiting similar or better performance than \bgrm. We highlight the performance benefit on the variants of IN-9 especially relevant to our work. All pre-training durations correspond to respective \underline{\textit{full}} settings. Note that IN-9 uses only 9 classes, so chance is $\sim$11.1\%.}
    \label{tab:background_challenge_deepusps_full}
\end{table}

\begin{table}
    \centering
    \resizebox{1\columnwidth}{!}{
    \begin{tabular}{lcccccccccl}\toprule
Data Set    & Supervised & \multicolumn{3}{c}{\textbf{\moco}}  & \multicolumn{3}{c}{\textbf{BYOL}} & \multicolumn{3}{c}{\textbf{SwAV}}
    \\\cmidrule(lr){3-5} \cmidrule(lr){6-8} \cmidrule(lr){9-11} 
     &      & baseline & \bgrm & \bgswaps & baseline & \bgrm & \bgrand & baseline & \bgrm & \bgrand \\ \midrule
Original    & 95.6 &  92.7 & 94.1  & 94.4  & 94.9  & 95.8  &   95.9  &   94.1 & 94.8 & 94.7 \\  
Only-BG-B $\downarrow$  & 11.4 &  6.1 &  7.2   & 4.2   & 5.4   & 4.9   & 5.8  &    10.9 & 8.5  & 8.7  \\
Only-BG-T $\downarrow$  & 16.3 & 14.8 &  13.8  & 10.7  & 12.7  & 12.0   &   12.2 &    15.8 & 16.6 & 17.2 \\
No-FG      & 45.9 & 37.8 &  43.5  & 41.7  & 43.9  & 46.5   &   46.2 &    41.3 & 46.1 & 45.3 \\
\rowcolor{lightgray}
Only-FG $\uparrow$     & 86.8 & 74.4 &  83.6{\scriptsize{$\pm$0.3}} (\textbf{\textcolor{ForestGreen}{+9.2}})  & 85.4{\scriptsize{$\pm$0.1}} (\textbf{\textcolor{ForestGreen}{+11}})  & 83.5  & 89.3{\scriptsize{$\pm$0.5}} (\textbf{\textcolor{ForestGreen}{+5.8}})  &   87.6{\scriptsize{$\pm$0.4}}  (\textbf{\textcolor{ForestGreen}{+4.1}}) &    79.4 & 85.4{\scriptsize{$\pm$0.2}}  (\textbf{\textcolor{ForestGreen}{+6.0}}) & 85.3{\scriptsize{$\pm$0.2}}  (\textbf{\textcolor{ForestGreen}{+5.9}}) \\
\rowcolor{lightgray}
Mixed-Same $\uparrow$ & 86.2 & 81.8 &  85.6{\scriptsize{$\pm$0.2}} (\textbf{\textcolor{ForestGreen}{+3.8}})  & 88.2{\scriptsize{$\pm$0.2}} (\textbf{\textcolor{ForestGreen}{+6.4}})  & 86.2  & 89.0{\scriptsize{$\pm$0.3}} (\textbf{\textcolor{ForestGreen}{+2.8}})   &   90.4{\scriptsize{$\pm$0.3}} (\textbf{\textcolor{ForestGreen}{+4.2}}) &    82.2 & 86.0{\scriptsize{$\pm$0.4}} (\textbf{\textcolor{ForestGreen}{+3.8}}) & 86.1{\scriptsize{$\pm$0.1}} (\textbf{\textcolor{ForestGreen}{+3.9}}) \\
\rowcolor{lightgray}
Mixed-Rand $\uparrow$ & 78.9 & 70.7 &  78.2{\scriptsize{$\pm$0.2}} (\textbf{\textcolor{ForestGreen}{+7.5}}) & 83.6{\scriptsize{$\pm$0.2}} (\textbf{\textcolor{ForestGreen}{+12.9}})  & 79.6  & 83.8{\scriptsize{$\pm$0.1}} (\textbf{\textcolor{ForestGreen}{+4.2}})   &   85.3{\scriptsize{$\pm$0.2}} (\textbf{\textcolor{ForestGreen}{+5.7}}) &    71.3 & 76.7{\scriptsize{$\pm$0.1}} (\textbf{\textcolor{ForestGreen}{+5.4}}) & 76.8{\scriptsize{$\pm$0.3}} (\textbf{\textcolor{ForestGreen}{+5.5}}) \\
\rowcolor{lightgray}
Mixed-Next $\uparrow$ & 77.2 & 67.0 &  75.2{\scriptsize{$\pm$0.1}} (\textbf{\textcolor{ForestGreen}{+8.2}})  & 81.2{\scriptsize{$\pm$0.2}} (\textbf{\textcolor{ForestGreen}{+14.2}})  & 77.6  & 81.7{\scriptsize{$\pm$0.1}} (\textbf{\textcolor{ForestGreen}{+4.1}})   &   83.5{\scriptsize{$\pm$0.3}} (\textbf{\textcolor{ForestGreen}{+5.9}}) &    69.0 & 73.8{\scriptsize{$\pm$0.4}} (\textbf{\textcolor{ForestGreen}{+4.8}}) & 74.1{\scriptsize{$\pm$0.2}} (\textbf{\textcolor{ForestGreen}{+5.1}}) \\\bottomrule
    \end{tabular}
    }
    \caption{\textbf{Robustness: Foreground-Background Shifts.} Background augmentations result in large performance gains on ImageNet-9 (IN-9) across all SSL methods, with \bgswaps~generally exhibiting similar or better performance than \bgrm. We highlight the performance benefit on the variants of IN-9 especially relevant to our work. All pre-training durations correspond to respective \underline{\textit{medium}} settings. Note that IN-9 uses only 9 classes, so chance is $\sim$11.1\%. Similar to Table \ref{tab:background_challenge_deepusps_expand}, but \underline{U$^2$Net} was used for FG extraction.}
    \label{tab:background_challenge_u2net}
\end{table}

\begin{table}
    \centering
    \resizebox{1\columnwidth}{!}{
    \begin{tabular}{lcccccccccc}\toprule
Data Set    & Supervised & \multicolumn{3}{c}{\textbf{\moco}}  & \multicolumn{3}{c}{\textbf{BYOL}} & \multicolumn{3}{c}{\textbf{SwAV}}
    \\\cmidrule(lr){3-5} \cmidrule(lr){6-8} \cmidrule(lr){9-11} 
    &      & baseline & \bgrm & \bgswaps & baseline & \bgrm & \bgrand & baseline & \bgrm & \bgrand \\ \midrule
Original & 95.6 &  94.7 & 94.8  & 95.2  & 95.2  & 96.1  &  96.1  &   94.6 & 95.3 & 95.4 \\  
Only-BG-B $\downarrow$  & 11.4 &  7.9 &  8.1   & 3.4   & 7.1   & 4.8   & 6.2  &   11.4 & 10.3  & 13.6  \\
Only-BG-T $\downarrow$  & 16.3 & 14.7 &  14.2  & 11.4  & 16.5  & 13.8  & 12.9 &  19.2 & 18.2 & 18.6 \\
No-FG      & 45.9 & 42.3 &  45.5  & 43.0  & 42.7  & 47.9   &   48.1 &    46.0 & 46.8 & 44.9 \\
\rowcolor{lightgray}
Only-FG $\uparrow$     & 86.8 & 79.7 &  87.1 (\textbf{\textcolor{ForestGreen}{+7.4}})  & 87.5 (\textbf{\textcolor{ForestGreen}{+7.8}})  & 81.4  & 89.1 (\textbf{\textcolor{ForestGreen}{+7.7}})  &  88.3  (\textbf{\textcolor{ForestGreen}{+6.9}}) &    81.9 & 86.8  (\textbf{\textcolor{ForestGreen}{+4.9}}) & 83.8  (\textbf{\textcolor{ForestGreen}{+1.9}}) \\
\rowcolor{lightgray}
Mixed-Same $\uparrow$ & 86.2 & 84.9 &  87.1 (\textbf{\textcolor{ForestGreen}{+2.2}})  & 89.6 (\textbf{\textcolor{ForestGreen}{+4.7}})  & 86.7  & 89.5 (\textbf{\textcolor{ForestGreen}{+2.8}})   &   90.2 (\textbf{\textcolor{ForestGreen}{+3.5}}) &  84.3 & 87.0 (\textbf{\textcolor{ForestGreen}{+2.7}}) & 86.9 (\textbf{\textcolor{ForestGreen}{+2.6}}) \\
\rowcolor{lightgray}
Mixed-Rand $\uparrow$ & 78.9 & 74.9 &  80.7 (\textbf{\textcolor{ForestGreen}{+5.8}}) & 85.2 (\textbf{\textcolor{ForestGreen}{+10.3}})  & 77.6  & 84.2 (\textbf{\textcolor{ForestGreen}{+6.6}}) &  85.2 (\textbf{\textcolor{ForestGreen}{+7.6}}) & 72.9 & 77.1 (\textbf{\textcolor{ForestGreen}{+4.2}}) & 76.6 (\textbf{\textcolor{ForestGreen}{+3.7}}) \\
\rowcolor{lightgray}
Mixed-Next $\uparrow$ & 77.2 & 72.9 &  78.1 (\textbf{\textcolor{ForestGreen}{+5.2}})  & 83.2 (\textbf{\textcolor{ForestGreen}{+10.3}})  & 75.7  & 81.7 (\textbf{\textcolor{ForestGreen}{+6.0}})   &   83.8 (\textbf{\textcolor{ForestGreen}{+8.1}}) &    70.2 & 75.6 (\textbf{\textcolor{ForestGreen}{+5.4}}) & 74.3 (\textbf{\textcolor{ForestGreen}{+4.1}}) \\\bottomrule
    \end{tabular}
    }
    \caption{\textbf{Robustness: Foreground-Background Shifts.} Background augmentations result in large performance gains on ImageNet-9 (IN-9) across all SSL methods, with \bgswaps~generally exhibiting similar or better performance than \bgrm. We highlight the performance benefit on the variants of IN-9 especially relevant to our work. All pre-training durations correspond to respective \underline{\textit{full}} settings. Note that IN-9 uses only 9 classes, so chance is $\sim$11.1\%. Similar to Table \ref{tab:background_challenge_deepusps_full}, but \underline{U$^2$Net} was used for FG extraction.}
    \label{tab:background_challenge_u2net_full}
\end{table}

\begin{table}
    \centering
    \resizebox{1\columnwidth}{!}{
    \begin{tabular}{cccccccccl}\toprule
     \shortstack{Pre-Train\\Duration} & \multicolumn{3}{c}{\textbf{\moco}}  & \multicolumn{3}{c}{\textbf{BYOL}} & \multicolumn{3}{c}{\textbf{SwAV}}
    \\\cmidrule(lr){2-4} \cmidrule(lr){5-7} \cmidrule(lr){8-10} 
 & baseline & \bgrm & \bgswaps & baseline & \bgrm & \bgrand & baseline & \bgrm & \bgrand \\ \midrule
Med.
& 11.1 & 7.4 (\textbf{\textcolor{ForestGreen}{-3.7}})  & 4.6 (\textbf{\textcolor{ForestGreen}{-6.5}})
& 6.6 & 5.2 (\textbf{\textcolor{ForestGreen}{-1.4}})   & 5.1 (\textbf{\textcolor{ForestGreen}{-1.5}}) 
& 10.9 & 9.3 (\textbf{\textcolor{ForestGreen}{-1.6}})  & 9.3 (\textbf{\textcolor{ForestGreen}{-1.6}})  \\
Full 
& 10.0  & 6.4 (\textbf{\textcolor{ForestGreen}{-3.6}})  & 4.4 (\textbf{\textcolor{ForestGreen}{-5.6}})
& 9.1  &  5.3 (\textbf{\textcolor{ForestGreen}{-3.8}})  & 5.0 (\textbf{\textcolor{ForestGreen}{-4.1}})  
& 11.4 &  9.9 (\textbf{\textcolor{ForestGreen}{-1.5}})  & 10.3 (\textbf{\textcolor{ForestGreen}{-1.1}})  \\  
\bottomrule
    \end{tabular}
    }
    \caption{\textbf{BG-Gap:} Background augmentations decrease BG-Gap of SSL Methods. Similar to Table \ref{tab:bg_gap}, but \underline{U$^2$Net} was used for FG extraction.}
    \label{tab:bg_gap_u2net}
\end{table}

\subsection{ImageNet-v2}
\label{app:additional_results_imnet_v2}
Here, we report evaluation on all variants of ImageNet-v2: MatchedFrequency (MF), Threshold0.7 (T0.7) and TopImages (TI). Background augmentations result in large performance gains across all variants. Corresponding results for U$^2$Net are shown in Table \ref{tab:expanded_imnetv2_u2net}.

\begin{table}
    \centering
     \begin{tabular}{lccc}\toprule
    Method  &  \multicolumn{3}{c}{ImageNet-v2 acc.}\\
           &  MF & T0.7 & TI \\\midrule
    \textcolor{gray}{Supervised}  
    & \textcolor{gray}{63.8} & \textcolor{gray}{72.6} & \textcolor{gray}{77.7} \\\midrule
    \multicolumn{1}{@{}l}{\textit{Pre-Train Duration: Medium}}\\
    \moco~{\scriptsize{(\textit{repro.})}} 
    & 54.7 & 63.6 & 69.6 \\
    \moco~+ \bgrm 
    & 56.7\gsem{0.1} (\textbf{\textcolor{ForestGreen}{+2.0}}) & 65.7\gsem{0.1} (\textbf{\textcolor{ForestGreen}{+2.1}}) & 71.5\gsem{0.1} (\textbf{\textcolor{ForestGreen}{+1.9}}) \\
    \moco~+ \bgswaps 
    & 57.2\gsem{0.1} (\textbf{\textcolor{ForestGreen}{+2.5}}) & 66.3\gsem{0.1} (\textbf{\textcolor{ForestGreen}{+2.7}}) & 72.0\gsem{0.2} (\textbf{\textcolor{ForestGreen}{+2.4}}) \\
    BYOL {\scriptsize{(\textit{repro.})}} & 60.7 & 70.2 & 75.6 \\
    BYOL + \bgrm & 61.7\gsem{0.2} (\textbf{\textcolor{ForestGreen}{+1.0}}) & 71.0\gsem{0.2} (\textbf{\textcolor{ForestGreen}{+0.8}}) & 76.3\gsem{0.2} (\textbf{\textcolor{ForestGreen}{+0.7}})   \\
    BYOL + \bgrand & \textbf{62.1}\gsem{0.1} (\textbf{\textcolor{ForestGreen}{+1.4}}) & \textbf{71.5}\gsem{0.0} (\textbf{\textcolor{ForestGreen}{+1.3}}) & \textbf{76.7}\gsem{0.1} (\textbf{\textcolor{ForestGreen}{+1.1}}) \\
    SwAV {\scriptsize{(\textit{repro.})}} & 59.3 & 68.5 & 73.9 \\
    SwAV + \bgrm  & 61.2\gsem{0.3} (\textbf{\textcolor{ForestGreen}{+1.9}}) & 70.3\gsem{0.1} (\textbf{\textcolor{ForestGreen}{+1.8}}) & 75.6\gsem{0.1} (\textbf{\textcolor{ForestGreen}{+1.7}})\\
    SwAV + \bgrand  & 60.7\gsem{0.0} (\textbf{\textcolor{ForestGreen}{+1.4}}) 
    &  70.0\gsem{0.2} (\textbf{\textcolor{ForestGreen}{+1.5}}) &  75.3\gsem{0.1} (\textbf{\textcolor{ForestGreen}{+1.4}}) \\
    \midrule
    \multicolumn{1}{@{}l}{\textit{Pre-Train Duration: Full}}\\
    \moco~{\scriptsize{(\textit{repro.})}} 
    & 58.9 & 67.4 & 73.1 \\
    \moco~+ \bgrm 
    & 59.6 (\textbf{\textcolor{ForestGreen}{+0.7}}) & 68.7 (\textbf{\textcolor{ForestGreen}{+1.3}}) 
    & 74.2 (\textbf{\textcolor{ForestGreen}{+1.1}}) \\
    \moco~+ \bgswaps 
    & 60.3 (\textbf{\textcolor{ForestGreen}{+1.4}}) & 68.8 (\textbf{\textcolor{ForestGreen}{+1.4}}) 
    & 74.9 (\textbf{\textcolor{ForestGreen}{+1.8}}) \\
    BYOL {\scriptsize{(\textit{repro.})}} & 61.9 & 71.2 & 76.2\\
    BYOL + \bgrm & 63.4 (\textbf{\textcolor{ForestGreen}{+1.5}}) & 72.4  (\textbf{\textcolor{ForestGreen}{+1.2}})& 77.7 (\textbf{\textcolor{ForestGreen}{+1.5}})   \\
    BYOL + \bgrand & 62.8 (\textbf{\textcolor{ForestGreen}{+0.9}}) & 72.4 (\textbf{\textcolor{ForestGreen}{+1.2}}) & 77.2 (\textbf{\textcolor{ForestGreen}{+1.0}})   \\
    SwAV {\scriptsize{(\textit{repro.})}} & 61.7 & 70.8 & 76.4 \\
    SwAV + \bgrm  & \textbf{63.8} (\textbf{\textcolor{ForestGreen}{+2.1}}) & \textbf{72.8} (\textbf{\textcolor{ForestGreen}{+2.0}}) & \textbf{77.9} (\textbf{\textcolor{ForestGreen}{+1.5}}) \\
    SwAV + \bgrand  & 63.4 (\textbf{\textcolor{ForestGreen}{+1.7}}) & 72.3 (\textbf{\textcolor{ForestGreen}{+1.5}}) & 77.9 (\textbf{\textcolor{ForestGreen}{+1.5}})  \\
    \bottomrule
    \end{tabular}
    \caption{
    \textbf{Robustness: Natural Distribution Shift.} Expanded version of Table \ref{tab:imagenet_v2_deepusps_sq}.  Background augmentations improve performance on all variants of ImageNet-v2. Notably, background augmentations enable \swav~to perform \textit{on par} with the standard supervised baseline. Best results are in \textbf{bold}. Notation: MF=MatchedFrequency, T0.7=Threshold0.7, TI=TopImages. Results in Table \ref{tab:imagenet_v2_deepusps_sq} correspond to MF and are included here for completeness.
    }
    \label{tab:expanded_imnetv2_deepusps_sq}
\end{table}

\begin{table}
    \centering
     \begin{tabular}{lccc}\toprule
    Method  &  \multicolumn{3}{c}{ImageNet-v2 acc.}\\
           &  MF & T0.7 & TI \\\midrule
    \textcolor{gray}{Supervised}  
    & \textcolor{gray}{63.8} & \textcolor{gray}{72.6} & \textcolor{gray}{77.7} \\\midrule
    \multicolumn{1}{@{}l}{\textit{Pre-Train Duration: Medium}}\\
    \moco~{\scriptsize{(\textit{repro.})}} 
    & 54.7 & 63.6 & 69.6 \\
    \moco~+ \bgrm 
    & 57.0\gsem{0.1} (\textbf{\textcolor{ForestGreen}{+2.3}}) & 66.2\gsem{0.2} (\textbf{\textcolor{ForestGreen}{+2.6}}) & 71.8\gsem{0.1} (\textbf{\textcolor{ForestGreen}{+2.2}}) \\
    \moco~+ \bgswaps 
    & 57.6\gsem{0.0} (\textbf{\textcolor{ForestGreen}{+2.9}}) & 66.5\gsem{0.2} (\textbf{\textcolor{ForestGreen}{+2.9}}) & 72.3\gsem{0.1} (\textbf{\textcolor{ForestGreen}{+2.7}}) \\
    BYOL {\scriptsize{(\textit{repro.})}} & 60.7 & 70.2 & 75.6 \\
    BYOL + \bgrm & \textbf{62.2}\gsem{0.2} (\textbf{\textcolor{ForestGreen}{+1.5}}) & 71.3\gsem{0.1} (\textbf{\textcolor{ForestGreen}{+1.1}}) & 76.5\gsem{0.1} (\textbf{\textcolor{ForestGreen}{+0.9}})   \\
    BYOL + \bgrand & \textbf{62.2}\gsem{0.4} (\textbf{\textcolor{ForestGreen}{+1.5}}) & \textbf{71.5}\gsem{0.1} (\textbf{\textcolor{ForestGreen}{+1.3}}) & \textbf{76.6}\gsem{0.2} (\textbf{\textcolor{ForestGreen}{+1.0}}) \\
    SwAV {\scriptsize{(\textit{repro.})}} & 59.3 & 68.5 & 73.9 \\
    SwAV + \bgrm  & 61.2\gsem{0.1} (\textbf{\textcolor{ForestGreen}{+1.9}}) & 70.2\gsem{0.2} (\textbf{\textcolor{ForestGreen}{+1.7}}) & 75.4\gsem{0.1} (\textbf{\textcolor{ForestGreen}{+1.5}})\\
    SwAV + \bgrand  & 61.0\gsem{0.3} (\textbf{\textcolor{ForestGreen}{+1.7}}) 
    &  69.9\gsem{0.2} (\textbf{\textcolor{ForestGreen}{+1.4}}) &  75.4\gsem{0.1} (\textbf{\textcolor{ForestGreen}{+1.5}}) \\
    \midrule
    \multicolumn{1}{@{}l}{\textit{Pre-Train Duration: Full}}\\
    \moco~{\scriptsize{(\textit{repro.})}} 
    & 58.9 & 67.4 & 73.1 \\
    \moco~+ \bgrm 
    & 60.2 (\textbf{\textcolor{ForestGreen}{+1.3}}) & 69.2 (\textbf{\textcolor{ForestGreen}{+1.8}}) 
    & 74.6 (\textbf{\textcolor{ForestGreen}{+1.5}}) \\
    \moco~+ \bgswaps 
    & 60.5 (\textbf{\textcolor{ForestGreen}{+1.6}}) & 69.3 (\textbf{\textcolor{ForestGreen}{+1.9}}) 
    & 75.1 (\textbf{\textcolor{ForestGreen}{+2.0}}) \\
    BYOL {\scriptsize{(\textit{repro.})}} & 61.9 & 71.2 & 76.2\\
    BYOL + \bgrm & 62.9 (\textbf{\textcolor{ForestGreen}{+1.0}}) & 72.2  (\textbf{\textcolor{ForestGreen}{+1.0}})& 77.5 (\textbf{\textcolor{ForestGreen}{+1.3}})   \\
    BYOL + \bgrand & 63.2 (\textbf{\textcolor{ForestGreen}{+1.3}}) & 71.9 (\textbf{\textcolor{ForestGreen}{+0.7}}) & 77.1 (\textbf{\textcolor{ForestGreen}{+0.9}})   \\
    SwAV {\scriptsize{(\textit{repro.})}} & 61.7 & 70.8 & 76.4 \\
    SwAV + \bgrm  & \textbf{63.7} (\textbf{\textcolor{ForestGreen}{+2.0}}) & \textbf{73.1} (\textbf{\textcolor{ForestGreen}{+2.3}}) & \textbf{78.2} (\textbf{\textcolor{ForestGreen}{+1.8}}) \\
    SwAV + \bgrand  & 63.5 (\textbf{\textcolor{ForestGreen}{+1.8}}) & 72.4 (\textbf{\textcolor{ForestGreen}{+1.6}}) & 77.7 (\textbf{\textcolor{ForestGreen}{+1.3}})  \\
    \bottomrule
    \end{tabular}
    \caption{
    \textbf{Robustness: Natural Distribution Shift.} Background augmentations improve performance on all variants of ImageNet-v2. Notably, background augmentations enable \swav~to perform \textit{on par} with the standard supervised baseline. Best results are in \textbf{bold}. Notation: MF=MatchedFrequency, T0.7=Threshold0.7, TI=TopImages. Similar to Table \ref{tab:expanded_imnetv2_deepusps_sq}, but \underline{U$^2$Net} was used for FG extraction.
    }
    \label{tab:expanded_imnetv2_u2net}
\end{table}

\begin{table}
    \centering
    \resizebox{1\columnwidth}{!}{
    \begin{tabular}{cccccccccl}\toprule
  \shortstack{Pre-Train\\Duration}  & \multicolumn{3}{c}{\textbf{\moco}}  & \multicolumn{3}{c}{\textbf{BYOL}} & \multicolumn{3}{c}{\textbf{SwAV}}
    \\\cmidrule(lr){2-4} \cmidrule(lr){5-7} \cmidrule(lr){8-10} 
 & baseline & \bgrm & \bgswaps & baseline & \bgrm & \bgrand & baseline & \bgrm & \bgrand \\ \midrule
Med.
& 14.4 & 17.1\sem{0.3} (\textbf{\textcolor{ForestGreen}{+2.7}})  & 18.5\sem{0.3} (\textbf{\textcolor{ForestGreen}{+4.1}})
& 20.4 & 22.9\sem{0.2} (\textbf{\textcolor{ForestGreen}{+2.5}})  & 22.4\sem{0.1} (\textbf{\textcolor{ForestGreen}{+2.0}}) 
& 16.1 & 19.0\sem{0.3\sem{0.1}} (\textbf{\textcolor{ForestGreen}{+2.9}})  & 18.3\sem{0.2} (\textbf{\textcolor{ForestGreen}{+2.2}})  \\
Full
& 17.4 & 20.2 (\textbf{\textcolor{ForestGreen}{+2.8}})  & 21.9 (\textbf{\textcolor{ForestGreen}{+4.5}})
& 20.8 & 24.3 (\textbf{\textcolor{ForestGreen}{+3.5}})  & 23.7 (\textbf{\textcolor{ForestGreen}{+2.9}}) 
& 19.3 & 23.1 (\textbf{\textcolor{ForestGreen}{+3.8}}) & 21.2 (\textbf{\textcolor{ForestGreen}{+1.9}})  \\
\bottomrule
    \end{tabular}
    }
    \caption{\textbf{Robustness: Rotation, Viewpoint, Background Shift.} Background augmentations improve  performance on ObjectNet, a challenging test set that controls object orientation, viewpoint and background. Similar to Table \ref{tab:objectnet_deepusps_sq}, but \underline{U$^2$Net} was used for FG extraction.}
    \label{tab:objectnet_u2net}
\end{table}

\subsection{ImageNet-A}
\label{app: only_fg_imagenet_a}

We expand on results in Table \ref{tab:nat_adv_examples_deepusps} in Table \ref{app:nat_adv_examples_deepusps_extended} to include evaluation on Only-FG ImageNet-A. While we report these numbers for completeness, as discussed in Section \ref{sec:nat_adv_examples}, evaluation on Only-FG ImageNet-A must be interpreted with caution, since this data set is also challenging for saliency detection. Corresponding results for U$^2$Net are shown in Table \ref{tab:nat_adv_examples_u2net_extended}.

\subsection{ImageNet-C}
\label{app:additional_results_imnet_c}

On ImageNet-C, we report the average performance on the four main categories of corruptions: noise, blur, weather and digital, see Table \ref{tab:imagenet_c}. While background augmentations generally result in improved robustness, they appear to decrease robustness to noise corruptions; this may be due to difficulty discerning between the foreground and background due to the high frequency noise added throughout the image.

\begin{table}
    \centering
    \resizebox{1\columnwidth}{!}{
    \begin{tabular}{cccccccccl}\toprule
  \shortstack{Pre-Train\\Duration}  & \multicolumn{3}{c}{\textbf{\moco}}  & \multicolumn{3}{c}{\textbf{BYOL}} & \multicolumn{3}{c}{\textbf{SwAV}}
    \\\cmidrule(lr){2-4} \cmidrule(lr){5-7} \cmidrule(lr){8-10} 
 & baseline & \bgrm & \bgswaps & baseline & \bgrm & \bgrand & baseline & \bgrm & \bgrand \\ \midrule
 \multicolumn{10}{@{}l}{\textit{ImageNet-A}}\\
Med.
& 3.1 & 3.3{\scriptsize{$\pm$0.1}} (\textbf{\textcolor{ForestGreen}{+0.2}})  & 3.6{\scriptsize{$\pm$0.1}} (\textbf{\textcolor{ForestGreen}{+0.5}})
& 4.4 & 5.8{\scriptsize{$\pm$0.3}} (\textbf{\textcolor{ForestGreen}{+1.4}})  & 6.1{\scriptsize{$\pm$0.1}} (\textbf{\textcolor{ForestGreen}{+1.7}}) 
& 3.7 & 4.2{\scriptsize{$\pm$0.1}} (\textbf{\textcolor{ForestGreen}{+0.5}}) & 4.1{\scriptsize{$\pm$0.1}} (\textbf{\textcolor{ForestGreen}{+0.4}})  \\
Full 
& 4.2  & 4.7 (\textbf{\textcolor{ForestGreen}{+0.5}})  & 5.3 (\textbf{\textcolor{ForestGreen}{+1.1}})
& 5.3  & 7.2 (\textbf{\textcolor{ForestGreen}{+1.9}})  & 7.2 (\textbf{\textcolor{ForestGreen}{+1.9}})  
& 5.2  & 6.0 (\textbf{\textcolor{ForestGreen}{+0.8}})  & 5.7 (\textbf{\textcolor{ForestGreen}{+0.5}})  \\   \multicolumn{10}{@{}l}{\textit{Only-FG ImageNet-A}}\\
 Med.
& 2.8 & 2.8{\scriptsize{$\pm$0.0}} (+0.0)  & 4.2{\scriptsize{$\pm$0.1}} (\textbf{\textcolor{ForestGreen}{+1.4}})
& 3.2 & 4.8{\scriptsize{$\pm$0.1}} (\textbf{\textcolor{ForestGreen}{+1.6}})  & 4.7{\scriptsize{$\pm$0.2}} (\textbf{\textcolor{ForestGreen}{+1.5}}) 
& 2.7 & 4.0{\scriptsize{$\pm$0.1}} (\textbf{\textcolor{ForestGreen}{+1.3}}) &3.5{\scriptsize{$\pm$0.1}} (\textbf{\textcolor{ForestGreen}{+0.8}})  \\
Full 
& 3.4  & 3.1 (\textbf{\textcolor{BrickRed}{-0.3}})  & 4.7 (\textbf{\textcolor{ForestGreen}{+1.3}})
& 3.1  & 5.8 (\textbf{\textcolor{ForestGreen}{+2.7}})  & 5.1 (\textbf{\textcolor{ForestGreen}{+2.0}})  
& 3.9  & 4.0 (\textbf{\textcolor{ForestGreen}{+0.1}})  & 3.8 (\textbf{\textcolor{BrickRed}{-0.1}})  \\  
\bottomrule
    \end{tabular}
    }
    \caption{\textbf{Robustness: Natural Adversarial Examples.} Background augmentations improve performance on ImageNet-A, a data set of natural adversarial examples.  Interestingly, the performance of all SSL methods drops when presented with only foreground, but background augmentations provide some robustness against this distribution shift as well. Similar to Table \ref{tab:nat_adv_examples_deepusps}, but expanded to include only FG ImageNet-A.}
    \label{app:nat_adv_examples_deepusps_extended}
\end{table}

\begin{table}
    \centering
    \resizebox{1\columnwidth}{!}{
    \begin{tabular}{cccccccccl}\toprule
  \shortstack{Pre-Train\\Duration}  & \multicolumn{3}{c}{\textbf{\moco}}  & \multicolumn{3}{c}{\textbf{BYOL}} & \multicolumn{3}{c}{\textbf{SwAV}}
    \\\cmidrule(lr){2-4} \cmidrule(lr){5-7} \cmidrule(lr){8-10} 
 & baseline & \bgrm & \bgswaps & baseline & \bgrm & \bgrand & baseline & \bgrm & \bgrand \\ \midrule
 \multicolumn{10}{@{}l}{\textit{ImageNet-A}}\\
Med.
& 3.1 & 3.4{\scriptsize{$\pm$0.1}} (\textbf{\textcolor{ForestGreen}{+0.3}})  & 3.9{\scriptsize{$\pm$0.0}} (\textbf{\textcolor{ForestGreen}{+0.8}})
& 4.4 & 6.0{\scriptsize{$\pm$0.2}} (\textbf{\textcolor{ForestGreen}{+1.6}})  & 6.3{\scriptsize{$\pm$0.1}} (\textbf{\textcolor{ForestGreen}{+1.9}}) 
& 3.7 & 4.1{\scriptsize{$\pm$0.0}} (\textbf{\textcolor{ForestGreen}{+0.4}}) & 4.1{\scriptsize{$\pm$0.1}} (\textbf{\textcolor{ForestGreen}{+0.4}})  \\
Full 
& 4.2  & 4.8 (\textbf{\textcolor{ForestGreen}{+0.4}})  & 5.4 (\textbf{\textcolor{ForestGreen}{+1.2}})
& 5.3  & 7.4 (\textbf{\textcolor{ForestGreen}{+2.1}})  & 7.1 (\textbf{\textcolor{ForestGreen}{+1.8}}) 
& 5.2  & 6.1 (\textbf{\textcolor{ForestGreen}{+0.9}})  & 6.1 (\textbf{\textcolor{ForestGreen}{+0.9}})  \\   \multicolumn{10}{@{}l}{\textit{Only-FG ImageNet-A}}\\
 Med.
& 2.8 & 3.2{\scriptsize{$\pm$0.3}} (\textbf{\textcolor{ForestGreen}{+0.4}})  & 4.4{\scriptsize{$\pm$0.1}} (\textbf{\textcolor{ForestGreen}{+1.6}})
& 3.2 & 4.9{\scriptsize{$\pm$0.2}} (\textbf{\textcolor{ForestGreen}{+1.7}})   & 4.8{\scriptsize{$\pm$0.2}} (\textbf{\textcolor{ForestGreen}{+1.6}}) 
& 2.7 & 4.0{\scriptsize{$\pm$0.1}} (\textbf{\textcolor{ForestGreen}{+1.3}}) & 3.6{\scriptsize{$\pm$0.1}} (\textbf{\textcolor{ForestGreen}{+0.9}})  \\
Full 
& 3.4  & 4.3 (\textbf{\textcolor{ForestGreen}{+0.9}})  & 4.6 (\textbf{\textcolor{ForestGreen}{+1.2}}) 
& 3.1  & 5.9 (\textbf{\textcolor{ForestGreen}{+2.8}})  & 5.0 (\textbf{\textcolor{ForestGreen}{+1.9}})  
& 3.9  & 4.7 (\textbf{\textcolor{ForestGreen}{+0.8}})  & 4.0 (\textbf{\textcolor{ForestGreen}{+0.1}})  \\  
\bottomrule
    \end{tabular}
    }
    \caption{ \textbf{Robustness: Natural Adversarial Examples.} Background augmentations improve performance on ImageNet-A, a data set of natural adversarial examples. Interestingly, the performance of all SSL methods drops when presented with only foreground, but background augmentations provide some robustness against this distribution shift as well. Similar to Table \ref{app:nat_adv_examples_deepusps_extended}, but \underline{U$^2$Net} was used for FG extraction.}
    \label{tab:nat_adv_examples_u2net_extended}
\end{table}

\begin{table}
    \centering
    \resizebox{1\columnwidth}{!}{
    \begin{tabular}{cccccccccl}\toprule
  \shortstack{Pre-Train\\Duration}  & \multicolumn{3}{c}{\textbf{\moco}}  & \multicolumn{3}{c}{\textbf{BYOL}} & \multicolumn{3}{c}{\textbf{SwAV}}
    \\\cmidrule(lr){2-4} \cmidrule(lr){5-7} \cmidrule(lr){8-10} 
  & baseline & \bgrm & \bgswaps & baseline & \bgrm & \bgrand & baseline & \bgrm & \bgrand \\ \midrule
Med.
& 27.7 & 31.1\sem{0.4} (\textbf{\textcolor{ForestGreen}{+3.4}})  & 32.3\sem{0.1} (\textbf{\textcolor{ForestGreen}{+4.6}})
& 36.3 & 39.7\sem{0.1} (\textbf{\textcolor{ForestGreen}{+3.4}})  & 38.4\sem{0.1} (\textbf{\textcolor{ForestGreen}{+2.1}}) 
& 27.9 & 32.0\sem{0.2} (\textbf{\textcolor{ForestGreen}{+4.1}}) & 31.4\sem{0.1} \textbf{\textcolor{ForestGreen}{+3.5}})  \\
Full
& 30.4 & 33.4 (\textbf{\textcolor{ForestGreen}{+3.0}})  & 34.2 (\textbf{\textcolor{ForestGreen}{+3.8}})
& 34.4 & 40.5 (\textbf{\textcolor{ForestGreen}{+6.1}})  & 39.7 (\textbf{\textcolor{ForestGreen}{+5.3}}) 
& 29.4 & 33.6 (\textbf{\textcolor{ForestGreen}{+4.2}}) & 32.5 (\textbf{\textcolor{ForestGreen}{+3.1}})  \\
\bottomrule
    \end{tabular}
    }
    \caption{\textbf{Robustness: Renditions.} Background augmentations improve performance on ImageNet-R, a data set of ImageNet-Renditions (e.g. paintings, sculpture). Similar to Table \ref{tab:imagenet_r_deepusps_sq}, but \underline{U$^2$Net} was used for FG extraction.}
    \label{tab:imagenet_r_u2net}
\end{table}

\begin{table}
    \centering
    \resizebox{1\columnwidth}{!}{
    \begin{tabular}{cccccccccl}\toprule
    \shortstack{Pre-Train\\Duration} & \multicolumn{3}{c}{\textbf{\moco}}  & \multicolumn{3}{c}{\textbf{BYOL}} & \multicolumn{3}{c}{\textbf{SwAV}}
    \\\cmidrule(lr){2-4} \cmidrule(lr){5-7} \cmidrule(lr){8-10} 
 & baseline & \bgrm & \bgswaps & baseline & \bgrm & \bgrand & baseline & \bgrm & \bgrand \\ \midrule
Med. & 4.5 & 6.0{\scriptsize{$\pm$0.1}} (\textbf{\textcolor{ForestGreen}{+1.5}})  & 8.6{\scriptsize{$\pm$0.2}} (\textbf{\textcolor{ForestGreen}{+4.1}})  & 10.6  & 11.7{\scriptsize{$\pm$0.2}} (\textbf{\textcolor{ForestGreen}{+1.1}})   & 11.6{\scriptsize{$\pm$0.2}} (\textbf{\textcolor{ForestGreen}{+1.0}})  & 6.0  &  6.4{\scriptsize{$\pm$0.0}} (\textbf{\textcolor{ForestGreen}{+0.4}})    &  6.7{\scriptsize{$\pm$0.1}} (\textbf{\textcolor{ForestGreen}{+0.7}})   \\
Full & 7.8 & 10.1 (\textbf{\textcolor{ForestGreen}{+2.3}})  & 12.6 (\textbf{\textcolor{ForestGreen}{+4.8}})  & 10.4  & 13.7 (\textbf{\textcolor{ForestGreen}{+3.3}})   & 13.5 (\textbf{\textcolor{ForestGreen}{+3.1}})  & 9.1  &  10.2 (\textbf{\textcolor{ForestGreen}{+1.1}})    &  10.5 (\textbf{\textcolor{ForestGreen}{+1.4}})    \\  
\bottomrule
    \end{tabular}
    }
    \caption{\textbf{Robustness: Adversarial Attack.} Background augmentations increase robustness to FGSM adversarial attacks. Similar to Table \ref{tab:adv_attacks_deepusps}, but \underline{U$^2$Net} was used for FG extraction.}
    \label{app:adv_attacks_u2net}
\end{table}

\begin{table}
    \centering
    \resizebox{1\columnwidth}{!}{
    \begin{tabular}{cccccccccl}\toprule
    Data Set & \multicolumn{3}{c}{\textbf{\moco}}  & \multicolumn{3}{c}{\textbf{BYOL}} & \multicolumn{3}{c}{\textbf{SwAV}}
    \\\cmidrule(lr){2-4} \cmidrule(lr){5-7} \cmidrule(lr){8-10} 
 & baseline & \bgrm & \bgswaps & baseline & \bgrm & \bgrand & baseline & \bgrm & \bgrand \\ \midrule
CIFAR-10 & 73.9 & 80.3 (\textbf{\textcolor{ForestGreen}{+6.4}}) & 77.5 (\textbf{\textcolor{ForestGreen}{+3.6}})  & 86.7  & 88.5 (\textbf{\textcolor{ForestGreen}{+1.8}}) & 87.9 (\textbf{\textcolor{ForestGreen}{+1.2}})  &  92.7 &   93.3 (\textbf{\textcolor{ForestGreen}{+0.6}})  &   93.3 (\textbf{\textcolor{ForestGreen}{+0.6}})    \\
CIFAR-100 & 40.8 & 51.4 (\textbf{\textcolor{ForestGreen}{+10.6}}) & 46.3 (\textbf{\textcolor{ForestGreen}{+5.5}})  & 67.6  & 68.2 (\textbf{\textcolor{ForestGreen}{+0.6}}) & 67.1 (\textbf{\textcolor{BrickRed}{-0.5}})  & 76.0  &  77.3   (\textbf{\textcolor{ForestGreen}{+1.3}})  &   76.8 (\textbf{\textcolor{ForestGreen}{+0.8}})    \\ \bottomrule
    \end{tabular}
    }
    \caption{\textbf{CIFAR-10, 100.} Background augmentations improve performance on linear evaluation on CIFAR-10 and 100. Similar to Table \ref{tab:cifar_deepusps}, but \underline{U$^2$Net} was used for FG extraction.}
    \label{app:cifar_u2net}
\end{table}

\begin{table}
    \centering
    \resizebox{1\columnwidth}{!}{
    \begin{tabular}{cccccccccc}\toprule
    & \multicolumn{3}{c}{VOC 07+12 detection}  & \multicolumn{3}{c}{COCO detection} & \multicolumn{3}{c}{COCO instance seg.}
    \\\cmidrule(lr){2-4} \cmidrule(lr){5-7} \cmidrule(lr){8-10} 
Method & AP$_{50}$ & AP & AP$_{75}$ & AP$_{50}$ & AP & AP$_{75}$ & AP$_{50}^{m}$ & AP$^{m}$ & AP$_{75}^{m}$ \\ \midrule
\moco {\scriptsize{(\textit{repro.})}}            & 82.7{\scriptsize{$\pm$0.0}} & 57.9{\scriptsize{$\pm$0.0}} & 64.5{\scriptsize{$\pm$0.1}} & 61.0 & 41.1 & 44.8 & 57.7 & 35.8 & 38.4\\
\rowcolor{lightgray}
\moco + \bgrm    & 82.6{\scriptsize{$\pm$0.1}} & 57.6{\scriptsize{$\pm$0.1}} & 64.5{\scriptsize{$\pm$0.2}} & 60.9 & 41.2 & 44.8 & 57.8 & 35.9 & 38.5\\
\rowcolor{lightgray}
\moco + \bgswaps & 82.7{\scriptsize{$\pm$0.0}} & 57.4{\scriptsize{$\pm$0.2}} & 64.0{\scriptsize{$\pm$0.3}} & 61.2 & 41.4 & 44.8 & 58.0 & 36.0 & 38.3\\ \midrule
BYOL {\scriptsize{(\textit{repro.})}}      & 82.7{\scriptsize{$\pm$0.1}} & 56.7{\scriptsize{$\pm$0.1}} & 63.0{\scriptsize{$\pm$0.3}} & 61.1 & 40.9 & 44.5 & 57.6 & 35.5 & 37.8\\
\rowcolor{lightgray}
BYOL + \bgrm      & 83.0{\scriptsize{$\pm$0.1}} & 57.0{\scriptsize{$\pm$0.1}} & 63.9{\scriptsize{$\pm$0.2}} & 61.5 & 41.1 & 44.3 & 57.8 & 35.5 & 37.8\\
\rowcolor{lightgray}
BYOL + \bgrand   & 83.2{\scriptsize{$\pm$0.1}} & 57.4{\scriptsize{$\pm$0.1}} & 64.1{\scriptsize{$\pm$0.2}} & 61.7 & 41.4 & 44.7 & 57.9 & 35.7 & 37.8\\ \midrule
SwAV {\scriptsize{(\textit{repro.})}}       & 82.3{\scriptsize{$\pm$0.1}} & 55.6{\scriptsize{$\pm$0.0}} & 61.9{\scriptsize{$\pm$0.2}} & 61.4 & 40.7 & 43.7 & 57.6 & 35.4 & 37.4\\
\rowcolor{lightgray}
SwAV + \bgrm      & 82.6{\scriptsize{$\pm$0.0}} & 55.8{\scriptsize{$\pm$0.1}} & 62.0{\scriptsize{$\pm$0.1}} & 61.2 & 40.6 & 43.8 & 57.4 & 35.1 & 37.0\\
\rowcolor{lightgray}
SwAV + \bgrand   & 82.5{\scriptsize{$\pm$0.1}} & 56.0{\scriptsize{$\pm$0.1}} & 62.7{\scriptsize{$\pm$0.1}} & 61.2 & 40.8 & 44.3 & 57.7 & 35.5 & 37.6\\
\bottomrule
    \end{tabular}
    }
    \caption{\textbf{Detection and Instance Segmentation}. Background Augmentations result in small improvements in detection and instance segmentation tasks, likely due to extensive supervision involved in subsequent training. All VOC metrics reported are average of 3 independent runs. Similar to Table \ref{tab:voc_coco_deepusp}, but \underline{U$^2$Net} was used for FG extraction.}
    \label{tab:voc_coco_u2net}
\end{table}

\begin{table}
    \centering
    \resizebox{1\columnwidth}{!}{
    \begin{tabular}{cccccccccc}\toprule
    Supervised & \multicolumn{3}{c}{\textbf{\moco}}  & \multicolumn{3}{c}{\textbf{BYOL}} & \multicolumn{3}{c}{\textbf{SwAV}}
    \\\cmidrule(lr){2-4} \cmidrule(lr){5-7} \cmidrule(lr){8-10} 
& baseline & \bgrm & \bgswaps & baseline & \bgrm & \bgrand & baseline & \bgrm & \bgrand \\ \midrule
 22.1 &  28.8 & 32.3  & 31.9  & 27.6  & 31.7  &   29.1  &   17.0 & 20.1 & 17.4 \\\bottomrule
    \end{tabular}
    }
    \caption{\textbf{Background augmentations increase shape bias.} SSL methods considered generally have a higher shape bias than the supervised baseline. SwAV deviates from this pattern due to \texttt{multi-crop} (SwAV w/o \texttt{multi-crop} shape bias: 27.4). Similar to Table \ref{tab:shape_bias_deepusps}, but \underline{U$^2$Net} was used for FG extraction.}
    \label{app:shape_bias_u2net}
\end{table}

\begin{table}
    \centering
    \resizebox{1\columnwidth}{!}{
    \begin{tabular}{cccccccccl}\toprule
  \shortstack{Corruption}  & \multicolumn{3}{c}{\textbf{\moco}}  & \multicolumn{3}{c}{\textbf{BYOL}} & \multicolumn{3}{c}{\textbf{SwAV}}
    \\\cmidrule(lr){2-4} \cmidrule(lr){5-7} \cmidrule(lr){8-10} 
  & baseline & \bgrm & \bgswaps & baseline & \bgrm & \bgrand & baseline & \bgrm & \bgrand \\ \midrule
  \multicolumn{10}{@{}l}{\rule{0pt}{3ex}\textit{Saliency Method: DeepUSPS}$^2$}\\
\rule{0pt}{2ex}Noise
& 30.3 & 25.9 \rdel{-4.4} & 30.6 \gdel{+0.3}
& 34.6 & 28.3 \rdel{-6.3} & 30.2 \rdel{-4.4}
& 33.0 & 32.6 \rdel{-0.4} & 33.3 \gdel{+0.3} \\
Blur
& 27.1 & 28.2 \gdel{+1.1} & 27.9 \gdel{+0.8}
& 31.3 & 32.4 \gdel{+1.1} & 33.0 \gdel{+1.7}
& 31.3 & 33.0 \gdel{+1.7} & 32.8 \gdel{+1.5} \\
Weather
& 40.1 & 41.7 \gdel{+1.6} & 42.3 \gdel{+2.2}
& 43.6 & 47.7 \gdel{+4.1} & 47.3 \gdel{+3.7}
& 43.7 & 46.2 \gdel{+2.5} & 45.5 \gdel{+1.8} \\
Digital
& 45.9 & 45.0 \rdel{-0.9} & 44.2 \rdel{-1.7}
& 48.9 & 49.7 \gdel{+0.8} & 49.4 \gdel{+0.5}
& 46.8 & 48.6 \gdel{+1.8} & 48.7 \gdel{+1.9} \\
\multicolumn{10}{@{}l}{\rule{0pt}{3ex}\textit{Saliency Method: U}$^2$\textit{Net}}\\
\rule{0pt}{2ex}Noise
& 30.3 & 30.3 (+0.0) & 33.1 \gdel{+2.8}
& 34.6 & 29.3 \rdel{-5.3} & 31.8 \rdel{-2.8}
& 33.0 & 31.8 \rdel{-1.2} & 30.5 \rdel{-2.5} \\
Blur
& 27.1 & 27.8 \gdel{+0.7} & 27.7 \gdel{+0.6}
& 31.3 & 32.4 \gdel{+1.1} & 32.8 \gdel{+1.5}
& 31.3 & 32.7 \gdel{+1.4} & 33.1 \gdel{+1.8} \\
Weather
& 40.1 & 41.6 \gdel{+1.5} & 42.5 \gdel{+2.4}
& 43.6 & 47.6 \gdel{+4.0} & 47.7 \gdel{+4.1}
& 43.7 & 46.7 \gdel{+3.0} & 45.5 \gdel{+1.8} \\
Digital
& 45.9 & 46.2 \gdel{+0.3} & 45.5 \rdel{-0.4}
& 48.9 & 50.3 \gdel{+1.4} & 50.6 \gdel{+1.7}
& 46.8 & 48.6 \gdel{+1.8} & 49.0 \gdel{+2.2} \\
\bottomrule
    \end{tabular}
    }
    \caption{\textbf{Robustness: Corruptions.} Background augmentations generally improve robustness to corruptions in ImageNet-C. We observe that across methods, robustness to added noise (e.g. Gaussian, Speckle) is reduced as a result of background augmentations, while there is improved robustness to blur, weather and digital corruptions. This maybe due to difficulty discerning between the foreground and background due to the high frequency noise added throughout the image. All models received full pre-training.}
    \label{tab:imagenet_c}
\end{table}
\newpage

\FloatBarrier
\vskip 0.2in
\bibliography{references}

\begin{thebibliography}{122}
\providecommand{\natexlab}[1]{#1}
\providecommand{\url}[1]{\texttt{#1}}
\expandafter\ifx\csname urlstyle\endcsname\relax
  \providecommand{\doi}[1]{doi: #1}\else
  \providecommand{\doi}{doi: \begingroup \urlstyle{rm}\Url}\fi

\bibitem[Achanta et~al.(2009)Achanta, Hemami, Estrada, and
  Susstrunk]{achanta_frequency-tuned_2009}
R.~Achanta, S.~Hemami, F.~Estrada, and S.~Susstrunk.
\newblock Frequency-tuned salient region detection.
\newblock In \emph{Proceedings of the {IEEE} {Conference} on Computer Vision
  and Pattern Recognition}, pages 1597--1604, June 2009.
\newblock \doi{10.1109/CVPR.2009.5206596}.
\newblock ISSN: 1063-6919.

\bibitem[Alcorn et~al.(2019)Alcorn, Li, Gong, Wang, Mai, Ku, and
  Nguyen]{alcorn_strike_2019}
M.~A. Alcorn, Q.~Li, Z.~Gong, C.~Wang, L.~Mai, W.-S. Ku, and A.~Nguyen.
\newblock Strike (with) a pose: {Neural} networks are easily fooled by strange
  poses of familiar objects.
\newblock In \emph{Proceedings of the {IEEE}/{CVF} {Conference} on {Computer}
  {Vision} and {Pattern} {Recognition}}, pages 4845--4854, 2019.

\bibitem[Asano et~al.(2020)Asano, Rupprecht, and
  Vedaldi]{asano_self-labelling_2020}
Y.~M. Asano, C.~Rupprecht, and A.~Vedaldi.
\newblock Self-labelling via simultaneous clustering and representation
  learning.
\newblock In \emph{International Conference on Learning Representations}, 2020.

\bibitem[Bachman et~al.(2019)Bachman, Hjelm, and
  Buchwalter]{bachman_learning_2019}
P.~Bachman, R.~D. Hjelm, and W.~Buchwalter.
\newblock Learning representations by maximizing mutual information across
  views.
\newblock \emph{arXiv preprint arXiv:1906.00910}, 2019.

\bibitem[Barbu et~al.(2019)Barbu, Mayo, Alverio, Luo, Wang, Gutfreund,
  Tenenbaum, and Katz]{objectnet}
A.~Barbu, D.~Mayo, J.~Alverio, W.~Luo, C.~Wang, D.~Gutfreund, J.~Tenenbaum, and
  B.~Katz.
\newblock Objectnet: {A} large-scale bias-controlled dataset for pushing the
  limits of object recognition models.
\newblock In \emph{Advances in {Neural} {Information} {Processing} {Systems}},
  2019.

\bibitem[Becker and Hinton(1992)]{becker_self-organizing_1992}
S.~Becker and G.~E. Hinton.
\newblock Self-organizing neural network that discovers surfaces in random-dot
  stereograms.
\newblock \emph{Nature}, 1992.

\bibitem[Beery et~al.(2018)Beery, Van~Horn, and Perona]{beery_recognition_2018}
S.~Beery, G.~Van~Horn, and P.~Perona.
\newblock Recognition in terra incognita.
\newblock In \emph{Proceedings of the {European} conference on computer vision
  ({ECCV})}, pages 456--473, 2018.

\bibitem[Beyer et~al.(2020)Beyer, H{\'e}naff, Kolesnikov, Zhai, and
  Oord]{beyer2020imagenetreal}
L.~Beyer, O.~J. H{\'e}naff, A.~Kolesnikov, X.~Zhai, and A.~v.~d. Oord.
\newblock Are we done with imagenet?
\newblock \emph{arXiv preprint arXiv:2006.07159}, 2020.

\bibitem[Bromley et~al.(1994)Bromley, Guyon, LeCun, Säckinger, and
  Shah]{siamese_bromley_1994}
J.~Bromley, I.~Guyon, Y.~LeCun, E.~Säckinger, and R.~Shah.
\newblock Signature verification using a “siamese” time delay neural
  network.
\newblock In \emph{Neural Information Processing Systems (NeurIPS)}, 1994.

\bibitem[Cai et~al.(2020)Cai, Frankle, Schwab, Morcos, et~al.]{cai_2020}
T.~T. Cai, J.~Frankle, D.~J. Schwab, A.~S. Morcos, et~al.
\newblock Are all negatives created equal in contrastive instance
  discrimination?
\newblock \emph{arXiv preprint arXiv:2010.06682}, 2020.

\bibitem[Caron et~al.(2018)Caron, Bojanowski, Joulin, and
  Douze]{caron_deep_2018}
M.~Caron, P.~Bojanowski, A.~Joulin, and M.~Douze.
\newblock Deep clustering for unsupervised learning of visual features.
\newblock In \emph{{European} {Conference} on {Computer} {Vision}}, 2018.

\bibitem[Caron et~al.(2019)Caron, Bojanowski, Mairal, and
  Joulin]{caron_unsupervised_2019}
M.~Caron, P.~Bojanowski, J.~Mairal, and A.~Joulin.
\newblock Unsupervised pre-training of image features on non-curated data.
\newblock In \emph{{International} {Conference} on {Computer} {Vision}}, 2019.

\bibitem[Caron et~al.(2020)Caron, Misra, Mairal, Goyal, Bojanowski, and
  Joulin]{swav}
M.~Caron, I.~Misra, J.~Mairal, P.~Goyal, P.~Bojanowski, and A.~Joulin.
\newblock Unsupervised learning of visual features by contrasting cluster
  assignments.
\newblock In \emph{Neural Information Processing Systems (NeurIPS)}, 2020.

\bibitem[Caron et~al.(2021)Caron, Touvron, Misra, Jégou, Mairal, Bojanowski,
  and Joulin]{caron_emerging_2021}
M.~Caron, H.~Touvron, I.~Misra, H.~Jégou, J.~Mairal, P.~Bojanowski, and
  A.~Joulin.
\newblock Emerging properties in self-supervised vision transformers.
\newblock \emph{arXiv preprint arXiv:2104.14294}, 2021.

\bibitem[Chen et~al.(2020{\natexlab{a}})Chen, Kornblith, Norouzi, and
  Hinton]{chen2020simple}
T.~Chen, S.~Kornblith, M.~Norouzi, and G.~Hinton.
\newblock A simple framework for contrastive learning of visual
  representations.
\newblock In \emph{International Conference of Machine Learning (ICML)},
  2020{\natexlab{a}}.

\bibitem[Chen et~al.(2020{\natexlab{b}})Chen, Kornblith, Swersky, Norouzi, and
  Hinton]{chen2020big}
T.~Chen, S.~Kornblith, K.~Swersky, M.~Norouzi, and G.~Hinton.
\newblock Big self-supervised models are strong semi-supervised learners.
\newblock In \emph{Neural Information Processing Systems (NeurIPS)},
  2020{\natexlab{b}}.

\bibitem[Chen and He(2020)]{simsiam}
X.~Chen and K.~He.
\newblock Exploring simple siamese representation learning.
\newblock \emph{arXiv preprint arXiv:2011.10566}, 2020.

\bibitem[Chen et~al.(2020{\natexlab{c}})Chen, Fan, Girshick, and
  He]{chen2020mocov2}
X.~Chen, H.~Fan, R.~Girshick, and K.~He.
\newblock Improved baselines with momentum contrastive learning.
\newblock \emph{arXiv preprint arXiv:2003.04297}, 2020{\natexlab{c}}.

\bibitem[Chen et~al.(2021)Chen, Xie, and He]{chen_empirical_2021}
X.~Chen, S.~Xie, and K.~He.
\newblock An empirical study of training self-supervised vision transformers.
\newblock \emph{arXiv preprint arXiv:2104.02057}, 2021.

\bibitem[Cordts et~al.(2016)Cordts, Omran, Ramos, Rehfeld, Enzweiler, Benenson,
  Franke, Roth, and Schiele]{cordts_cityscapes_2016}
M.~Cordts, M.~Omran, S.~Ramos, T.~Rehfeld, M.~Enzweiler, R.~Benenson,
  U.~Franke, S.~Roth, and B.~Schiele.
\newblock The {Cityscapes} {Dataset} for {Semantic} {Urban} {Scene}
  {Understanding}.
\newblock In \emph{Proceedings of the {IEEE} International Conference on
  Computer Vision}, pages 3213--3223, 2016.

\bibitem[Doersch and Zisserman(2017)]{doersch_multi-task_2017}
C.~Doersch and A.~Zisserman.
\newblock Multi-task {Self}-{Supervised} {Visual} {Learning}.
\newblock In \emph{Proceedings of the {IEEE}/{CVF} {International} {Conference}
  on {Computer} {Vision}}, 2017.

\bibitem[Dosovitskiy et~al.(2014)Dosovitskiy, Springenberg, Riedmiller, and
  Brox]{dosovitskiy_discriminative_2014}
A.~Dosovitskiy, J.~T. Springenberg, M.~Riedmiller, and T.~Brox.
\newblock Discriminative {Unsupervised} {Feature} {Learning} with
  {Convolutional} {Neural} {Networks}.
\newblock In \emph{Advances in {Neural} {Information} {Processing} {Systems}},
  2014.

\bibitem[Dosovitskiy et~al.(2015)Dosovitskiy, Fischer, Ilg, Hausser, Hazirbas,
  Golkov, van~der Smagt, Cremers, and Brox]{dosovitskiy_flownet_2015}
A.~Dosovitskiy, P.~Fischer, E.~Ilg, P.~Hausser, C.~Hazirbas, V.~Golkov,
  P.~van~der Smagt, D.~Cremers, and T.~Brox.
\newblock {FlowNet}: {Learning} {Optical} {Flow} {With} {Convolutional}
  {Networks}.
\newblock In \emph{Proceedings of the {IEEE} International Conference on
  Computer Vision}, pages 2758--2766, 2015.

\bibitem[Dosovitskiy et~al.(2021)Dosovitskiy, Beyer, Kolesnikov, Weissenborn,
  Zhai, Unterthiner, Dehghani, Minderer, Heigold, and
  Gelly]{vit_dosovitskiy_2021}
A.~Dosovitskiy, L.~Beyer, A.~Kolesnikov, D.~Weissenborn, X.~Zhai,
  T.~Unterthiner, M.~Dehghani, M.~Minderer, G.~Heigold, and S.~Gelly.
\newblock An image is worth 16x16 words: {Transformers} for image recognition
  at scale.
\newblock In \emph{International Conference on Learning Representations}, 2021.

\bibitem[Dvornik et~al.(2018)Dvornik, Mairal, and Schmid]{Dvornik_2018_ECCV}
N.~Dvornik, J.~Mairal, and C.~Schmid.
\newblock Modeling visual context is key to augmenting object detection
  datasets.
\newblock In \emph{Proceedings of the European Conference on Computer Vision
  (ECCV)}, September 2018.

\bibitem[Dwibedi et~al.(2017)Dwibedi, Misra, and Hebert]{Dwibedi_2017_ICCV}
D.~Dwibedi, I.~Misra, and M.~Hebert.
\newblock Cut, paste and learn: Surprisingly easy synthesis for instance
  detection.
\newblock In \emph{The IEEE International Conference on Computer Vision
  (ICCV)}, Oct 2017.

\bibitem[Fang et~al.(2019)Fang, Sun, Wang, Gou, Li, and
  Lu]{fang_instaboost_2019}
H.-S. Fang, J.~Sun, R.~Wang, M.~Gou, Y.-L. Li, and C.~Lu.
\newblock Instaboost: {Boosting} instance segmentation via probability map
  guided copy-pasting.
\newblock In \emph{Proceedings of the {IEEE}/{CVF} {International} {Conference}
  on {Computer} {Vision}}, pages 682--691, 2019.

\bibitem[Geirhos et~al.(2019)Geirhos, Rubisch, Michaelis, Bethge, Wichmann, and
  Brendel]{geirhos2019imagenettrained}
R.~Geirhos, P.~Rubisch, C.~Michaelis, M.~Bethge, F.~A. Wichmann, and
  W.~Brendel.
\newblock Imagenet-trained {CNN}s are biased towards texture; increasing shape
  bias improves accuracy and robustness.
\newblock In \emph{International Conference on Learning Representations}, 2019.

\bibitem[Geirhos et~al.(2020)Geirhos, Narayanappa, Mitzkus, Bethge, Wichmann,
  and Brendel]{geirhos2020surprising}
R.~Geirhos, K.~Narayanappa, B.~Mitzkus, M.~Bethge, F.~A. Wichmann, and
  W.~Brendel.
\newblock On the surprising similarities between supervised and self-supervised
  models.
\newblock \emph{arXiv preprint arXiv:2010.08377}, 2020.

\bibitem[Georgakis et~al.(2017)Georgakis, Mousavian, Berg, and
  Kosecka]{georgakis_synthesizing_2017}
G.~Georgakis, A.~Mousavian, A.~C. Berg, and J.~Kosecka.
\newblock Synthesizing training data for object detection in indoor scenes.
\newblock \emph{arXiv preprint arXiv:1702.07836}, 2017.

\bibitem[Ghiasi et~al.(2020)Ghiasi, Cui, Srinivas, Qian, Lin, Cubuk, Le, and
  Zoph]{ghiasi2020simple}
G.~Ghiasi, Y.~Cui, A.~Srinivas, R.~Qian, T.-Y. Lin, E.~D. Cubuk, Q.~V. Le, and
  B.~Zoph.
\newblock Simple copy-paste is a strong data augmentation method for instance
  segmentation.
\newblock \emph{arXiv preprint arXiv:2012.07177}, 2020.

\bibitem[Gidaris et~al.(2018)Gidaris, Singh, and
  Komodakis]{gidaris2018unsupervised}
S.~Gidaris, P.~Singh, and N.~Komodakis.
\newblock Unsupervised representation learning by predicting image rotations.
\newblock In \emph{International Conference on Learning Representations}, 2018.

\bibitem[Goodfellow et~al.(2015)Goodfellow, Shlens, and
  Szegedy]{goodfellow_fgsm}
I.~Goodfellow, J.~Shlens, and C.~Szegedy.
\newblock Explaining and harnessing adversarial examples.
\newblock In \emph{International Conference on Learning Representations}, 2015.

\bibitem[Gordon et~al.(2020)Gordon, Ehsani, Fox, and
  Farhadi]{gordon_watching_2020}
D.~Gordon, K.~Ehsani, D.~Fox, and A.~Farhadi.
\newblock Watching the world go by: {Representation} learning from unlabeled
  videos.
\newblock \emph{arXiv preprint arXiv:2003.07990}, 2020.

\bibitem[Goyal et~al.(2018)Goyal, Dollár, Girshick, Noordhuis, Wesolowski,
  Kyrola, Tulloch, Jia, and He]{goyal_accurate_2018}
P.~Goyal, P.~Dollár, R.~Girshick, P.~Noordhuis, L.~Wesolowski, A.~Kyrola,
  A.~Tulloch, Y.~Jia, and K.~He.
\newblock Accurate, {Large} {Minibatch} {SGD}: {Training} {ImageNet} in 1
  {Hour}.
\newblock \emph{arXiv:1706.02677 [cs]}, Apr. 2018.

\bibitem[Grill et~al.(2020)Grill, Strub, Altch{\'e}, Tallec, Richemond,
  Buchatskaya, Doersch, Pires, Guo, Azar, et~al.]{grill2020bootstrap}
J.-B. Grill, F.~Strub, F.~Altch{\'e}, C.~Tallec, P.~H. Richemond,
  E.~Buchatskaya, C.~Doersch, B.~A. Pires, Z.~D. Guo, M.~G. Azar, et~al.
\newblock Bootstrap your own latent: A new approach to self-supervised
  learning.
\newblock \emph{Neural Information Processing Systems (NeurIPS)}, 2020.

\bibitem[Gupta et~al.(2016)Gupta, Vedaldi, and Zisserman]{gupta_synthetic_2016}
A.~Gupta, A.~Vedaldi, and A.~Zisserman.
\newblock Synthetic data for text localisation in natural images.
\newblock In \emph{Proceedings of the {IEEE} Conference on Computer Vision and
  Pattern Recognition}, 2016.

\bibitem[Hadsell et~al.(2006)Hadsell, Chopra, and
  LeCun]{hadsell_dimensionality_2006}
R.~Hadsell, S.~Chopra, and Y.~LeCun.
\newblock Dimensionality reduction by learning an invariant mapping.
\newblock In \emph{{Computer} {Vision} and {Pattern} {Recognition}}, 2006.

\bibitem[Han et~al.(2019)Han, Xie, and Zisserman]{han_video_2019}
T.~Han, W.~Xie, and A.~Zisserman.
\newblock Video representation learning by dense predictive coding.
\newblock In \emph{Proceedings of the {IEEE}/{CVF} {International} {Conference}
  on {Computer} {Vision} {Workshops}}, 2019.

\bibitem[Harris et~al.(2020)Harris, Millman, van~der Walt, Gommers, Virtanen,
  Cournapeau, Wieser, Taylor, Berg, Smith, Kern, Picus, Hoyer, van Kerkwijk,
  Brett, Haldane, del R{\'{i}}o, Wiebe, Peterson, G{\'{e}}rard-Marchant,
  Sheppard, Reddy, Weckesser, Abbasi, Gohlke, and Oliphant]{numpy}
C.~R. Harris, K.~J. Millman, S.~J. van~der Walt, R.~Gommers, P.~Virtanen,
  D.~Cournapeau, E.~Wieser, J.~Taylor, S.~Berg, N.~J. Smith, R.~Kern, M.~Picus,
  S.~Hoyer, M.~H. van Kerkwijk, M.~Brett, A.~Haldane, J.~F. del R{\'{i}}o,
  M.~Wiebe, P.~Peterson, P.~G{\'{e}}rard-Marchant, K.~Sheppard, T.~Reddy,
  W.~Weckesser, H.~Abbasi, C.~Gohlke, and T.~E. Oliphant.
\newblock Array programming with {NumPy}.
\newblock \emph{Nature}, 585\penalty0 (7825):\penalty0 357--362, Sept. 2020.
\newblock \doi{10.1038/s41586-020-2649-2}.

\bibitem[He et~al.(2016)He, Zhang, Ren, and Sun]{resnet}
K.~He, X.~Zhang, S.~Ren, and J.~Sun.
\newblock Deep {Residual} {Learning} for {Image} {Recognition}.
\newblock In \emph{Proceedings of the {IEEE}/{CVF} {Conference} on {Computer}
  {Vision} and {Pattern} {Recognition}}, Dec. 2016.

\bibitem[He et~al.(2017)He, Gkioxari, Dollar, and Girshick]{he_mask_2017}
K.~He, G.~Gkioxari, P.~Dollar, and R.~Girshick.
\newblock Mask {R}-{CNN}.
\newblock In \emph{Proceedings of the {IEEE} International Conference on
  Computer Vision}, pages 2961--2969, 2017.

\bibitem[He et~al.(2020)He, Fan, Wu, Xie, and Girshick]{he2019moco}
K.~He, H.~Fan, Y.~Wu, S.~Xie, and R.~Girshick.
\newblock Momentum contrast for unsupervised visual representation learning.
\newblock In \emph{Conference on Computer Vision and Pattern Recognition},
  2020.

\bibitem[H{\'e}naff et~al.(2020)H{\'e}naff, Srinivas, De~Fauw, Razavi, Doersch,
  Eslami, and Oord]{henaff_data-efficient_2020}
O.~H{\'e}naff, A.~Srinivas, J.~De~Fauw, A.~Razavi, C.~Doersch, S.~M.~A. Eslami,
  and A.~v.~d. Oord.
\newblock Data-efficient image recognition with contrastive predictive coding.
\newblock In \emph{International {Conference} on {Machine} {Learning}}, 2020.

\bibitem[Hendrycks and Dietterich(2019)]{hendrycks_benchmarking_2019}
D.~Hendrycks and T.~Dietterich.
\newblock Benchmarking neural network robustness to common corruptions and
  perturbations.
\newblock In \emph{International Conference on Learning Representations}, 2019.

\bibitem[Hendrycks et~al.(2019{\natexlab{a}})Hendrycks, Lee, and
  Mazeika]{hendrycks_using_2019}
D.~Hendrycks, K.~Lee, and M.~Mazeika.
\newblock Using pre-training can improve model robustness and uncertainty.
\newblock In \emph{International {Conference} on {Machine} {Learning}},
  2019{\natexlab{a}}.

\bibitem[Hendrycks et~al.(2019{\natexlab{b}})Hendrycks, Zhao, Basart,
  Steinhardt, and Song]{hendrycks2019nae}
D.~Hendrycks, K.~Zhao, S.~Basart, J.~Steinhardt, and D.~Song.
\newblock Natural adversarial examples.
\newblock \emph{arXiv preprint arXiv:1907.07174}, 2019{\natexlab{b}}.

\bibitem[Hendrycks et~al.(2021)Hendrycks, Basart, Mu, Kadavath, Wang, Dorundo,
  Desai, Zhu, Parajuli, Guo, Song, Steinhardt, and Gilmer]{hendrycks_many_2021}
D.~Hendrycks, S.~Basart, N.~Mu, S.~Kadavath, F.~Wang, E.~Dorundo, R.~Desai,
  T.~Zhu, S.~Parajuli, M.~Guo, D.~Song, J.~Steinhardt, and J.~Gilmer.
\newblock The {Many} {Faces} of {Robustness}: {A} {Critical} {Analysis} of
  {Out}-of-{Distribution} {Generalization}.
\newblock In \emph{Proceedings of the {IEEE} International Conference on
  Computer Vision}, 2021.

\bibitem[Hermann et~al.(2020)Hermann, Chen, and Kornblith]{hermann2019origins}
K.~L. Hermann, T.~Chen, and S.~Kornblith.
\newblock The origins and prevalence of texture bias in convolutional neural
  networks.
\newblock \emph{Neural Information Processing Systems (NeurIPS)}, 2020.

\bibitem[Hjelm et~al.(2019)Hjelm, Fedorov, Lavoie-Marchildon, Grewal, Bachman,
  Trischler, and Bengio]{hjelm_learning_2019}
R.~D. Hjelm, A.~Fedorov, S.~Lavoie-Marchildon, K.~Grewal, P.~Bachman,
  A.~Trischler, and Y.~Bengio.
\newblock Learning deep representations by mutual information estimation and
  maximization.
\newblock In \emph{International Conference on Learning Representations}, 2019.

\bibitem[Hou et~al.(2017)Hou, Cheng, Hu, Borji, Tu, and Torr]{hou_deeply_2017}
Q.~Hou, M.-M. Cheng, X.~Hu, A.~Borji, Z.~Tu, and P.~H. Torr.
\newblock Deeply supervised salient object detection with short connections.
\newblock In \emph{Proceedings of the {IEEE} conference on Computer Vision and
  Pattern Recognition}, pages 3203--3212, 2017.

\bibitem[Huynh et~al.(2020)Huynh, Kornblith, Walter, Maire, and Khademi]{fnc}
T.~Huynh, S.~Kornblith, M.~R. Walter, M.~Maire, and M.~Khademi.
\newblock Boosting contrastive self-supervised learning with false negative
  cancellation.
\newblock \emph{arXiv preprint arXiv:2011.11765}, 2020.

\bibitem[Ilyas et~al.(2019)Ilyas, Santurkar, Tsipras, Engstrom, Tran, and
  Madry]{ilyas_neurips_2020_features_not_bugs}
A.~Ilyas, S.~Santurkar, D.~Tsipras, L.~Engstrom, B.~Tran, and A.~Madry.
\newblock Adversarial examples are not bugs, they are features.
\newblock In H.~Wallach, H.~Larochelle, A.~Beygelzimer, F.~d\textquotesingle
  Alch\'{e}-Buc, E.~Fox, and R.~Garnett, editors, \emph{Advances in Neural
  Information Processing Systems}, volume~32, pages 125--136. Curran
  Associates, Inc., 2019.

\bibitem[Ioffe and Szegedy(2015)]{ioffe_batch_2015}
S.~Ioffe and C.~Szegedy.
\newblock Batch {Normalization}: {Accelerating} {Deep} {Network} {Training} by
  {Reducing} {Internal} {Covariate} {Shift}.
\newblock In \emph{International {Conference} on {Machine} {Learning}}, pages
  448--456, 2015.

\bibitem[Ji et~al.(2019)Ji, Henriques, and Vedaldi]{ji_invariant_2019}
X.~Ji, J.~F. Henriques, and A.~Vedaldi.
\newblock Invariant {Information} {Clustering} for {Unsupervised} {Image}
  {Classification} and {Segmentation}.
\newblock In \emph{Proceedings of the {IEEE}/{CVF} {International} {Conference}
  on {Computer} {Vision}}, 2019.

\bibitem[Jing and Tian(2020)]{ssl_review}
L.~Jing and Y.~Tian.
\newblock Self-supervised visual feature learning with deep neural networks: A
  survey.
\newblock \emph{IEEE Transactions on Pattern Analysis and Machine
  Intelligence}, 2020.

\bibitem[Jo and Bengio(2017)]{jo2017measuring}
J.~Jo and Y.~Bengio.
\newblock Measuring the tendency of cnns to learn surface statistical
  regularities.
\newblock \emph{arXiv preprint arXiv:1711.11561}, 2017.

\bibitem[Kalantidis et~al.(2020)Kalantidis, Sariyildiz, Pion, Weinzaepfel, and
  Larlus]{kalantidis2020hard}
Y.~Kalantidis, M.~B. Sariyildiz, N.~Pion, P.~Weinzaepfel, and D.~Larlus.
\newblock Hard negative mixing for contrastive learning.
\newblock In \emph{Neural Information Processing Systems (NeurIPS)}, 2020.

\bibitem[Kolesnikov et~al.(2019)Kolesnikov, Zhai, and
  Beyer]{kolesnikov_revisiting_2019}
A.~Kolesnikov, X.~Zhai, and L.~Beyer.
\newblock Revisiting {Self}-{Supervised} {Visual} {Representation} {Learning}.
\newblock In \emph{The IEEE Conference on Computer Vision and Pattern
  Recognition (CVPR)}, 2019.

\bibitem[Krizhevsky et~al.(2012)Krizhevsky, Sutskever, and Hinton]{alexnet}
A.~Krizhevsky, I.~Sutskever, and G.~E. Hinton.
\newblock {ImageNet} {Classification} with {Deep} {Convolutional} {Neural}
  {Networks}.
\newblock In \emph{Advances in {Neural} {Information} {Processing} {Systems}},
  2012.

\bibitem[Kurakin et~al.(2016)Kurakin, Goodfellow, and
  Bengio]{kurakin_adversarial_2016}
A.~Kurakin, I.~J. Goodfellow, and S.~Bengio.
\newblock Adversarial examples in the physical world.
\newblock \emph{ICLR 2017 Workshop}, Nov. 2016.

\bibitem[Li et~al.(2021{\natexlab{a}})Li, Yang, Zhang, Gao, Xiao, Dai, Yuan,
  and Gao]{li_efficient_2021}
C.~Li, J.~Yang, P.~Zhang, M.~Gao, B.~Xiao, X.~Dai, L.~Yuan, and J.~Gao.
\newblock Efficient {Self}-supervised {Vision} {Transformers} for
  {Representation} {Learning}.
\newblock \emph{arXiv preprint arXiv:2106.09785}, June 2021{\natexlab{a}}.

\bibitem[Li et~al.(2021{\natexlab{b}})Li, Zhou, Xiong, and
  Hoi]{li_prototypical_2021}
J.~Li, P.~Zhou, C.~Xiong, and S.~Hoi.
\newblock Prototypical {Contrastive} {Learning} of {Unsupervised}
  {Representations}.
\newblock In \emph{International Conference on Learning Representations},
  2021{\natexlab{b}}.

\bibitem[Li et~al.(2016)Li, Zhao, Wei, Yang, Wu, Zhuang, Ling, and
  Wang]{li_deepsaliency_2016}
X.~Li, L.~Zhao, L.~Wei, M.-H. Yang, F.~Wu, Y.~Zhuang, H.~Ling, and J.~Wang.
\newblock Deepsaliency: {Multi}-task deep neural network model for salient
  object detection.
\newblock \emph{IEEE transactions on Image Processing}, 25\penalty0
  (8):\penalty0 3919--3930, 2016.
\newblock Publisher: IEEE.

\bibitem[Liu et~al.(2011)Liu, Yuan, Sun, Wang, Zheng, Tang, and
  Shum]{liu_learning_2011}
T.~Liu, Z.~Yuan, J.~Sun, J.~Wang, N.~Zheng, X.~Tang, and H.-Y. Shum.
\newblock Learning to {Detect} a {Salient} {Object}.
\newblock \emph{IEEE Transactions on Pattern Analysis and Machine
  Intelligence}, 33\penalty0 (2):\penalty0 353--367, Feb. 2011.
\newblock ISSN 1939-3539.
\newblock \doi{10.1109/TPAMI.2010.70}.

\bibitem[Loshchilov and Hutter(2017)]{loshchilov_sgdr_2016}
I.~Loshchilov and F.~Hutter.
\newblock Sgdr: {Stochastic} gradient descent with warm restarts.
\newblock In \emph{International Conference on Learning Representations}, 2017.

\bibitem[Luo et~al.(2017)Luo, Mishra, Achkar, Eichel, Li, and
  Jodoin]{luo_non-local_2017}
Z.~Luo, A.~Mishra, A.~Achkar, J.~Eichel, S.~Li, and P.-M. Jodoin.
\newblock Non-local deep features for salient object detection.
\newblock In \emph{Proceedings of the {IEEE} {Conference} on Computer Vision
  and Pattern Recognition}, pages 6609--6617, 2017.

\bibitem[Madry et~al.(2018)Madry, Makelov, Schmidt, Tsipras, and
  Vladu]{madry2018towards}
A.~Madry, A.~Makelov, L.~Schmidt, D.~Tsipras, and A.~Vladu.
\newblock Towards deep learning models resistant to adversarial attacks.
\newblock In \emph{International Conference on Learning Representations}, 2018.

\bibitem[Marcel and Rodriguez(2010)]{marcel_torchvision_2010}
S.~Marcel and Y.~Rodriguez.
\newblock Torchvision the machine-vision package of torch.
\newblock In \emph{Proceedings of the 18th {ACM} International Conference on
  {Multimedia}}, 2010.

\bibitem[Misra and van~der Maaten(2020)]{misra2020pirl}
I.~Misra and L.~van~der Maaten.
\newblock Self-supervised learning of pretext-invariant representations.
\newblock In \emph{Proceedings of the {IEEE} {Conference} on Computer Vision
  and Pattern Recognition}, 2020.

\bibitem[Nado et~al.(2020)Nado, Padhy, Sculley, D'Amour, Lakshminarayanan, and
  Snoek]{nado_evaluating_2020}
Z.~Nado, S.~Padhy, D.~Sculley, A.~D'Amour, B.~Lakshminarayanan, and J.~Snoek.
\newblock Evaluating prediction-time batch normalization for robustness under
  covariate shift.
\newblock \emph{arXiv preprint arXiv:2006.10963}, 2020.

\bibitem[Nguyen et~al.(2019)Nguyen, Dax, Mummadi, Ngo, Nguyen, Lou, and
  Brox]{nguyen_deepusps_2019}
D.~T. Nguyen, M.~Dax, C.~K. Mummadi, T.~P.~N. Ngo, T.~H.~P. Nguyen, Z.~Lou, and
  T.~Brox.
\newblock Deepusps: {Deep} robust unsupervised saliency prediction with
  self-supervision.
\newblock \emph{arXiv preprint arXiv:1909.13055}, 2019.

\bibitem[Noroozi and Favaro(2016)]{jigsaw_ssl}
M.~Noroozi and P.~Favaro.
\newblock Unsupervised learning of visual representations by solving jigsaw
  puzzles.
\newblock In \emph{European Conference on Computer Vision}. Springer, 2016.

\bibitem[Oord et~al.(2018)Oord, Li, and Vinyals]{infonce_2018}
A.~v.~d. Oord, Y.~Li, and O.~Vinyals.
\newblock Representation {Learning} with {Contrastive} {Predictive} {Coding}.
\newblock \emph{arXiv:1807.03748 [cs, stat]}, Jan. 2018.

\bibitem[Paszke et~al.(2019)Paszke, Gross, Massa, Lerer, Bradbury, Chanan,
  Killeen, Lin, Gimelshein, Antiga, Desmaison, Kopf, Yang, DeVito, Raison,
  Tejani, Chilamkurthy, Steiner, Fang, Bai, and Chintala]{pytorch_cite}
A.~Paszke, S.~Gross, F.~Massa, A.~Lerer, J.~Bradbury, G.~Chanan, T.~Killeen,
  Z.~Lin, N.~Gimelshein, L.~Antiga, A.~Desmaison, A.~Kopf, E.~Yang, Z.~DeVito,
  M.~Raison, A.~Tejani, S.~Chilamkurthy, B.~Steiner, L.~Fang, J.~Bai, and
  S.~Chintala.
\newblock Pytorch: An imperative style, high-performance deep learning library.
\newblock In H.~Wallach, H.~Larochelle, A.~Beygelzimer, F.~d\textquotesingle
  Alch\'{e}-Buc, E.~Fox, and R.~Garnett, editors, \emph{Advances in Neural
  Information Processing Systems 32}, pages 8024--8035. Curran Associates,
  Inc., 2019.

\bibitem[Pathak et~al.(2016)Pathak, Krahenbuhl, Donahue, Darrell, and
  Efros]{inpainting_ssl}
D.~Pathak, P.~Krahenbuhl, J.~Donahue, T.~Darrell, and A.~A. Efros.
\newblock Context {Encoders}: {Feature} {Learning} by {Inpainting}.
\newblock In \emph{{Computer} {Vision} and {Pattern} {Recognition}}, 2016.

\bibitem[Purushwalkam and Gupta(2020)]{purushwalkam2020demystifying}
S.~Purushwalkam and A.~Gupta.
\newblock Demystifying contrastive self-supervised learning: Invariances,
  augmentations and dataset biases.
\newblock \emph{Neural Information Processing Systems (NeurIPS)}, 2020.

\bibitem[Qin et~al.(2020)Qin, Zhang, Huang, Dehghan, Zaiane, and
  Jagersand]{u2net}
X.~Qin, Z.~Zhang, C.~Huang, M.~Dehghan, O.~Zaiane, and M.~Jagersand.
\newblock U2-net: Going deeper with nested u-structure for salient object
  detection.
\newblock \emph{Pattern Recognition}, 106:\penalty0 107404, 2020.

\bibitem[Rauber et~al.(2020)Rauber, Zimmermann, Bethge, and
  Brendel]{foolbox_rauber2017}
J.~Rauber, R.~Zimmermann, M.~Bethge, and W.~Brendel.
\newblock Foolbox native: Fast adversarial attacks to benchmark the robustness
  of machine learning models in pytorch, tensorflow, and jax.
\newblock \emph{Journal of Open Source Software}, 5\penalty0 (53):\penalty0
  2607, 2020.

\bibitem[Recht et~al.(2019)Recht, Roelofs, Schmidt, and Shankar]{imagenetv2}
B.~Recht, R.~Roelofs, L.~Schmidt, and V.~Shankar.
\newblock Do {ImageNet} {Classifiers} {Generalize} to {ImageNet}?
\newblock In \emph{{International} {Conference} on {Machine} {Learning}}, 2019.

\bibitem[Remez et~al.(2018)Remez, Huang, and Brown]{remez_learning_2018}
T.~Remez, J.~Huang, and M.~Brown.
\newblock Learning to segment via cut-and-paste.
\newblock In \emph{Proceedings of the {European} conference on computer vision
  ({ECCV})}, 2018.

\bibitem[Ren et~al.(2015)Ren, He, Girshick, and Sun]{ren_faster_2015}
S.~Ren, K.~He, R.~Girshick, and J.~Sun.
\newblock Faster {R}-{CNN}: {Towards} {Real}-{Time} {Object} {Detection} with
  {Region} {Proposal} {Networks}.
\newblock In \emph{Advances in {Neural} {Information} {Processing} {Systems}},
  volume~28, 2015.

\bibitem[Robinson et~al.(2021)Robinson, Chuang, Sra, and
  Jegelka]{robinson2021contrastive}
J.~D. Robinson, C.-Y. Chuang, S.~Sra, and S.~Jegelka.
\newblock Contrastive learning with hard negative samples.
\newblock In \emph{International Conference on Learning Representations}, 2021.

\bibitem[Russakovsky et~al.(2015)Russakovsky, Deng, Su, Krause, Satheesh, Ma,
  Huang, Karpathy, Khosla, Bernstein, Berg, and Fei-Fei]{imnet1k}
O.~Russakovsky, J.~Deng, H.~Su, J.~Krause, S.~Satheesh, S.~Ma, Z.~Huang,
  A.~Karpathy, A.~Khosla, M.~Bernstein, A.~C. Berg, and L.~Fei-Fei.
\newblock {ImageNet} {Large} {Scale} {Visual} {Recognition} {Challenge}.
\newblock \emph{International Journal of Computer Vision}, 115\penalty0
  (3):\penalty0 211--252, Dec. 2015.
\newblock ISSN 1573-1405.

\bibitem[Schneider et~al.(2020)Schneider, Rusak, Eck, Bringmann, Brendel, and
  Bethge]{schneider_improving_2020}
S.~Schneider, E.~Rusak, L.~Eck, O.~Bringmann, W.~Brendel, and M.~Bethge.
\newblock Improving robustness against common corruptions by covariate shift
  adaptation.
\newblock In \emph{Advances in Neural Information Processing Systems}, 2020.

\bibitem[Sehwag et~al.(2020)Sehwag, Oak, Chiang, and Mittal]{sehwag2020time}
V.~Sehwag, R.~Oak, M.~Chiang, and P.~Mittal.
\newblock Time for a background check! uncovering the impact of background
  features on deep neural networks.
\newblock \emph{arXiv preprint arXiv:2006.14077}, 2020.

\bibitem[Selvaraju et~al.(2017)Selvaraju, Cogswell, Das, Vedantam, Parikh, and
  Batra]{grad-cam_2017}
R.~R. Selvaraju, M.~Cogswell, A.~Das, R.~Vedantam, D.~Parikh, and D.~Batra.
\newblock Grad-cam: {Visual} explanations from deep networks via gradient-based
  localization.
\newblock In \emph{Proceedings of the {IEEE} International Conference on
  Computer Vision}, pages 618--626, 2017.

\bibitem[Selvaraju et~al.(2020)Selvaraju, Desai, Johnson, and
  Naik]{selvaraju2020casting}
R.~R. Selvaraju, K.~Desai, J.~Johnson, and N.~Naik.
\newblock Casting your model: Learning to localize improves self-supervised
  representations.
\newblock \emph{arXiv preprint arXiv:2012.04630}, 2020.

\bibitem[Sermanet et~al.(2018)Sermanet, Lynch, Chebotar, Hsu, Jang, Schaal, and
  Levine]{sermanet_time-contrastive_2018}
P.~Sermanet, C.~Lynch, Y.~Chebotar, J.~Hsu, E.~Jang, S.~Schaal, and S.~Levine.
\newblock Time-contrastive networks: {Self}-supervised learning from video.
\newblock In \emph{2018 {IEEE} international conference on robotics and
  automation ({ICRA})}, 2018.

\bibitem[Simonyan et~al.(2013)Simonyan, Vedaldi, and
  Zisserman]{simonyan_deep_2013}
K.~Simonyan, A.~Vedaldi, and A.~Zisserman.
\newblock Deep {Inside} {Convolutional} {Networks}: {Visualising} {Image}
  {Classification} {Models} and {Saliency} {Maps}.
\newblock \emph{arXiv:1312.6034 [cs]}, 2013.

\bibitem[Sohn et~al.(2020)Sohn, Berthelot, Li, Zhang, Carlini, Cubuk, Kurakin,
  Zhang, and Raffel]{fixmatch_sohn_2020}
K.~Sohn, D.~Berthelot, C.-L. Li, Z.~Zhang, N.~Carlini, E.~D. Cubuk, A.~Kurakin,
  H.~Zhang, and C.~Raffel.
\newblock Fixmatch: {Simplifying} semi-supervised learning with consistency and
  confidence.
\newblock In \emph{Neural Information Processing Systems (NeurIPS)}, 2020.

\bibitem[Stock and Cisse(2018)]{stock_convnets_2018}
P.~Stock and M.~Cisse.
\newblock Convnets and imagenet beyond accuracy: {Understanding} mistakes and
  uncovering biases.
\newblock In \emph{Proceedings of the {European} {Conference} on {Computer}
  {Vision} ({ECCV})}, 2018.

\bibitem[Tamkin et~al.(2021)Tamkin, Wu, and Goodman]{tamkin2021viewmaker}
A.~Tamkin, M.~Wu, and N.~Goodman.
\newblock Viewmaker networks: Learning views for unsupervised representation
  learning.
\newblock In \emph{International Conference on Learning Representations}, 2021.

\bibitem[Tian et~al.(2020{\natexlab{a}})Tian, Krishnan, and
  Isola]{tian_contrastive_2020}
Y.~Tian, D.~Krishnan, and P.~Isola.
\newblock Contrastive multiview coding.
\newblock In \emph{European Conference on Computer Vision}, pages 776--794,
  2020{\natexlab{a}}.

\bibitem[Tian et~al.(2020{\natexlab{b}})Tian, Sun, Poole, Krishnan, Schmid, and
  Isola]{tian_what_2020}
Y.~Tian, C.~Sun, B.~Poole, D.~Krishnan, C.~Schmid, and P.~Isola.
\newblock What {Makes} for {Good} {Views} for {Contrastive} {Learning}?
\newblock In \emph{Neural Information Processing Systems (NeurIPS)}, Dec.
  2020{\natexlab{b}}.

\bibitem[Vincent et~al.(2008)Vincent, Larochelle, Bengio, and
  Manzagol]{vincent_extracting_2008}
P.~Vincent, H.~Larochelle, Y.~Bengio, and P.-A. Manzagol.
\newblock Extracting and composing robust features with denoising autoencoders.
\newblock In \emph{International Conference on {Machine} Learning}, 2008.

\bibitem[Wang et~al.(2016)Wang, Wang, Lu, Zhang, and Ruan]{wang_saliency_2016}
L.~Wang, L.~Wang, H.~Lu, P.~Zhang, and X.~Ruan.
\newblock Saliency detection with recurrent fully convolutional networks.
\newblock In \emph{European conference on computer vision}, pages 825--841.
  Springer, 2016.

\bibitem[Wang et~al.(2017{\natexlab{a}})Wang, Lu, Wang, Feng, Wang, Yin, and
  Ruan]{duts_tr_wang2017}
L.~Wang, H.~Lu, Y.~Wang, M.~Feng, D.~Wang, B.~Yin, and X.~Ruan.
\newblock Learning to detect salient objects with image-level supervision.
\newblock In \emph{Proceedings of the {IEEE} conference on Computer Vision and
  Pattern Recognition (CVPR)}, 2017{\natexlab{a}}.

\bibitem[Wang et~al.(2017{\natexlab{b}})Wang, Borji, Zhang, Zhang, and
  Lu]{wang_stagewise_2017}
T.~Wang, A.~Borji, L.~Zhang, P.~Zhang, and H.~Lu.
\newblock A stagewise refinement model for detecting salient objects in images.
\newblock In \emph{Proceedings of the {IEEE} {International} {Conference} on
  {Computer} {Vision}}, pages 4019--4028, 2017{\natexlab{b}}.

\bibitem[Wu et~al.(2021)Wu, Mosse, Zhuang, Yamins, and
  Goodman]{wu2021conditional}
M.~Wu, M.~Mosse, C.~Zhuang, D.~Yamins, and N.~Goodman.
\newblock Conditional negative sampling for contrastive learning of visual
  representations.
\newblock In \emph{International Conference on Learning Representations}, 2021.

\bibitem[Wu et~al.(2019)Wu, Kirillov, Massa, Lo, and
  Girshick]{wu2019detectron2}
Y.~Wu, A.~Kirillov, F.~Massa, W.-Y. Lo, and R.~Girshick.
\newblock Detectron2.
\newblock \url{https://github.com/facebookresearch/detectron2}, 2019.

\bibitem[Wu et~al.(2018)Wu, Xiong, Yu, and Lin]{wu_unsupervised_2018}
Z.~Wu, Y.~Xiong, S.~X. Yu, and D.~Lin.
\newblock Unsupervised feature learning via non-parametric instance
  discrimination.
\newblock In \emph{Proceedings of the {IEEE} conference on Computer Vision and
  Pattern Recognition}, pages 3733--3742, 2018.

\bibitem[Xiao et~al.(2021{\natexlab{a}})Xiao, Engstrom, Ilyas, and
  Madry]{xiao2020noise}
K.~Xiao, L.~Engstrom, A.~Ilyas, and A.~Madry.
\newblock Noise or signal: The role of image backgrounds in object recognition.
\newblock In \emph{International Conference on Learning Representations
  (ICLR)}, 2021{\natexlab{a}}.

\bibitem[Xiao et~al.(2021{\natexlab{b}})Xiao, Wang, Efros, and
  Darrell]{xiao_multihead_2021}
T.~Xiao, X.~Wang, A.~A. Efros, and T.~Darrell.
\newblock What should not be contrastive in contrastive learning.
\newblock In \emph{International Conference on Learning Representations},
  2021{\natexlab{b}}.

\bibitem[Yamins et~al.(2014)Yamins, Hong, Cadieu, Solomon, Seibert, and
  DiCarlo]{yamins_pnas_2014}
D.~L.~K. Yamins, H.~Hong, C.~F. Cadieu, E.~A. Solomon, D.~Seibert, and J.~J.
  DiCarlo.
\newblock Performance-optimized hierarchical models predict neural responses in
  higher visual cortex.
\newblock \emph{Proceedings of the National Academy of Sciences}, 111\penalty0
  (23):\penalty0 8619--8624, June 2014.

\bibitem[Yan et~al.(2013)Yan, Xu, Shi, and Jia]{yan_hierarchical_2013}
Q.~Yan, L.~Xu, J.~Shi, and J.~Jia.
\newblock Hierarchical {Saliency} {Detection}.
\newblock In \emph{Proceedings of the {IEEE} {Conference} on Computer Vision
  and Pattern Recognition}, pages 1155--1162, 2013.

\bibitem[Ye et~al.(2019)Ye, Zhang, Yuen, and Chang]{ye_unsupervised_2019}
M.~Ye, X.~Zhang, P.~C. Yuen, and S.-F. Chang.
\newblock Unsupervised embedding learning via invariant and spreading instance
  feature.
\newblock In \emph{{Computer} {Vision} and {Pattern} {Recognition}}, 2019.

\bibitem[You et~al.(2017)You, Gitman, and Ginsburg]{you2017large}
Y.~You, I.~Gitman, and B.~Ginsburg.
\newblock Large batch training of convolutional networks.
\newblock \emph{arXiv preprint arXiv:1708.03888}, 2017.

\bibitem[Yu et~al.(2017)Yu, Koltun, and Funkhouser]{drn_yu_2017}
F.~Yu, V.~Koltun, and T.~Funkhouser.
\newblock Dilated residual networks.
\newblock In \emph{Proceedings of the {IEEE} conference on Computer Vision and
  Pattern Recognition (CVPR)}, 2017.

\bibitem[Yun et~al.(2019)Yun, Han, Oh, Chun, Choe, and Yoo]{yun_cutmix_2019}
S.~Yun, D.~Han, S.~J. Oh, S.~Chun, J.~Choe, and Y.~Yoo.
\newblock {CutMix}: {Regularization} {Strategy} to {Train} {Strong}
  {Classifiers} with {Localizable} {Features}.
\newblock In \emph{Proceedings of the {IEEE} International Conference on
  Computer Vision}, 2019.

\bibitem[Zbontar et~al.(2021)Zbontar, Jing, Misra, LeCun, and
  Deny]{zbontar_barlow_2021}
J.~Zbontar, L.~Jing, I.~Misra, Y.~LeCun, and S.~Deny.
\newblock Barlow twins: {Self}-supervised learning via redundancy reduction.
\newblock In \emph{International Conference of Machine Learning (ICML)}, 2021.

\bibitem[Zhai et~al.(2019)Zhai, Oliver, Kolesnikov, and Beyer]{zhai_s4l_2019}
X.~Zhai, A.~Oliver, A.~Kolesnikov, and L.~Beyer.
\newblock S4l: {Self}-supervised semi-supervised learning.
\newblock In \emph{Proceedings of the {IEEE}/{CVF} {International} {Conference}
  on {Computer} {Vision}}, pages 1476--1485, 2019.

\bibitem[Zhang et~al.(2017{\natexlab{a}})Zhang, Han, and
  Zhang]{zhang_supervision_2017}
D.~Zhang, J.~Han, and Y.~Zhang.
\newblock Supervision by {Fusion}: {Towards} {Unsupervised} {Learning} of
  {Deep} {Salient} {Object} {Detector}.
\newblock In \emph{Proceedings of the {IEEE} International Conference on
  Computer Vision}, pages 4048--4056, 2017{\natexlab{a}}.

\bibitem[Zhang et~al.(2018{\natexlab{a}})Zhang, Cisse, Dauphin, and
  Lopez-Paz]{mixup}
H.~Zhang, M.~Cisse, Y.~N. Dauphin, and D.~Lopez-Paz.
\newblock mixup: {Beyond} empirical risk minimization.
\newblock In \emph{International Conference on Learning Representations},
  2018{\natexlab{a}}.

\bibitem[Zhang et~al.(2018{\natexlab{b}})Zhang, Zhang, Dai, Harandi, and
  Hartley]{zhang_deep_2018}
J.~Zhang, T.~Zhang, Y.~Dai, M.~Harandi, and R.~Hartley.
\newblock Deep {Unsupervised} {Saliency} {Detection}: {A} {Multiple} {Noisy}
  {Labeling} {Perspective}.
\newblock In \emph{Proceedings of the {IEEE} {Conference} on Computer Vision
  and Pattern Recognition}, pages 9029--9038, 2018{\natexlab{b}}.

\bibitem[Zhang et~al.(2017{\natexlab{b}})Zhang, Wang, Lu, Wang, and
  Ruan]{zhang_amulet_2017}
P.~Zhang, D.~Wang, H.~Lu, H.~Wang, and X.~Ruan.
\newblock Amulet: {Aggregating} multi-level convolutional features for salient
  object detection.
\newblock In \emph{Proceedings of the {IEEE} {International} {Conference} on
  {Computer} {Vision}}, pages 202--211, 2017{\natexlab{b}}.

\bibitem[Zhang et~al.(2017{\natexlab{c}})Zhang, Wang, Lu, Wang, and
  Yin]{zhang_learning_2017}
P.~Zhang, D.~Wang, H.~Lu, H.~Wang, and B.~Yin.
\newblock Learning uncertain convolutional features for accurate saliency
  detection.
\newblock In \emph{Proceedings of the {IEEE} {International} {Conference} on
  Computer Vision}, pages 212--221, 2017{\natexlab{c}}.

\bibitem[Zhang et~al.(2016)Zhang, Isola, and Efros]{colorization_ssl}
R.~Zhang, P.~Isola, and A.~A. Efros.
\newblock Colorful image colorization.
\newblock In \emph{European Conference on Computer Vision}, pages 649--666.
  Springer, 2016.

\bibitem[Zhang et~al.(2017{\natexlab{d}})Zhang, Isola, and
  Efros]{split-brain_2017}
R.~Zhang, P.~Isola, and A.~A. Efros.
\newblock Split-brain autoencoders: {Unsupervised} learning by cross-channel
  prediction.
\newblock In \emph{{Computer} {Vision} and {Pattern} {Recognition}},
  2017{\natexlab{d}}.

\bibitem[Zhao et~al.(2021)Zhao, Wu, Lau, and Lin]{ZhaoAAAI2021}
N.~Zhao, Z.~Wu, R.~W. Lau, and S.~Lin.
\newblock Distilling localization for self-supervised representation learning.
\newblock In \emph{Proceedings of the AAAI Conference on Artificial
  Intelligence}, 2021.

\bibitem[Zhuang et~al.(2019)Zhuang, Zhai, and Yamins]{zhuang_local_2019}
C.~Zhuang, A.~L. Zhai, and D.~Yamins.
\newblock Local aggregation for unsupervised learning of visual embeddings.
\newblock In \emph{Proceedings of the {IEEE}/{CVF} {International} {Conference}
  on {Computer} {Vision}}, pages 6002--6012, 2019.

\bibitem[Zhuang et~al.(2020)Zhuang, She, Andonian, Mark, and
  Yamins]{zhuang_unsupervised_2020}
C.~Zhuang, T.~She, A.~Andonian, M.~S. Mark, and D.~Yamins.
\newblock Unsupervised learning from video with deep neural embeddings.
\newblock In \emph{Proceedings of the {IEEE}/{CVF} {Conference} on {Computer}
  {Vision} and {Pattern} {Recognition}}, 2020.

\end{thebibliography}

\end{document}